%% file: main.tex
\documentclass[lettersize,journal]{IEEEtran}
\usepackage{amsmath,amsfonts}
\usepackage{algorithmic}
\usepackage{algorithm}
\usepackage{array}
\usepackage[caption=false,font=small]{subfig}
\usepackage{textcomp}
\usepackage{stfloats}
\usepackage{url}
\usepackage{verbatim}
\usepackage{graphicx}
\usepackage{cite}
\usepackage[T1]{fontenc}

\usepackage{graphicx}
\usepackage{comment}
\usepackage{amsmath,amssymb,amsfonts}
\usepackage{booktabs}
\usepackage[dvipsnames]{xcolor}
\usepackage{soul}

\usepackage{siunitx}
\usepackage{diagbox}

\usepackage{makecell}
\usepackage{adjustbox}
\usepackage{cite}

\newif\ifarxiv
\arxivtrue

\usepackage{threeparttable} 

\usepackage[pagebackref=false,breaklinks=true,colorlinks,bookmarks=true,bookmarksnumbered=true]{hyperref}

\usepackage[capitalize]{cleveref}
\crefname{section}{Sec.}{Secs.}
\crefname{table}{Tab.}{Tabs.} 
\crefname{figure}{Fig.}{Figs.}
\Crefname{section}{Section}{Sections}
\Crefname{table}{Table}{Tables}
\Crefname{figure}{Figure}{Figures}

\usepackage{multirow}

\usepackage{url}
\pdfpxdimen=\dimexpr 1 in/192\relax  

\usepackage{graphicx}

\usepackage{amsmath}
\DeclareMathOperator*{\argmax}{argmax}
\DeclareMathOperator*{\argmin}{argmin}

\usepackage{caption}
\usepackage{subcaption}
\usepackage{tikz,pgfplots}
\usetikzlibrary{spy,calc}

\usepackage{prettyref}
\usepackage[percent]{overpic}

\newrefformat{fig}{Fig.~\ref{#1}}
\newrefformat{sec}{Sec.~\ref{#1}}
\newrefformat{tab}{Tab.~\ref{#1}}
\newrefformat{algo}{Algo.~\ref{#1}}
\newrefformat{eq}{Eq.~\ref{#1}}

\colorlet{RevisionColor}{PineGreen!80!black}
\newif\ifshowrevisions
\showrevisionsfalse

\newenvironment{revisedtextblock}{%
  \ifshowrevisions
    \color{RevisionColor}
    \captionsetup{textfont={color=RevisionColor}}
    \ignorespaces
  \fi
}{
  \ifshowrevisions
    \captionsetup{textfont={color=black}} 
  \fi
  \unskip
  \ignorespaces
}

\colorlet{RevisionColor}{PineGreen!80!black}
\newif\ifshowrevisionsCR
\showrevisionsCRfalse

\newenvironment{revisedtextblockCR}{%
  \ifshowrevisionsCR%
    \color{RevisionColor}%
    \captionsetup{textfont={color=RevisionColor}}%
  \fi
  \ignorespaces
}{%
  \ifshowrevisionsCR%
    \captionsetup{textfont={color=black}}%
  \fi
  \unskip
  \ignorespaces
}

\begin{document}

\title{EROAM: Event-based Camera Rotational Odometry and Mapping in Real-time}

\author{
Wanli Xing$^{1,2}$, Shijie Lin$^{1,2}$, Linhan Yang$^{1,2}$, Zeqing Zhang$^{1,2}$, Yanjun Du$^{3}$,\\ Maolin Lei$^{4}$, Yipeng Pan$^{1,2}$, Chen Wang$^{1,2}$, and Jia Pan$^{1,2}$$^{\dagger}$
\thanks{This research is supported by the Jiangsu-Hong Kong-Macau Project BZ2024061, RGC grants (GRF 17201025, GRF 17200924, NSFC-RGC Joint Research Scheme N\_HKU705/24, CRF C1073-24GF), and the Natural Science Foundation of China (Project Number 62461160309).}
\thanks{
$^{1}$Department of Computer Science, The University of Hong Kong, Hong Kong SAR, China. {\tt\footnotesize e-mail: wlxing@connect.hku.hk}
\endgraf
$^{2}$Centre for Transformative Garment Production, Hong Kong SAR, China.
\endgraf
$^{3}$Department of Mechanical and Automation Engineering, The Chinese University of Hong Kong, Hong Kong SAR, China.
\endgraf
$^{4}$Humanoids and Human Centered Mechatronics Research Line, Istituto Italiano Di Tecnologia (IIT), Genoa, Italy.
\endgraf
$^\dagger$Corresponding author.}
}

\maketitle

\begin{abstract}
This paper presents EROAM, a novel event-based rotational odometry and mapping system that achieves real-time, accurate camera rotation estimation. Unlike existing approaches that rely on event generation models or contrast maximization, EROAM employs a spherical event representation by projecting events onto a unit sphere and introduces Event Spherical Iterative Closest Point (ES-ICP), a novel geometric optimization framework designed specifically for event camera data. \begin{revisedtextblockCR}The spherical representation simplifies rotational motion formulation while operating in a continuous spherical domain, enabling enhanced spatial resolution.\end{revisedtextblockCR} Our system features an efficient map management approach using incremental k-d tree structures and intelligent regional density control, ensuring optimal computational performance during long-term operation. \begin{revisedtextblock}Combined with parallel point-to-line optimization, EROAM achieves efficient computation without compromising accuracy. Extensive experiments on both synthetic and real-world datasets show that EROAM significantly outperforms state-of-the-art methods in terms of accuracy, robustness, and computational efficiency. Our method maintains consistent performance under challenging conditions, including high angular velocities and extended sequences, where other methods often fail or show significant drift. Additionally, EROAM produces high-quality panoramic reconstructions with preserved fine structural details.\end{revisedtextblock}
\end{abstract}


\section*{Video, Source Code and Data}
Project page: \url{https://wlxing1901.github.io/eroam/}

\input{tools/zoombox}
\input{sections/introduction}
\input{sections/related_works}
\input{sections/methodology}

\input{sections/experiments}

\begin{revisedtextblock}

\begin{revisedtextblockCR}
\input{sections/experiments_ecd}
\end{revisedtextblockCR}

\input{sections/experiments_ablation}
\input{sections/limitation}

\end{revisedtextblock}

\input{sections/conclusion}

\bibliographystyle{IEEEtran}
\bibliography{IEEEabrv,references}

\vfill

\end{document}

%% file: tools/zoombox.tex
\newif\ifblackandwhitecycle
\gdef\patternnumber{0}

\pgfkeys{/tikz/.cd,
    zoombox paths/.style={
        draw=orange,
        very thick
    },
    black and white/.is choice,
    black and white/.default=static,
    black and white/static/.style={ 
        draw=white,   
        zoombox paths/.append style={
            draw=white,
            postaction={
                draw=black,
                loosely dashed
            }
        }
    },
    black and white/static/.code={
        \gdef\patternnumber{1}
    },
    black and white/cycle/.code={
        \blackandwhitecycletrue
        \gdef\patternnumber{1}
    },
    black and white pattern/.is choice,
    black and white pattern/0/.style={},
    black and white pattern/1/.style={    
            draw=white,
            postaction={
                draw=black,
                dash pattern=on 2pt off 2pt
            }
    },
    black and white pattern/2/.style={    
            draw=white,
            postaction={
                draw=black,
                dash pattern=on 4pt off 4pt
            }
    },
    black and white pattern/3/.style={    
            draw=white,
            postaction={
                draw=black,
                dash pattern=on 4pt off 4pt on 1pt off 4pt
            }
    },
    black and white pattern/4/.style={    
            draw=white,
            postaction={
                draw=black,
                dash pattern=on 4pt off 2pt on 2 pt off 2pt on 2 pt off 2pt
            }
    },
    zoomboxarray inner gap/.initial=5pt,
    zoomboxarray columns/.initial=2,
    zoomboxarray rows/.initial=2,
    subfigurename/.initial={},
    figurename/.initial={zoombox},
    zoomboxarray/.style={
        execute at begin picture={
            \begin{scope}[
                spy using outlines={%
                    zoombox paths,
                    width=\imagewidth / \pgfkeysvalueof{/tikz/zoomboxarray columns} - (\pgfkeysvalueof{/tikz/zoomboxarray columns} - 1) / \pgfkeysvalueof{/tikz/zoomboxarray columns} * \pgfkeysvalueof{/tikz/zoomboxarray inner gap} -\pgflinewidth,
                    height=\imageheight / \pgfkeysvalueof{/tikz/zoomboxarray rows} - (\pgfkeysvalueof{/tikz/zoomboxarray rows} - 1) / \pgfkeysvalueof{/tikz/zoomboxarray rows} * \pgfkeysvalueof{/tikz/zoomboxarray inner gap}-\pgflinewidth,
                    magnification=3,
                    every spy on node/.style={
                        zoombox paths
                    },
                    every spy in node/.style={
                        zoombox paths
                    }
                }
            ]
        },
        execute at end picture={
            \end{scope}
            \node at (image.north) [anchor=north,inner sep=0pt] {};
            \node at (zoomboxes container.north) [anchor=north,inner sep=0pt] {};
     \gdef\patternnumber{0}
        },
        spymargin/.initial=0.5em,
        zoomboxes xshift/.initial=1,
        zoomboxes right/.code=\pgfkeys{/tikz/zoomboxes xshift=1},
        zoomboxes left/.code=\pgfkeys{/tikz/zoomboxes xshift=-1},
        zoomboxes yshift/.initial=0,
        zoomboxes above/.code={
            \pgfkeys{/tikz/zoomboxes yshift=1},
            \pgfkeys{/tikz/zoomboxes xshift=0}
        },
        zoomboxes below/.code={
            \pgfkeys{/tikz/zoomboxes yshift=-1},
            \pgfkeys{/tikz/zoomboxes xshift=0}
        },
        caption margin/.initial=4ex,
    },
    adjust caption spacing/.code={},
    image container/.style={
        inner sep=0pt,
        at=(image.north),
        anchor=north,
        adjust caption spacing
    },
    zoomboxes container/.style={
        inner sep=0pt,
        at=(image.north),
        anchor=north,
        name=zoomboxes container,
        xshift=\pgfkeysvalueof{/tikz/zoomboxes xshift}*(\imagewidth+\pgfkeysvalueof{/tikz/spymargin}),
        yshift=\pgfkeysvalueof{/tikz/zoomboxes yshift}*(\imageheight+\pgfkeysvalueof{/tikz/spymargin}+\pgfkeysvalueof{/tikz/caption margin}),
        adjust caption spacing
    },
    calculate dimensions/.code={
        \pgfpointdiff{\pgfpointanchor{image}{south west} }{\pgfpointanchor{image}{north east} }
        \pgfgetlastxy{\imagewidth}{\imageheight}
        \global\let\imagewidth=\imagewidth
        \global\let\imageheight=\imageheight
        \gdef\columncount{1}
        \gdef\rowcount{1}
        \gdef\zoomboxcount{1}
    },
    image node/.style={
        inner sep=0pt,
        name=image,
        anchor=south west,
        append after command={
            [calculate dimensions]
            node [image container,subfigurename=\pgfkeysvalueof{/tikz/figurename}-image] {\phantomimage}
            node [zoomboxes container,subfigurename=\pgfkeysvalueof{/tikz/figurename}-zoom] {\phantomimage}
        }
    },
    color code/.style={
        zoombox paths/.append style={draw=#1}
    },
    connect zoomboxes/.style={
    spy connection path={\draw[draw=none,zoombox paths] (tikzspyonnode) -- (tikzspyinnode);}
    },
    help grid code/.code={
        \begin{scope}[
                x={(image.south east)},
                y={(image.north west)},
                font=\footnotesize,
                help lines,
                overlay
            ]
            \foreach \x in {0,1,...,9} { 
                \draw(\x/10,0) -- (\x/10,1);
                \node [anchor=north] at (\x/10,0) {0.\x};
            }
            \foreach \y in {0,1,...,9} {
                \draw(0,\y/10) -- (1,\y/10);                        \node [anchor=east] at (0,\y/10) {0.\y};
            }
        \end{scope}    
    },
    help grid/.style={
        append after command={
            [help grid code]
        }
    },
}

\newcommand\phantomimage{%
    \phantom{%
        \rule{\imagewidth}{\imageheight}%
    }%
}
\newcommand\zoombox[2][]{
    \begin{scope}[zoombox paths]
        \pgfmathsetmacro\xpos{
            (\columncount-1)*(\imagewidth / \pgfkeysvalueof{/tikz/zoomboxarray columns} + \pgfkeysvalueof{/tikz/zoomboxarray inner gap} / \pgfkeysvalueof{/tikz/zoomboxarray columns} ) + \pgflinewidth
        }
        \pgfmathsetmacro\ypos{
            (\rowcount-1)*( \imageheight / \pgfkeysvalueof{/tikz/zoomboxarray rows} + \pgfkeysvalueof{/tikz/zoomboxarray inner gap} / \pgfkeysvalueof{/tikz/zoomboxarray rows} ) + 0.5*\pgflinewidth
        }
        \edef\dospy{\noexpand\spy [
            #1,
            zoombox paths/.append style={
                black and white pattern=\patternnumber
            },
            every spy on node/.append style={#1},
            x=\imagewidth,
            y=\imageheight
        ] on (#2) in node [anchor=north west] at ($(zoomboxes container.north west)+(\xpos pt,-\ypos pt)$);}
        \dospy
        \pgfmathtruncatemacro\pgfmathresult{ifthenelse(\columncount==\pgfkeysvalueof{/tikz/zoomboxarray columns},\rowcount+1,\rowcount)}
        \global\let\rowcount=\pgfmathresult
        \pgfmathtruncatemacro\pgfmathresult{ifthenelse(\columncount==\pgfkeysvalueof{/tikz/zoomboxarray columns},1,\columncount+1)}
        \global\let\columncount=\pgfmathresult
        \ifblackandwhitecycle
            \pgfmathtruncatemacro{\newpatternnumber}{\patternnumber+1}
            \global\edef\patternnumber{\newpatternnumber}
        \fi
    \end{scope}
}

%% file: sections/introduction.tex
\input{figs_and_tabs/fig_head_graph}

\section{Introduction}

\IEEEPARstart{R}{otational} motion estimation represents a fundamental challenge in computer vision and robotics, serving as a cornerstone for various applications from visual odometry to camera stabilization and panoramic image creation \cite{martinec2007robust}. While traditional frame-based cameras have been widely used for this task, they struggle with rapid rotations due to inherent limitations such as motion blur, limited exposure control, and large inter-frame displacements that compromise data association and motion estimation accuracy.

Event cameras \cite{lichtsteiner2008128, brandli2014240, gallego2020event} offer a promising alternative through their bio-inspired design. Unlike conventional cameras that capture intensity frames at fixed intervals, these sensors operate asynchronously, measuring and reporting per-pixel brightness changes with microsecond precision. Each event is encoded as a tuple $e_k = (\mathbf{x_k}, t_k, p_k)$, where $\mathbf{x_k} = (u_k, v_k)^T$ denotes the pixel location, $t_k$ the timestamp, and $p_k$ the polarity of the brightness change \cite{cm-w}. This unique operating principle enables exceptional temporal resolution, high dynamic range (up to \SI{140}{dB}), and low power consumption (\SI{20}{mW}) \cite{gallego2020event}, making event cameras particularly well-suited for challenging scenarios involving high-speed motion or extreme lighting conditions.

The advantages of event cameras have been demonstrated across numerous applications, including SLAM and odometry \cite{niu2024esvo2, 9809788, 9810191, vidal2018ultimate, lin2023fast, qu2024implicit, kueng2016low}, robotic navigation \cite{sun2021autonomous, falanga2020dynamic, iaboni2021event, bhattacharya2024monocular}, feature extraction and tracking \cite{afshar2020event, seok2020robust, huang2023eventpoint, lin2020efficient, zhu2017event}, optical flow estimation \cite{shiba2022fast, gehrig2021raft, shiba2022secrets, lee2020spike}, video enhancement \cite{scheerlinck2020fast, rebecq2019events, rebecq2019high}, 3D reconstruction \cite{klenk2023nerf, hwang2023ev, rudnev2023eventnerf, xue2024event}, and optical systems \cite{lin2022autofocus, liao2022synthetic, he2024microsaccade, lin2024neuromorphic}. Recent manufacturing advances have further accelerated their adoption by enabling higher-resolution sensors at more accessible price points \cite{finateu20205}.

In the specific context of rotational motion estimation, event cameras offer unique advantages through their high temporal resolution and freedom from motion blur, enabling potential accurate capture of rapid rotational movements that conventional cameras struggle to handle. Existing event-based approaches broadly fall into two categories. Event Generation Model (EGM) based methods \cite{smt, rtpt} explicitly model the event triggering process, requiring careful consideration of the contrast threshold and event generation mechanisms. In contrast, Contrast Maximization (CM) based approaches \cite{cm-w, cmax-slam, cm-gae} estimate angular velocity by optimizing the alignment of events over temporal windows. While both approaches have demonstrated promising performance in scenarios with moderate and short-term rotational motion, they face significant challenges in more complex scenarios: EGM methods struggle with the complexity of accurately modeling event generation, while CM techniques may lack robustness during highly dynamic rotations involving rapid direction changes.  Moreover, both approaches operate on discretized representations with predefined resolutions, introducing inherent quantization errors. Beyond these algorithmic limitations, both approaches often incur substantial computational overhead, limiting their real-time applicability. \begin{revisedtextblock}Additionally, these methods typically lack efficient mechanisms for incremental map updates during long-term operation, leading to increasing computational burden and memory consumption as the observed scene expands.\end{revisedtextblock}

Drawing inspiration from successful techniques in LiDAR SLAM \cite{zhang2014loam, shan2020lio, xu2021fast, bai2022faster, xu2022fast}, we propose a novel approach that addresses these limitations. As illustrated in \prettyref{fig:head_fig}, our method introduces three key innovations: a spherical event representation, the Event Spherical Iterative Closest Point (ES-ICP) algorithm, and an efficient incremental map management approach. The spherical representation projects events onto a unit sphere, assigning continuous $\mathbb{R}^3$ coordinates instead of working with discrete pixel locations. ES-ICP is a novel geometric optimization framework specifically designed for event data, which efficiently aligns sparse event point clouds on the unit sphere through parallel point-to-line optimization. Unlike existing methods that operate on discretized maps with fixed resolutions, our approach maintains both tracking and mapping in continuous spherical space, completely decoupling the core estimation pipeline from any discretization requirements. This continuous representation, combined with the innovative ES-ICP algorithm, not only enables precise motion estimation but also allows for flexible panoramic image generation at arbitrary resolutions as a post-processing step. \begin{revisedtextblock}Our incremental map management strategy utilizing dynamic k-d tree structures and regional density control ensures optimal computational performance even during extended operation.\end{revisedtextblock} The result is a real-time capable system that provides accurate and robust event camera rotation estimation.

Our main contributions are:
\begin{enumerate}
    \item We introduce a novel spherical event representation that operates entirely in continuous $\mathbb{R}^3$ space, with efficient incremental map management, enabling accurate motion estimation while decoupling the tracking process from panorama generation, which allows for flexible panoramic image creation at arbitrary resolutions;
    
    \item We develop the Event Spherical Iterative Closest Point (ES-ICP) algorithm, which efficiently matches and aligns event projections in continuous spherical space for robust rotational motion estimation;
    
    \item We conduct comprehensive experimental validation on both synthetic and real-world datasets, demonstrating superior accuracy and robustness across diverse scenarios, and release our complete implementation and datasets to benefit the research community.
\end{enumerate}

%% file: figs_and_tabs/fig_head_graph.tex
\begin{figure}[t]
    \centering
    {
    \newlength{\trimlen}
    \setlength{\trimlen}{44mm} 
    \newlength{\trimlenB}
    \setlength{\trimlenB}{28mm} 
    
    \subfloat[3DoF rotational motion]{%
        \includegraphics[width=0.49\columnwidth]
        {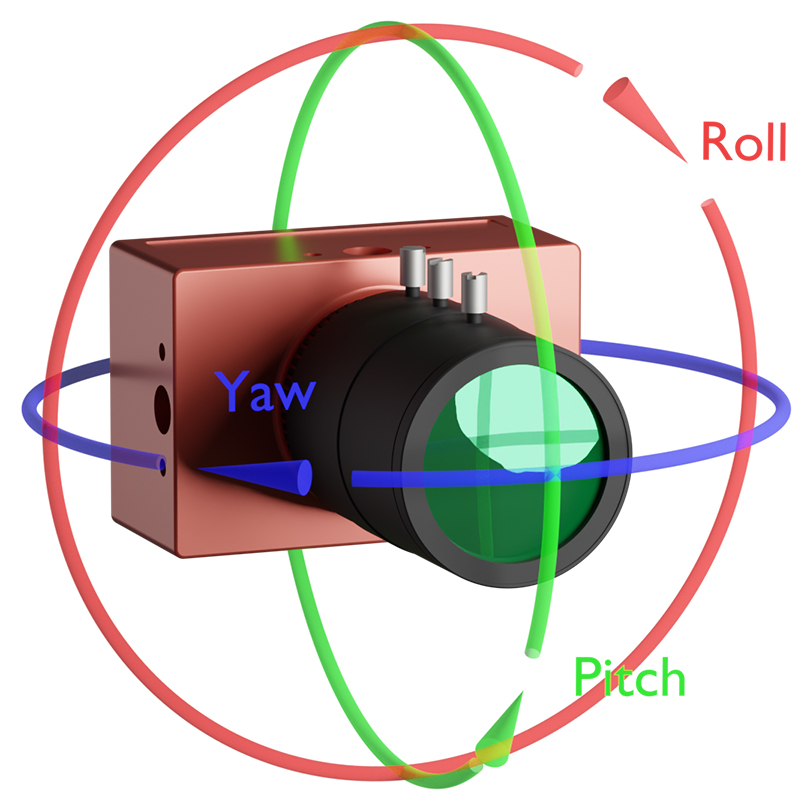}%
        \label{fig:subfig1}%
    }%
    \hfill
    \subfloat[ES-ICP alignment]{%
        \includegraphics[width=0.49\columnwidth,
        trim={\trimlen} {47mm} {\trimlen} {\trimlen},
        clip]{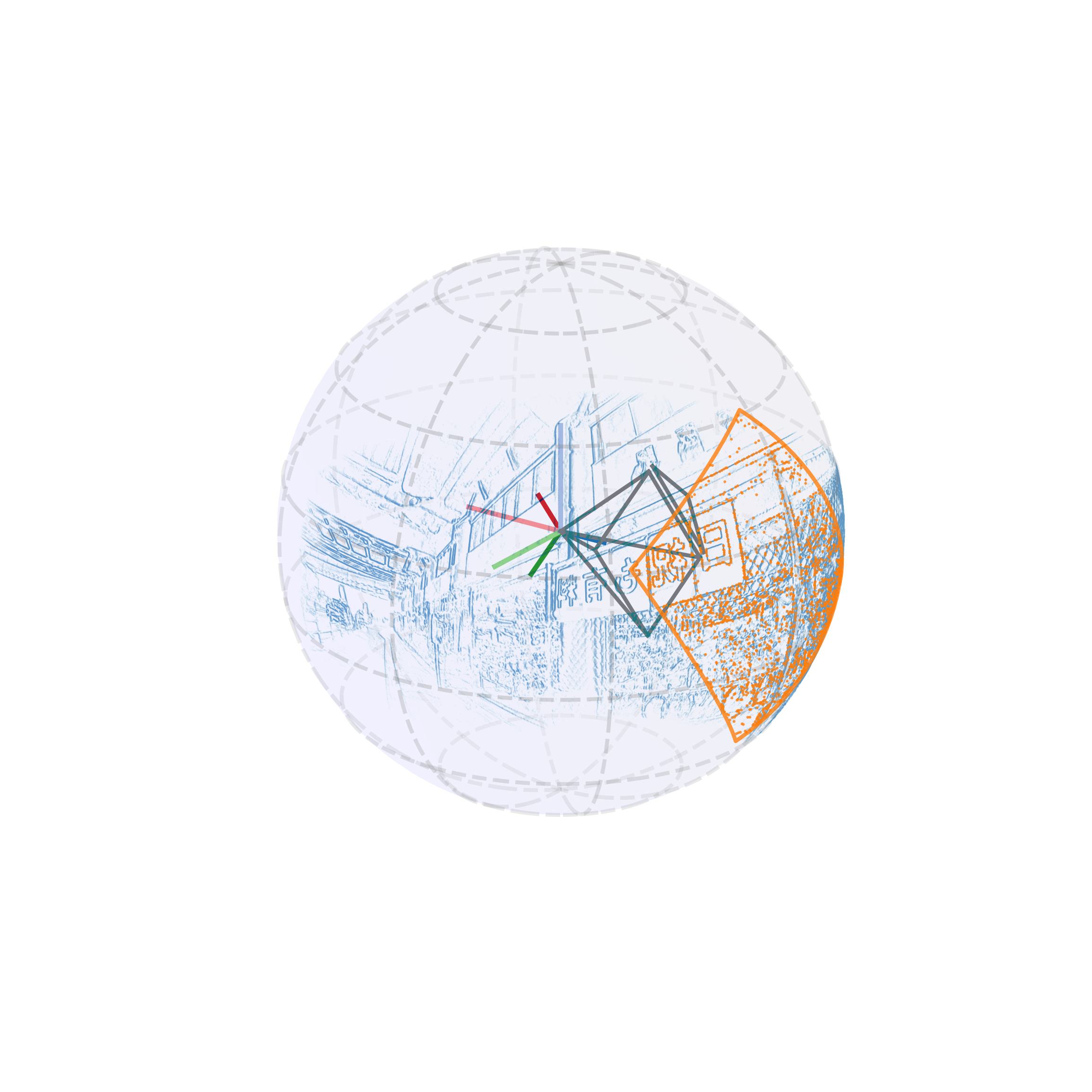}%
        \label{fig:spherical_icp}%
    }%

    \setlength{\fboxrule}{0.1pt}
    \vspace{1em} 
    \subfloat[Panoramic reconstruction with detailed views]{%
    \begin{tikzpicture}[
        zoomboxarray,
        spymargin=0.1em,  
        caption margin=2ex,  
        zoomboxes below,
        zoomboxarray columns=3,
        zoomboxarray rows=1,
        connect zoomboxes,
        zoombox paths/.append style={thick, red}]
        \node [image node] { {\color{gray}\fbox{\includegraphics[width=0.98\columnwidth]{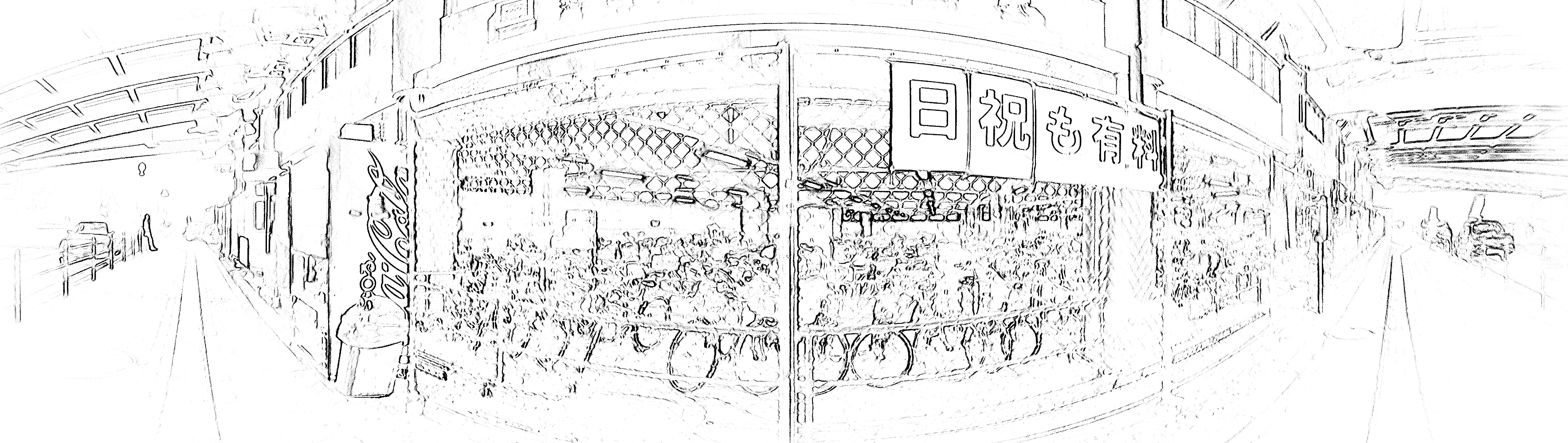}}} };
        
        \zoombox[magnification=4]{0.08,0.70}
        \zoombox[magnification=4]{0.23,0.40}
        \zoombox[magnification=4]{0.615,0.73}
    \end{tikzpicture}
    \label{fig:panoramic_head}%
    }%
    }
    \caption{Event-based 3DoF camera pose estimation and panoramic reconstruction. (a) Our method accurately estimates the 3DoF pose of an event camera from continuous event streams. (b) Events are projected onto a unit sphere and aligned using our novel Event Spherical Iterative Closest Point (ES-ICP) algorithm. (c) The aligned event sphere enables the reconstruction of a clear panoramic image with preserved fine details.}
    \label{fig:head_fig}
\end{figure}

%% file: sections/related_works.tex
\section{Related Works}

\input{figs_and_tabs/tab_related_works}

Event-based rotational motion estimation presents unique challenges and opportunities due to the asynchronous nature of event cameras. As summarized in \prettyref{tab:related_works_comparison}, existing approaches have evolved along two main paradigms: Event Generation Model (EGM) based methods, which explicitly model the event triggering process, and Contrast Maximization (CM) based methods, which focus on event alignment optimization. This section analyzes the fundamental principles, practical implementations, and inherent limitations of each approach, providing insights into their relative strengths and weaknesses.

\subsection{Event Generation Model Based Methods}
\begin{revisedtextblock}
EGM-based approaches model event triggering conditions. An event triggers when log intensity change exceeds the threshold $C$:
\end{revisedtextblock}

\begin{equation}
\Delta L(\mathbf{x}_k, t_k) = L(\mathbf{x}_k, t_k) - L(\mathbf{x}_k, t_k - \Delta t_k) = p_k C
\end{equation}
where $\mathbf{x}_k = (u_k, v_k)^\top$ represents the pixel coordinates, $L(\mathbf{x}_k, t_k)$ is the log intensity at pixel $\mathbf{x}_k$ at time $t_k$, $\Delta t_k$ is the time elapsed since the last event at the same pixel, $p_k \in \{+1, -1\}$ is the polarity of the event, and $C > 0$ is the contrast threshold \cite{lichtsteiner2008128, gallego2020event}.

In practice, real event cameras exhibit more complex behavior due to sensor noise, transistor mismatch, and varying illumination conditions. This probabilistic nature is often modeled using normal distributions centered at the contrast threshold $C$ \cite{lichtsteiner2008128, guo2022low, joubert2021event}. For small time intervals $\Delta t_k$, the relationship between events and temporal brightness changes can be approximated as:
\begin{equation}
\frac{\partial L}{\partial t}(\mathbf{x}_k, t_k) \approx \frac{p_k C}{\Delta t_k}
\end{equation}
This approximation serves as the foundation for various event-based algorithms \cite{pan2019bringing, pan2020high, paredes2021back, munda2018real}.

\subsubsection{Simultaneous Mosaicing and Tracking (SMT)}

SMT \cite{smt} introduced the first comprehensive approach to event-based visual processing by simultaneously addressing camera tracking and scene reconstruction. The method employs a dual-filter architecture: a particle filter for camera rotation tracking and pixel-wise Extended Kalman Filters (EKFs) for mapping. In the tracking module, camera poses are estimated by evaluating how well each pose hypothesis aligns with the current gradient map using an event likelihood model. Concurrently, the mapping module maintains and updates a gradient map through EKF filtering of incoming events, with each pixel independently tracking its gradient estimate and uncertainty.

The key innovation of SMT lies in its probabilistic treatment of both tracking and mapping components, which work symbiotically: the tracker leverages the gradient map for pose evaluation, while the mapping system utilizes pose estimates for gradient updates. This approach effectively handles the asynchronous nature of event data while maintaining a coherent scene representation.

\subsubsection{Real-Time Panoramic Tracking (RTPT)}

RTPT \cite{rtpt} advances event-based motion estimation through a direct optimization approach. The method introduces a probabilistic map that tracks event occurrence likelihoods at each spatial location, dynamically updating as the camera moves. Camera pose estimation is formulated as an energy minimization problem that elegantly combines re-projection error with temporal motion smoothness.

A significant contribution of RTPT is its probabilistic map representation, which maintains both observed event counts and possible event occurrences for each location. This formulation naturally accommodates the sparse and asynchronous characteristics of event data while preserving scene geometry in map updates. However, achieving real-time performance requires GPU acceleration, and the method faces challenges with map updates at higher resolutions.

\noindent\textbf{Limitations of EGM Based Methods.} EGM-based approaches face several fundamental challenges:
\begin{itemize}
    \item Both SMT and RTPT struggle with robustness to sensor noise and varying illumination conditions, despite their different approaches to handling contrast threshold specification.
    \item The computational requirements for map updates become prohibitive at higher resolutions, even with GPU acceleration.
    \item The use of discretized map storage inherently introduces quantization errors that affect motion estimation accuracy.
\end{itemize}

\subsection{Contrast Maximization Based Methods}

The contrast maximization framework, introduced by Gallego and Scaramuzza \cite{cm-w, gallego2018unifying}, represents a fundamentally different approach to event-based motion estimation. The key insight is that when events are warped (rotated) according to the true motion parameters, they align to create sharp edge patterns.

The framework formulates motion estimation as an optimization problem:
\begin{equation}
\boldsymbol{\omega}^* = \argmax\nolimits_{\boldsymbol{\omega}} \text{Var}(I(\boldsymbol{\omega}, \mathcal{E}))
\end{equation}
where $\boldsymbol{\omega}$ is the angular velocity vector to be estimated, and $I(\boldsymbol{\omega}, \mathcal{E})$ represents an image formed by warped events. This image is created by:
\begin{equation}
I(\mathbf{x}; \boldsymbol{\omega}) = \sum\nolimits_{k} \pm_k \delta(\mathbf{x} - \mathbf{x}'_k(\boldsymbol{\omega}))
\end{equation}
where $\pm_k$ is the polarity of each event, and $\mathbf{x}'_k(\boldsymbol{\omega})$ represents the warped (rotated) position of the original event according to the candidate angular velocity $\boldsymbol{\omega}$. The Dirac delta function $\delta$ accumulates event polarities at the warped locations.

The quality of motion estimation is measured through the image variance:
\begin{equation}
\text{Var}(I(\boldsymbol{\omega}, \mathcal{E})) = \frac{1}{|\Omega|} \int_{\Omega} (I(\boldsymbol{\omega}, \mathcal{E})(\mathbf{x}) - \mu(I(\boldsymbol{\omega}, \mathcal{E})))^2 d\mathbf{x}
\end{equation}
where $\Omega$ is the image domain. The variance measures the contrast of the warped event image - when events are correctly aligned according to the true motion, they form sharp edges resulting in maximum contrast.

\subsubsection{Global Events Alignment for Rotational Motion Estimation (CM-GAE)}

CM-GAE \cite{cm-gae} enhances the contrast maximization framework through a dual-optimization strategy aimed at reducing drift. The method maintains a global event image by continuously aligning observed events to the initial camera coordinate frame, enabling both local and global consistency checks. Beyond maximizing contrast within temporal windows, CM-GAE introduces a second optimization step that aligns locally warped event images with the accumulated global event image, effectively reducing the drift accumulation common in local optimization methods.

\subsubsection{Event-based Rotational SLAM System (CMax-SLAM)}

CMax-SLAM \cite{cmax-slam} represents a significant advancement as the first complete event-based rotational SLAM system incorporating both front-end and back-end components. The front-end employs contrast maximization for angular velocity estimation, while the back-end performs continuous-time trajectory refinement using B-splines. A key innovation lies in its formulation of bundle adjustment based on contrast maximization, optimizing camera trajectories while simultaneously generating sharp panoramic maps as a by-product of the optimization process.

\noindent\textbf{Limitations of CM Based Methods.} Contrast maximization approaches face several key challenges:
\begin{itemize}
    \item The temporal window selection presents an inherent trade-off between having sufficient events for optimization and maintaining the constant angular velocity assumption.
    \item Iterative optimization and repeated event warping operations impose significant computational overhead.
    \item The optimization process can encounter local optima \cite{shiba2022event} and degenerate solutions affecting estimation reliability.
    \item Global optimization approaches introduce substantial computational demands that impact real-time performance.
\end{itemize}

%% file: figs_and_tabs/tab_related_works.tex
\begin{table*}[ht]
\centering
\caption{\begin{revisedtextblockCR}Comparison of Event-based Rotational Motion Estimation Methods\end{revisedtextblockCR}}
\label{tab:related_works_comparison}
\resizebox{\textwidth}{!}{%
\begin{tabular}{lcccccc}
\toprule
Method & Principle & Real-time & GPU Required & Representation & Threshold $C$ Dependency & Polarity Dependency \\
\midrule
SMT \cite{smt} & EGM & No & No & Discrete & Yes & Yes \\
RTPT \cite{rtpt} & EGM & No & Yes & Discrete & No & No \\
CM-GAE \cite{cm-gae} & Contrast Maximization & No & No & Discrete & No & Yes \\
CMax-SLAM \cite{cmax-slam} & Contrast Maximization & No & No & Discrete & No & No \\
EROAM (\textbf{ours}) & ES-ICP & Yes & No & Continuous Spherical & No & No \\
\bottomrule
\end{tabular}%
}
\end{table*}

%% file: sections/methodology.tex
\section{Methodology}
\input{figs_and_tabs/fig_system_overview}
\input{figs_and_tabs/fig_spherical_projection}
\input{sections/methodology_01_system_overview}
\input{figs_and_tabs/tab_symbols}

\input{sections/methodology_02_event_spherical_representation}
\input{figs_and_tabs/fig_pixel_error}


\input{figs_and_tabs/fig_frame_formation}
\input{sections/methodology_04_event_spherical_formation}

\input{figs_and_tabs/fig_es_icp}
\input{sections/methodology_05_es_icp}

\input{sections/methodology_06_map}

\input{figs_and_tabs/fig_spherical_to_panoramic}
\input{sections/methodology_07_panoramic}

%% file: figs_and_tabs/fig_system_overview.tex
\definecolor{trackingcolor}{RGB}{182,40,28}  
\definecolor{mappingcolor}{RGB}{74,113,164}   

\begin{figure}[!t]
    \centering
    \includegraphics[width=\columnwidth,
        trim={17mm} {45mm} {17mm} {48mm},
        clip]{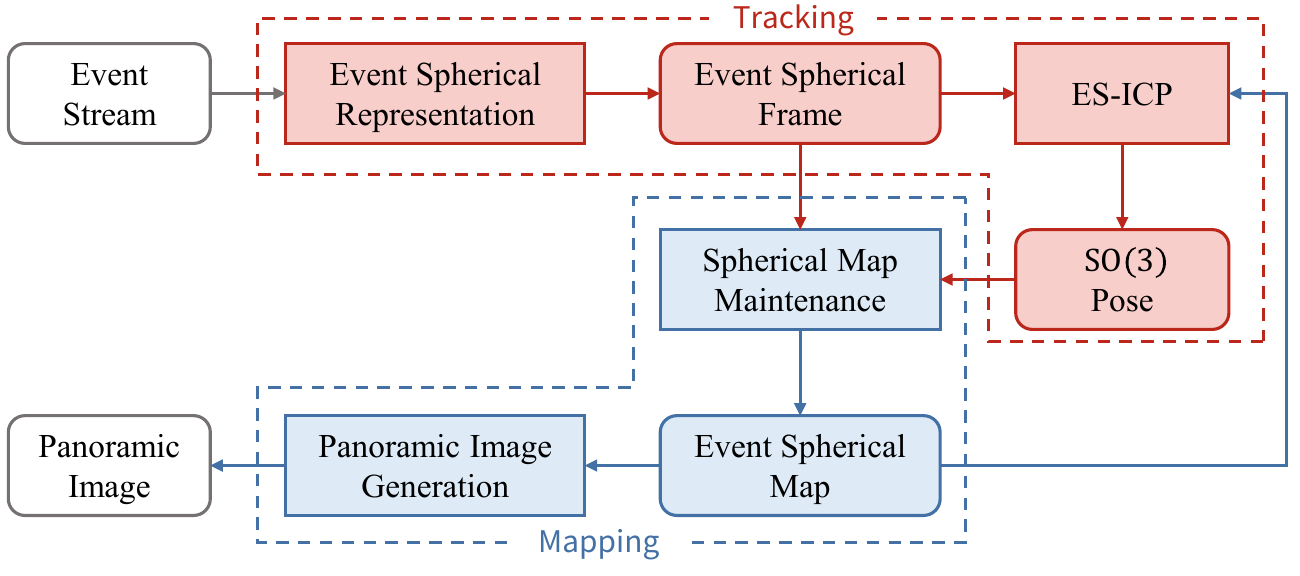}%

    \caption{System Overview: The proposed event-based rotational motion estimation system consists of two main modules. The \textcolor{trackingcolor}{tracking module} processes event streams to estimate $SO(3)$ pose using spherical representation and ES-ICP. The \textcolor{mappingcolor}{mapping module} maintains and updates the spherical event map, which supports both tracking and panoramic image generation.}
    \label{fig:system_overview}
\end{figure}


%% file: figs_and_tabs/fig_spherical_projection.tex
\definecolor{pixel_image_projection}{RGB}{255,135,135}
\definecolor{spherical_point_projection}{RGB}{238,129,27}
\definecolor{normalized_image_projection}{RGB}{55,159,49}
\definecolor{original_point}{RGB}{48,120,179}

\begin{figure}[!t]
    \vspace{-3mm}

    \centering
    \begin{overpic}[width=0.98\columnwidth,
        trim={30mm} {45mm} {30mm} {45mm},
        clip]{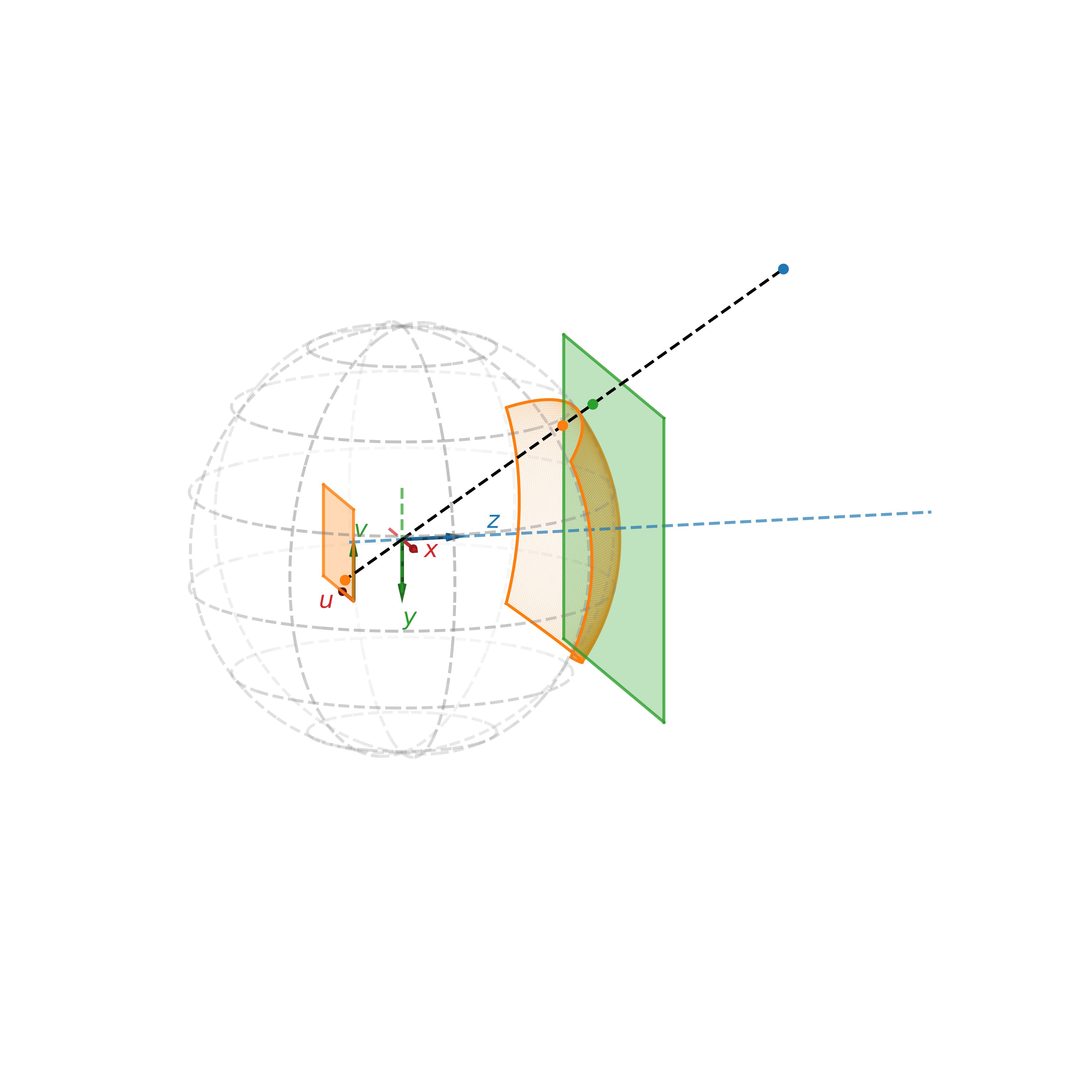}
        \put(82,70){\color{original_point}$\mathbf{P}_i$}
        \put(46,62){\color{spherical_point_projection}$\mathbf{p}^s_i$}
        \put(58,65){\color{normalized_image_projection}$\mathbf{p}^n_i$}
        \put(15,40){\color{pixel_image_projection}$\mathbf{x}^I_i$}
    \end{overpic}
    \vspace{-8mm}
    \caption{Spherical projection process: The 3D point \textcolor{original_point}{$\mathbf{P}_i$} is first projected onto the image plane as \textcolor{pixel_image_projection}{$\mathbf{x}^I_i$}. After undistortion and normalization, it becomes \textcolor{normalized_image_projection}{$\mathbf{p}^n_i$} on the normalized image plane. Finally, \textcolor{normalized_image_projection}{$\mathbf{p}^n_i$} is projected onto the unit sphere, resulting in the spherical point \textcolor{spherical_point_projection}{$\mathbf{p}^s_i$}.}
    \label{fig:spherical_projection}
\end{figure}
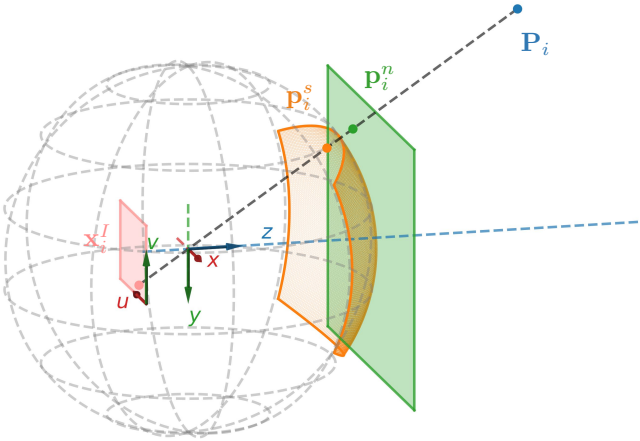

%% file: sections/methodology_01_system_overview.tex
\subsection{System Overview}
\label{subsec:system_overview}

Our proposed event-based rotational motion estimation system, illustrated in \prettyref{fig:system_overview}, comprises two main components: a tracking module and a mapping module. \begin{revisedtextblock}The system is designed for real-time operation at high frequencies (\SI{1000}{\Hz}) with stable computational costs throughout extended operation.\end{revisedtextblock}

The tracking module begins by transforming incoming events into a spherical representation (\prettyref{subsec:event_spherical_representation}). These spherical events are then grouped into event frames and undergo motion compensation (\prettyref{subsec:event_spherical_frame_formation}). The compensated event frames serve as input for our novel Event Spherical Iterative Closest Point (ES-ICP) algorithm (\prettyref{subsec:es_icp}). \begin{revisedtextblock}High-frequency operation is crucial for ES-ICP's success, as it ensures that consecutive frames have minimal rotation differences, providing excellent initial guesses that help the algorithm avoid local minima inherent to ICP-based optimization.\end{revisedtextblock} The ES-ICP algorithm estimates the camera's rotational pose in $SO(3)$, leveraging the spherical event map maintained by the mapping module.

Concurrently, the mapping module selectively updates the spherical event map using key frames determined by significant rotational motion (\prettyref{subsec:map_maintenance}). \begin{revisedtextblock}Unlike existing methods that may suffer from increasing computational burden over time, our mapping strategy employs incremental updates with ikd-Tree and our Regional Density Management to ensure that computational costs remain stable and bounded regardless of sequence length or scene complexity.\end{revisedtextblock} This updated map not only aids in subsequent tracking iterations but also serves as the basis for panoramic image generation (\prettyref{subsec:panoramic_generation}).

This iterative tracking and mapping process achieves robust high-frequency rotation estimation while maintaining predictable real-time performance.

\prettyref{tab:symbols} summarizes the key symbols and variables used throughout our methodology.

%% file: figs_and_tabs/tab_symbols.tex
\begin{table}[t]
\centering
\caption{Summary of Key Symbols and Variables}
\label{tab:symbols}
\begin{tabular}{cl}
\toprule
Symbol & Description \\
\midrule
$e_i$ & Individual event $(u_i, v_i, t_i, p_i)$ \\
$\mathbf{x}^I_i$ & Pixel coordinates $(u_i, v_i)^T$ \\
$\mathbf{K}$ & Camera intrinsic matrix \\
$\mathbf{f}(\cdot)$ & Distortion function \\
$\mathbf{f}^{-1}(\cdot)$ & Undistortion function \\
$\mathbf{x}^u_i$ & Undistorted pixel coordinates \\
$\mathbf{p}^n_i$ & Normalized coordinates on virtual imaging plane \\
$\mathbf{p}^s_i$ & Spherical representation of an event \\
$\mathbf{p}^c_i$ & Cylindrical projection of a point \\
$\mathcal{E}^s$ & Event spherical frame \\
$f$ & Event spherical frame formation frequency \\
$n$ & Number of events per spherical frame \\
$\mathbf{R}$ & Rotation matrix in $SO(3)$ \\
$\mathbf{t}$ & Translation vector \\
$\boldsymbol{\omega}$ & Angular velocity of the event camera \\
$\mathcal{M}^s$ & Spherical event map \\
$\theta_t$ & Rotation threshold for key frames \\
$\mathbf{d}$ & Direction vector of fitted line (ES-ICP) \\
$\mathbf{c}$ & Point on fitted line (ES-ICP) \\
$\mathbf{J}_i$ & Jacobian matrix for each point (ES-ICP) \\
$\mathbf{H}$ & Approximated Hessian matrix (ES-ICP) \\
$\Delta\mathbf{x}$ & Incremental update in $\mathfrak{so}(3)$ (ES-ICP) \\
$\mathbf{g}$ & Gradient vector (ES-ICP) \\
$\phi_v$ & Vertical field of view (panoramic) \\
$\phi_h$ & Horizontal span angle (panoramic) \\
$\exp(\cdot^\wedge)$ & Exponential map from $\mathfrak{so}(3)$ to $SO(3)$ \\
$\log(\cdot)^\vee$ & Logarithm map from $SO(3)$ to $\mathfrak{so}(3)$ \\
$(\cdot)^\wedge$ & Hat operator (vector to skew-symmetric matrix) \\
$(\cdot)^\vee$ & Vee operator (skew-symmetric matrix to vector) \\
$\|\cdot\|_2$ & Euclidean (L2) norm \\
\bottomrule
\end{tabular}
\end{table}

%% file: sections/methodology_02_event_spherical_representation.tex
\subsection{Event Spherical Representation}
\label{subsec:event_spherical_representation}

In event-based rotational motion estimation, the choice of event representation is crucial for both accuracy and computational efficiency. We propose a spherical representation of events, which offers two significant advantages over traditional pixel-plane approaches:
1) Simplification of rotational motion formula, and
2) \begin{revisedtextblockCR}Enhanced spatial resolution and mapping within a continuous domain.\end{revisedtextblockCR}

\subsubsection{Simplified Motion Geometry on Unit Sphere}

In pure rotational motion, considering events on the pixel plane involves a complex series of transformations including rotation, perspective projection, and distortion. \prettyref{fig:spherical_projection} illustrates this process, showing the transformation from a 3D point to its spherical representation. Let's consider a 3D point $\mathbf{P}_i$ and its projection onto the image plane $\mathbf{x}^I_i$. The process can be described as follows \cite{zhang2000flexible}:
\begin{equation}
\mathbf{x}^I_i = \mathbf{f}(\mathbf{K}[\mathbf{R}|\mathbf{t}]\mathbf{P}_i)
\end{equation}
where $\mathbf{K}$ is the camera intrinsic matrix, $\mathbf{f}$ is the distortion function, and $[\mathbf{R}|\mathbf{t}]$ represents the rigid body transformation from the world coordinate system to the camera coordinate system.

Now, consider the same point $\mathbf{P}_i$ observed from two camera views $I_1$ and $I_2$ related by a rotation $\mathbf{R}_{12}$ (denoting the rotation from reference frame 2 to reference frame 1, and $\mathbf{t}_{12}=\mathbf{0}$). The corresponding pixel coordinates in the two views, $\mathbf{x}^{I_1}_i$ and $\mathbf{x}^{I_2}_i$, are related by:
\begin{equation}
\mathbf{x}^{I_1}_i = \mathbf{f}(\mathbf{K}\mathbf{R}_{12}\mathbf{K}^{-1}\mathbf{f}^{-1}(\mathbf{x}^{I_2}_i))
\end{equation}
where $\mathbf{f}^{-1}$ is the undistortion function. This expression, involving multiple non-linear transformations, is both mathematically complex and computationally demanding.

In contrast, on a unit sphere, the projection of a 3D point $\mathbf{P}_i$ can be expressed as:
\begin{equation}
\mathbf{p}^s_i = \frac{[\mathbf{R}|\mathbf{t}]\mathbf{P}_i}{\|[\mathbf{R}|\mathbf{t}]\mathbf{P}_i\|_2}
\end{equation}
For pure rotational motion, $\mathbf{t} = \mathbf{0}$. Moreover, since $\mathbf{R}$ is a rotation matrix, its determinant is 1, which preserves the vector's magnitude. Therefore, the equation simplifies to:
\begin{equation}
\mathbf{p}^s_i = \frac{\mathbf{R}\mathbf{P}_i}{\|\mathbf{P}_i\|_2}
\end{equation}
Consequently, for two spherical projections $\mathbf{p}^{s_1}_{i}$ and $\mathbf{p}^{s_2}_{i}$ from views differing by a rotation $\mathbf{R}_{12}$, we have:
\begin{equation}
\mathbf{p}^{s_1}_{i} = \mathbf{R}_{12}\mathbf{p}^{s_2}_{i}
\end{equation}
This formulation provides a direct and simple representation of rotational motion on the unit sphere, significantly reducing computational complexity compared to the pixel-plane formulation.

\subsubsection{\begin{revisedtextblockCR}Enhanced Spatial Resolution in Continuous Spherical Domain\end{revisedtextblockCR}}
While event cameras provide discrete pixel locations, our spherical representation offers significant advantages in spatial resolution and mapping continuity. On the pixel plane, each event is confined to a discrete grid, with each pixel corresponding to a specific field of view (FOV). This discretization limits the precision of motion estimation, as movements within a single pixel's FOV become indistinguishable. 

In contrast, the spherical representation, although derived from discrete pixel data, allows for a more nuanced positioning of events in 3D space. This approach enables us to maintain sub-pixel precision in event positioning, as the spherical coordinates are not confined to a discrete grid. \begin{revisedtextblockCR}Consequently, we maintain a discretely sampled spherical event map within this continuous domain, where events can be placed at arbitrary positions on the unit sphere.\end{revisedtextblockCR} This enhanced spatial resolution is particularly beneficial for representing and tracking small motions, even when they don't result in a change of pixel coordinates.

\prettyref{fig:pixel_error} illustrates this advantage, showing how a small camera rotation can lead to a detectable change in the spherical representation (\textcolor{spherical_point_projection}{$\Delta\mathbf{p}^s_i$}), while potentially resulting in no change in the pixel coordinates (\textcolor{pixel_image_projection}{$\Delta\mathbf{x}^I_i = 0$}). This capability significantly enhances the precision of our rotation estimation algorithm, allowing for more accurate tracking of camera motion.

Given these advantages, we implement the Event Spherical Representation as follows:

Given an event $e_i = (\mathbf{x}^I_i, t_i, p_i)$, where $\mathbf{x}^I_i = (u_i, v_i)^\top$ denotes the pixel coordinates, $t_i$ the timestamp, and $p_i$ the polarity, we perform the following steps (illustrated in \prettyref{fig:spherical_projection}):

\begin{enumerate}
    \item \textbf{Undistortion}: First, we undistort the pixel coordinates $\mathbf{x}^I_i$ using the camera's intrinsic parameters and distortion coefficients:
    \begin{equation}
    \mathbf{x}_i^u = \mathbf{f}^{-1}(\mathbf{x}^I_i)
    \end{equation}

    \item \textbf{Normalization}: We then normalize the undistorted coordinates using the camera's intrinsic matrix $\mathbf{K}$:
    \begin{equation}
    \mathbf{p}_i^n = \mathbf{K}^{-1} \begin{pmatrix} \mathbf{x}_i^u \\ 1 \end{pmatrix}
    \end{equation}

    \item \textbf{Spherical Projection}: Finally, we project the normalized coordinates onto the unit sphere:
    \begin{equation}
    \mathbf{p}^s_i = \frac{\mathbf{p}_i^n}{\|\mathbf{p}_i^n\|_2}
    \end{equation}
\end{enumerate}

The resulting $\mathbf{p}^s_i$ is a 3D point on the unit sphere, representing the direction of the event in 3D space. This spherical representation allows us to treat events as points on a continuous surface, facilitating more accurate rotational motion estimation.

%% file: figs_and_tabs/fig_pixel_error.tex
\begin{figure}[!t]
    \centering
    \begin{overpic}[width=0.98\columnwidth,
        trim={30mm} {45mm} {30mm} {40mm},
        clip]{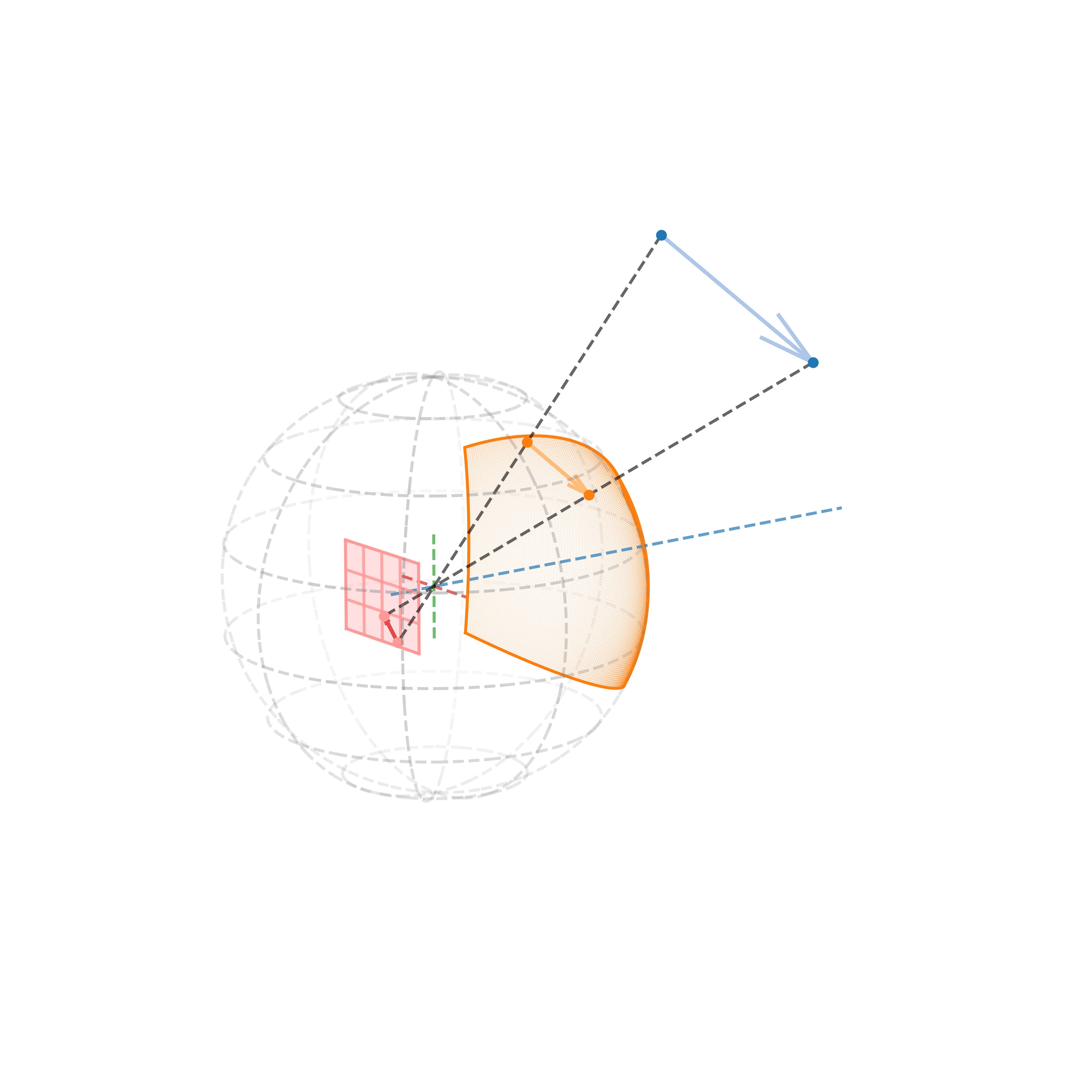}
        \put(74,74){\color{original_point}$\Delta\mathbf{P}_i$}
        \put(44,45){\color{spherical_point_projection}$\Delta\mathbf{p}^s_i$}
        \put(28,20){\color{pixel_image_projection}$\Delta\mathbf{x}^I_i$}
    \end{overpic}
    \caption{Illustration of motion sensitivity differences between spherical and pixel representations. When the event camera undergoes a small rotation, the 3D point position relative to the camera changes by \textcolor{original_point}{$\Delta\mathbf{P}_i$}. This results in a measurable change \textcolor{spherical_point_projection}{$\Delta\mathbf{p}^s_i$} on the spherical surface. However, in the pixel coordinate system, the motion may not cause a change in pixel location, resulting in \textcolor{pixel_image_projection}{$\Delta\mathbf{x}^I_i = 0$}. This demonstrates the higher sensitivity and continuous nature of the spherical representation compared to the discrete pixel representation.}
    \label{fig:pixel_error}
\end{figure}

%% file: figs_and_tabs/fig_frame_formation.tex
\begin{figure}[!t]
    \centering
    \begin{overpic}[width=0.98\columnwidth]{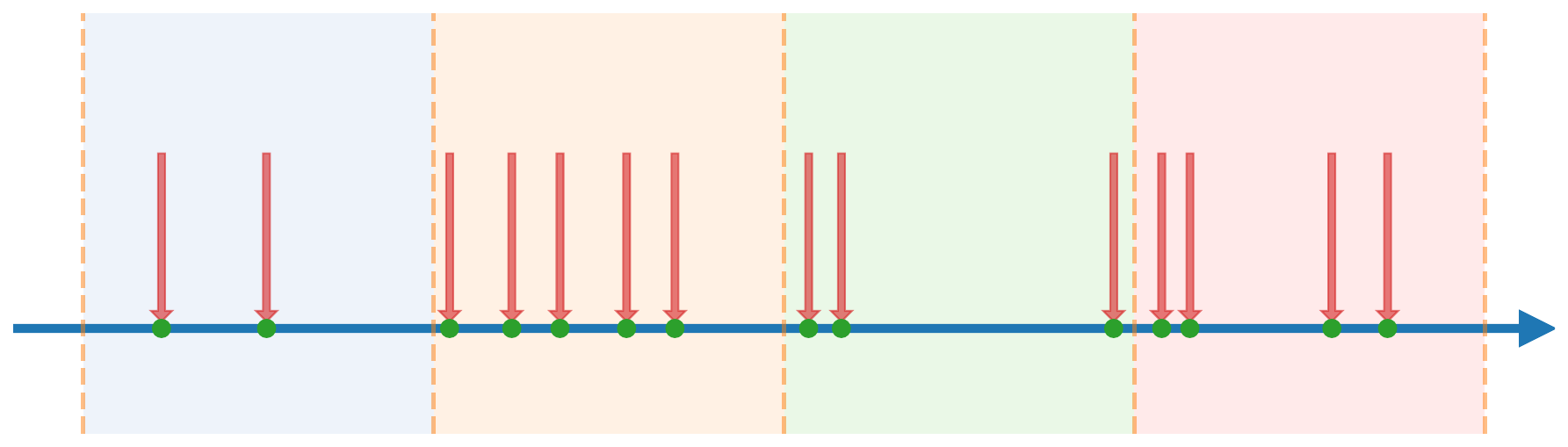}
        \put(35,23){\vector(-1,0){7}}
        \put(43,23){\vector(1,0){7}}
        \put(36,22){$1/f$}
        \put(61,21){$e_i$}
        \put(60.5,20){\vector(-1,-1){6}}
        \put(28,7){$\underbrace{\hspace{25pt}}_{n}$} 
        \put(100,5.5){$t$}
    \end{overpic}
    \caption{Event spherical frame formation process: Each downward arrow represents a triggered event $e_i$. Colored backgrounds indicate time segments of duration $1/f$. The brace underneath shows the selection of the first $n$ events within each time segment to form an event spherical frame.}
    \label{fig:frame_formation}
\end{figure}
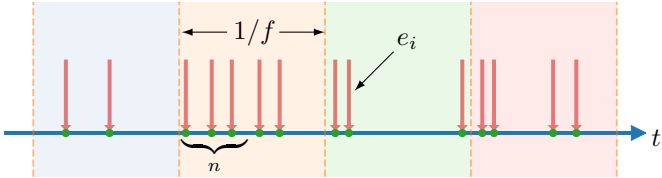

%% file: sections/methodology_04_event_spherical_formation.tex
\subsection{Event Spherical Frame Formation}
\label{subsec:event_spherical_frame_formation}

The continuous nature of event streams reflects the ongoing motion of the event camera. While contrast maximization methods typically process events over relatively long time windows (e.g., \SI{0.1}{s}) under the assumption of constant angular velocity, we propose a different approach that operates on much shorter time scales to better handle varying motion patterns. We segment the event stream at a frequency $f$ and select the first $n$ events from each segment, as illustrated in \prettyref{fig:frame_formation}. These events are then transformed into spherical coordinates as described in \prettyref{subsec:event_spherical_representation} to form event spherical frames.

To illustrate why our frame selection strategy effectively minimizes intra-frame camera motion, let us analyze a specific example from our implementation on ECRot \cite{cmax-slam} dataset. For ECRot sequences, we set $f = \SI{1000}{Hz}$ and $n = 1500$. Our analysis shows that these $n = 1500$ events span an average duration of only \SI{0.121}{ms}. Taking the average angular velocity of $\omega = \SI{120}{\degree/\second}$ as an example, this brief time span results in an intra-frame rotation of \SI{0.0145}{\degree}, leading to very small motion-induced pixel displacement within each frame. Each frame containing only 1500 events is too sparse to reveal the scene structure, making it challenging for contrast maximization methods to process effectively. However, our ES-ICP algorithm is specifically designed to handle such sparse point clouds effectively. This configuration allows our subsequent ES-ICP method to estimate $SO(3)$ pose at $\SI{1000}{Hz}$.

Although the intra-frame motion is already very small, we further apply motion compensation to the events within each frame to achieve the highest possible accuracy. For each frame, we estimate a constant angular velocity $\boldsymbol{\omega}$ based on the rotational change between the two most recent pose estimates. Given this short time span (\SI{0.121}{ms} on average), the constant angular velocity assumption is much more reasonable compared to applying it over longer windows. Using this angular velocity, we then warp all events within the frame to the timestamp of the first event in the frame. Specifically, for an event $e_i$ with timestamp $t_i$ and its spherical representation $\mathbf{p}^s_i$, we apply the following transformation:
\begin{equation}
    \mathbf{p}^s_i \leftarrow \exp((t_0 - t_i)\boldsymbol{\omega}^\wedge)\mathbf{p}^s_i
\end{equation}
where $t_0$ is the timestamp of the first event in the frame, $t_i$ is the timestamp of the $i$-th event, $\boldsymbol{\omega}^\wedge$ denotes the skew-symmetric matrix of $\boldsymbol{\omega}$, and $\exp(\cdot^\wedge)$ is the exponential map from $\mathfrak{so}(3)$ to $SO(3)$. After processing all $n$ events, we obtain the motion-compensated event spherical frame $\mathcal{E}^s = \{\mathbf{p}^s_k\}_{i=0}^{n-1}$.

%% file: figs_and_tabs/fig_es_icp.tex
\definecolor{c_color}{RGB}{44,160,44}
\begin{figure}[!t]
    \centering
    \begin{overpic}[width=0.88\columnwidth,
        trim={15mm} {45mm} {15mm} {35mm},
        clip]{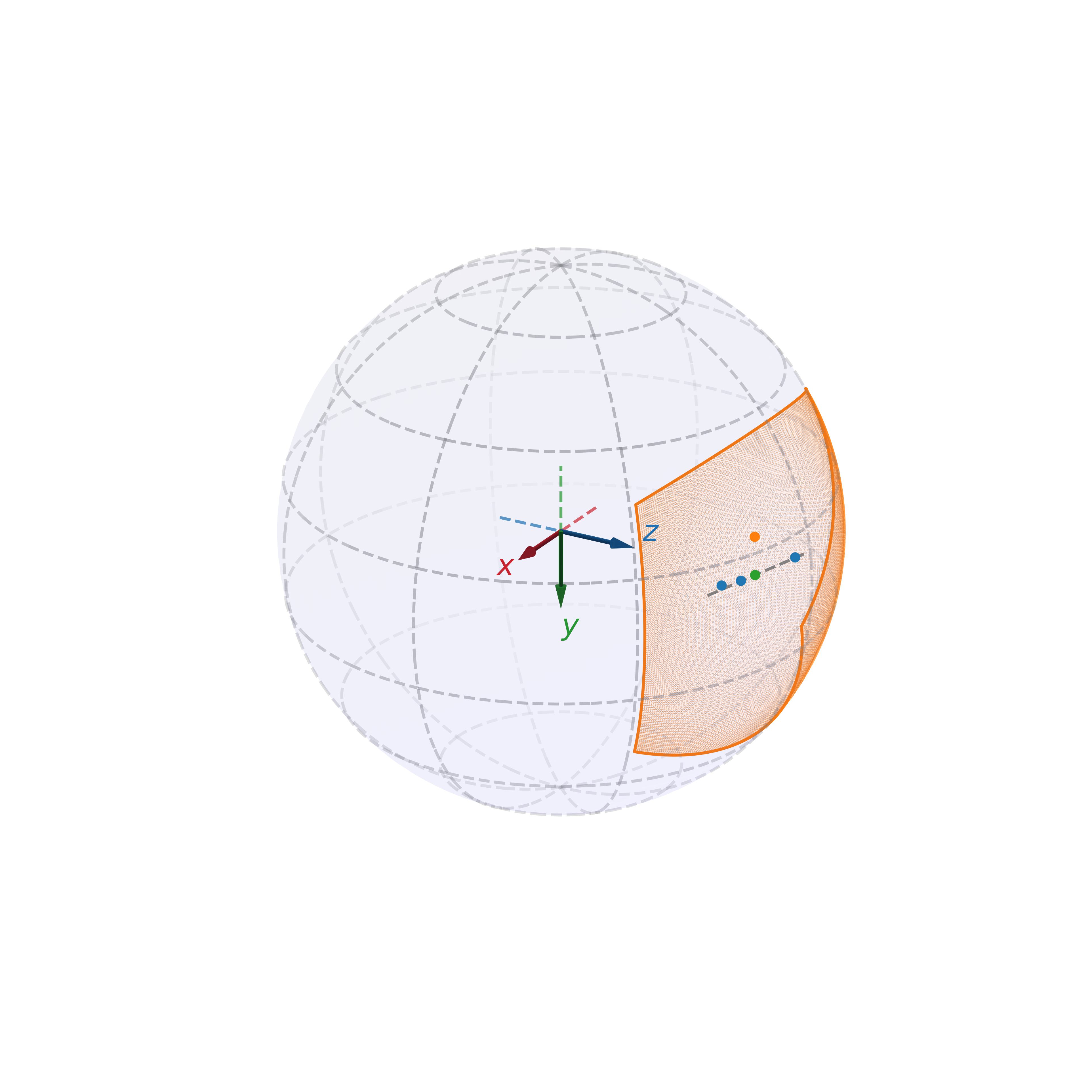}
        \put(72,35){\color{spherical_point_projection}$\mathbf{p}^s_i$}
        \put(66,24){\color{original_point}$\mathbf{q}^s_j$}
        \put(72,25){\color{c_color}$\mathbf{c}$}
    \end{overpic}
    \caption{ES-ICP error visualization. The shaded region represents the spherical event map $\color{spherical_point_projection}\mathcal{M}^s$. Our objective is to minimize the distances between all \textcolor{spherical_point_projection}{$\mathbf{p}^s_i$} in the current event frame and the line fitted to their $k$ nearest neighbors \textcolor{original_point}{$\mathbf{q}^s_j$} in the map. The point \textcolor{c_color}{$\mathbf{c}$} represents the centroid of the $k$ nearest neighbors used for line fitting.}
    \label{fig:spherical_icp_error}
\end{figure}

%% file: sections/methodology_05_es_icp.tex
\subsection{Event Spherical Iterative Closest Point (ES-ICP) Algorithm}
\label{subsec:es_icp}

The core of our approach lies in the Event Spherical Iterative Closest Point (ES-ICP) algorithm, which is designed to efficiently estimate rotational motion using spherical event representations. As illustrated in \prettyref{fig:spherical_icp}, our goal is to align the sparse event point cloud with a denser spherical event map on the surface of a unit sphere.

Since events are primarily triggered by edges in the environment, and these edges typically form continuous structures on the spherical surface, we optimize the distance between points and nearby line segments rather than point-to-point distances for improved robustness and accuracy. This approach better captures the underlying edge structure of the scene while being more resilient to noise and sparse event distributions. As visualized in \prettyref{fig:spherical_icp_error}, both the map and the frame exist on a curved surface (the unit sphere), where these line segments locally approximate the edge structures projected onto the sphere.

Given a set of event projections on a unit sphere $\mathcal{E}^s = \{\mathbf{p}^s_i\}_{i=0}^{n-1}$, representing a sparse event spherical frame of $n$ points, and a spherical event map $\mathcal{M}^s = \{\mathbf{q}^s_j\}_{j=0}^{m-1}$, which is an accumulated representation of previously aligned event frames, our objective is to optimize the pose $\mathbf{R}$ in $SO(3)$ to minimize the distance between points in $\mathcal{E}^s$ and $\mathcal{M}^s$. This can be formulated as:
\begin{equation}
\mathbf{R}^* = \argmin\nolimits_{\mathbf{R} \in SO(3)} \sum\nolimits_{i=0}^{n-1} d(\mathbf{R}\mathbf{p}^s_i, \mathcal{M}^s)
\label{eq:es_icp_objective}
\end{equation}
where $d(\cdot, \cdot)$ represents a distance metric and $\mathbf{R}^*$ is the optimal rotation we seek to estimate.

To implement this approach, for each $\mathbf{p}^s_i$, we first apply the current estimate of the rotation matrix $\mathbf{R} \in SO(3)$:
\begin{equation}
\tilde{\mathbf{p}}^s_i = \mathbf{R} \mathbf{p}^s_i
\end{equation}

We then employ a k-d tree structure to efficiently find the $k=5$ nearest neighbors of $\tilde{\mathbf{p}}^s_i$ in $\mathcal{M}^s$. Using these neighbors, we fit a line $l: \mathbf{d}\tau + \mathbf{c}$, where $\mathbf{d}$ is the direction vector and $\mathbf{c}$ is the centroid of the $k$ nearest neighboring points. \begin{revisedtextblock}We choose $k=5$ as it provides sufficient points for robust line fitting while maintaining computational efficiency and focusing on local geometric structure.\end{revisedtextblock} It's worth noting that while \prettyref{fig:spherical_icp_error} illustrates this concept with exaggerated distances for clarity, in practice, the nearest neighbors are typically very close to each other on the unit sphere. As a result, the fitted line closely approximates the tangent to the unit sphere at that point.

The optimization problem can then be reformulated as:
\begin{equation}
\mathbf{R}^* = \argmin\nolimits_{\mathbf{R} \in SO(3)} \sum\nolimits_{i=0}^{n-1} \|\mathbf{d} \times (\mathbf{R}\mathbf{p}^s_i - \mathbf{c})\|^2_2
\label{eq:es_icp_point_to_line}
\end{equation}

We solve this non-linear least squares optimization problem using the Gauss-Newton method. For each point, we derive the Jacobian matrix $\mathbf{J}_i \in \mathbb{R}^{3\times3}$ with respect to the Lie algebra of $SO(3)$:
\begin{equation}
\mathbf{J}_i = -\mathbf{d}^\wedge \mathbf{R} {\mathbf{p}^s_i}^\wedge
\end{equation}
where $\mathbf{d}^\wedge$ and ${\mathbf{p}^s_i}^\wedge$ denote the skew-symmetric matrices of $\mathbf{d}$ and $\mathbf{p}^s_i$, respectively.

In each Gauss-Newton iteration, we construct and solve the normal equation:
\begin{equation}
\mathbf{H}\Delta\mathbf{x} = \mathbf{g}
\label{eq:gauss-newton}
\end{equation}
Here, $\mathbf{H} \in \mathbb{R}^{3\times3}$ is the approximated Hessian matrix, $\Delta\mathbf{x} \in \mathbb{R}^3$ is the incremental update we seek in the Lie algebra $\mathfrak{so}(3)$, and $\mathbf{g} \in \mathbb{R}^3$ is the gradient vector. These are computed as:
\begin{equation}
\mathbf{H} = \sum\nolimits_{i=0}^{n-1} \mathbf{J}_i^\top \mathbf{J}_i, \quad \mathbf{g} = -\sum\nolimits_{i=0}^{n-1} \mathbf{J}_i^\top [\mathbf{d} \times (\mathbf{R}\mathbf{p}^s_i - \mathbf{c})]
\end{equation}

After solving \prettyref{eq:gauss-newton} to obtain the update $\Delta\mathbf{x}$ in the Lie algebra $\mathfrak{so}(3)$, we update the current pose estimate $\mathbf{R}_{it}$ using the exponential map:
\begin{equation}
\mathbf{R}_{it+1} = \mathbf{R}_{it} \exp(\Delta\mathbf{x}^\wedge)
\end{equation}
where $\exp(\cdot^\wedge)$ maps from $\mathfrak{so}(3)$ to $SO(3)$, and $it$ denotes the iteration number.

The optimization is considered converged when the magnitude of the update $\|\Delta\mathbf{x}\|$ falls below a predefined threshold or reaches a maximum number of iterations, yielding an optimal rotation estimate that precisely aligns the event frame with the spherical map.
\begin{revisedtextblock}
A crucial aspect of our ES-ICP implementation is its high-frequency operation at \SI{1000}{Hz}. \begin{revisedtextblockCR}This design choice helps mitigate a fundamental challenge inherent to ICP-based algorithms: sensitivity to initialization and susceptibility to local minima.\end{revisedtextblockCR} By operating at such high frequencies, the actual rotation between consecutive frames becomes very small, ensuring that the previous ES-ICP solution provides an excellent initial guess for the current frame. \begin{revisedtextblockCR}While ICP remains a local optimizer, this strategy significantly improves the likelihood of ES-ICP converging to the correct solution\end{revisedtextblockCR}, which is a core reason for EROAM's superior accuracy compared to methods that operate over longer time windows.
\end{revisedtextblock}

%% file: sections/methodology_06_map.tex
\subsection{Event Spherical Map Maintenance and Update}
\label{subsec:map_maintenance}
\begin{revisedtextblock}
\input{figs_and_tabs/fig_regional_density_management}
\end{revisedtextblock}
To maintain an accurate and efficient representation of the environment while ensuring predictable computational costs for real-time operation, we employ an intelligent map maintenance approach that combines incremental updates with regional density control. We selectively update our spherical event map based on significant camera motion, designating certain frames as key frames based on a rotation threshold criterion.

Let $\mathbf{R}_c$ be the current frame's estimated rotation and $\mathbf{R}_k$ be the rotation of the last key frame. We consider the current frame as a new key frame if:
\begin{equation}
\|\log(\mathbf{R}_c \mathbf{R}_k^{-1})^\vee\|_2 > \theta_t
\label{eq:keyframe_criterion}
\end{equation}
where $\log(\cdot)^\vee$ is the logarithm map from $SO(3)$ to $\mathfrak{so}(3)$, and $\theta_t$ is a predefined rotation threshold.
\begin{revisedtextblock}
Upon identifying a new key frame, we update our map using an incremental k-d tree (ikd-Tree) data structure \cite{cai2021ikd} and our novel Regional Density Management (RDM) approach. Unlike traditional static k-d trees that require complete rebuilding after updates, the ikd-Tree supports efficient incremental insertions and deletions, allowing us to dynamically update the spherical map without the computational overhead of rebuilding the entire tree structure. This design choice is fundamental to our real-time performance guarantee, as it ensures that map update operations maintain bounded and predictable execution times regardless of map size or update frequency.

A significant challenge in event-based mapping arises when the camera repeatedly observes the same area, generating redundant points that can degrade performance and increase memory consumption. Our RDM approach addresses this by intelligently controlling point density across the spherical surface while maintaining stable computational complexity. As illustrated in \prettyref{fig:regional_density_management}, RDM partitions the unit sphere into grid cells based on latitude-longitude lines, each with a maximum point capacity. Since grid cells near the poles have smaller surface areas than those at the equator, their maximum point capacity is proportionally scaled according to their surface area. When a new keyframe is added to the spherical map, events are only inserted into regions that haven't reached their capacity limits. The pink grid cells in the figure have already reached their capacity, so any new points from the keyframe that fall within these saturated regions are not added to the map.

This density management mechanism maintains a uniform spatial density across the map, preventing over-representation of frequently observed regions while preserving structural details and feature richness. Crucially, RDM ensures that computational costs remain bounded and predictable throughout extended operation, as the maximum number of points in the map is inherently limited by the grid structure. The combination of ikd-Tree and RDM creates a spherical map that dynamically evolves as the camera explores the environment, efficiently balancing computational resources with geometric integrity while guaranteeing the stable execution times essential for real-time operation.
\end{revisedtextblock}

%% file: figs_and_tabs/fig_regional_density_management.tex
\begin{figure}[!t]
    \centering
    \begin{overpic}[width=0.90\columnwidth,
        trim={25mm} {65mm} {25mm} {65mm},
        clip]{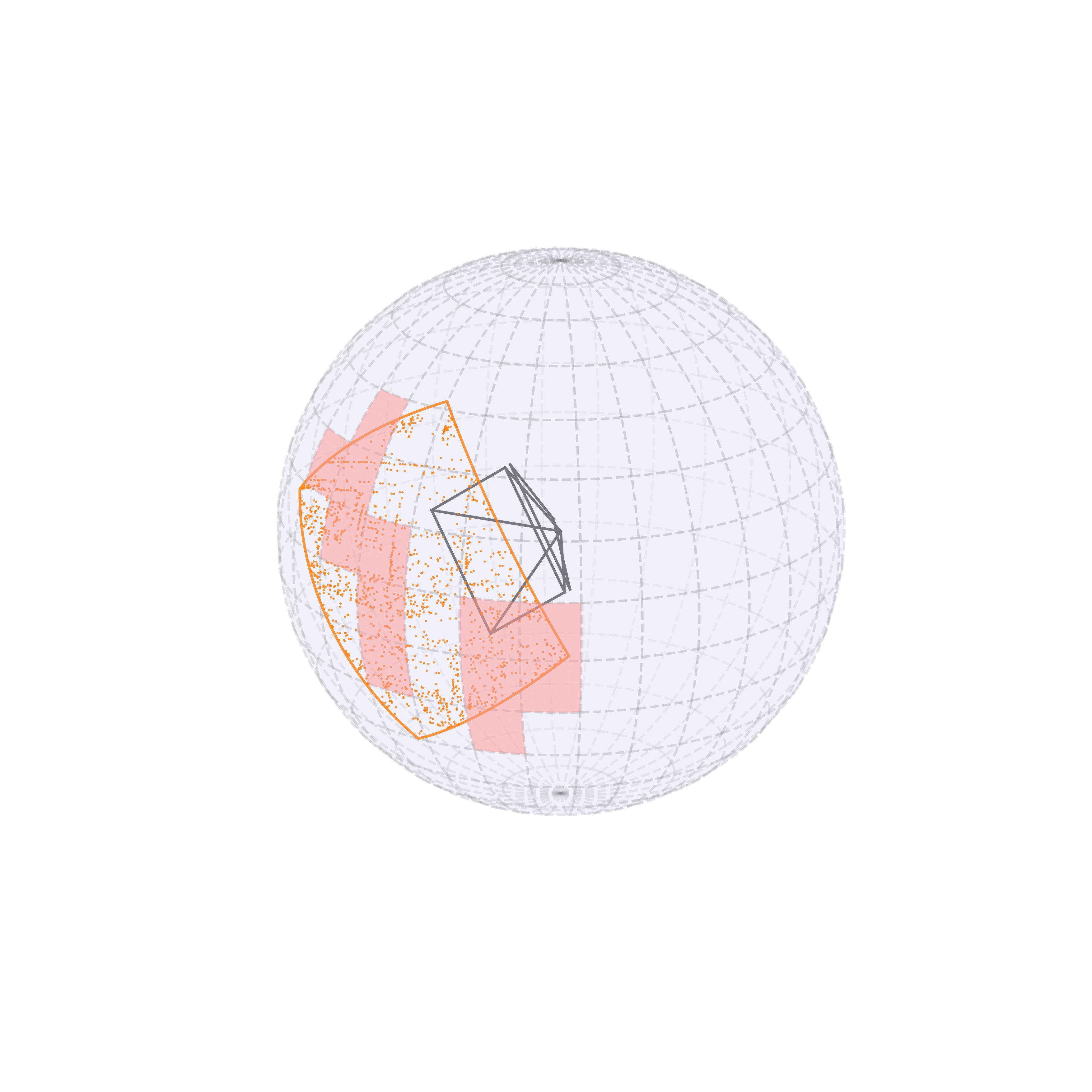}
    \end{overpic}
    \caption{Illustration of Regional Density Management (RDM) for spherical event representation. The unit sphere is divided into grid cells based on latitude-longitude lines, each with a maximum point capacity. Since grid cells near the poles have smaller surface areas than those at the equator, their maximum point capacity is proportionally scaled according to their surface area. When a new keyframe is added to the spherical map, events are only inserted into regions that haven't reached their capacity limits. As shown in the figure, the pink grid cells have already reached their capacity, so any new points from the keyframe that fall within these saturated regions are not added to the map.}
    \label{fig:regional_density_management}
\end{figure}

%% file: figs_and_tabs/fig_spherical_to_panoramic.tex
\begin{figure}[!t]
    \vspace{-3mm}
    \centering
    \begin{overpic}[width=0.98\columnwidth,
        trim={15mm} {45mm} {15mm} {35mm},
        clip]{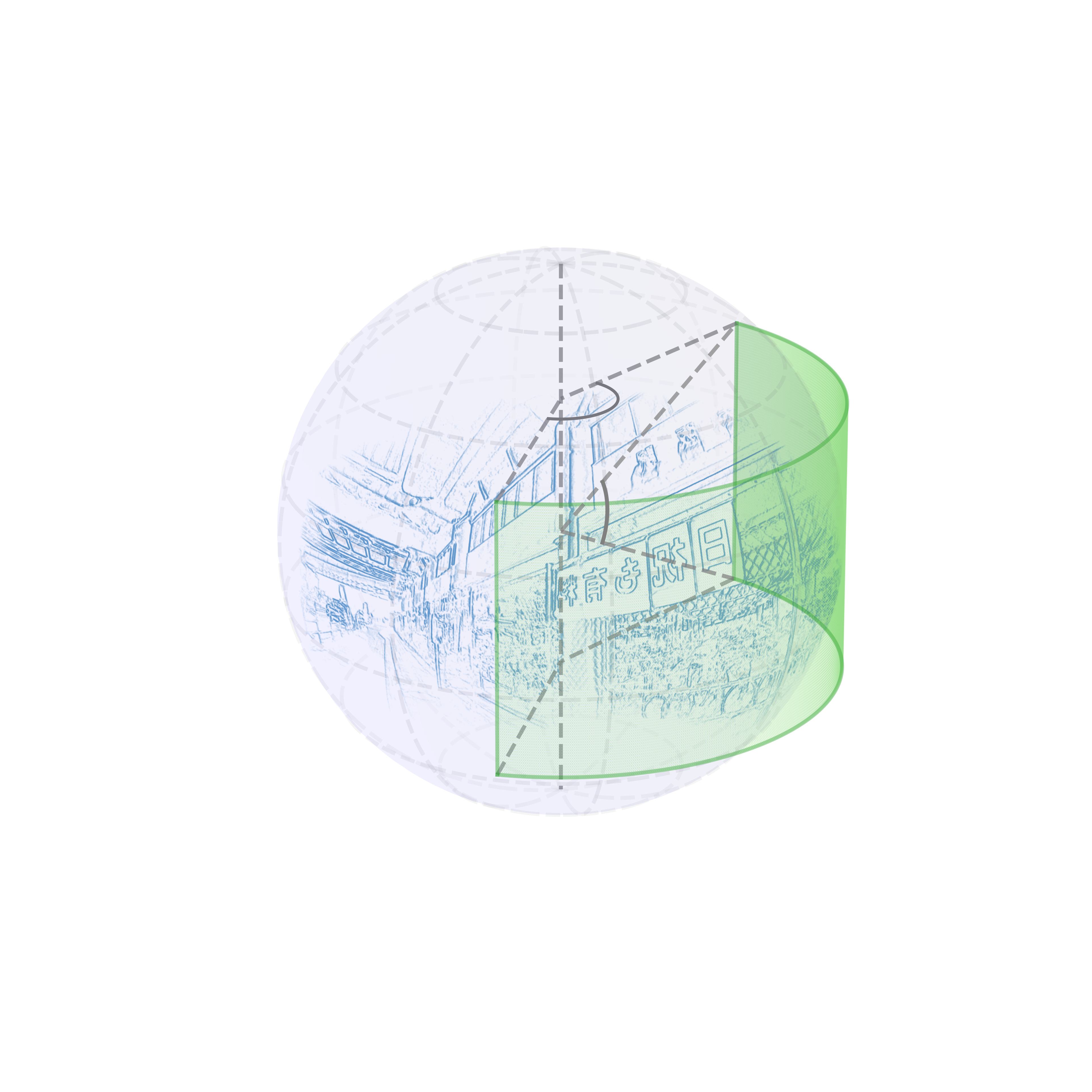}
        \put(58,34){$\phi_v$}
        \put(45,45){$\phi_h$}
    \end{overpic}
    \caption{Projection from spherical event map to panoramic image. Events on the unit sphere are projected onto a surrounding cylinder, then unwrapped to form the panorama. $\phi_v$: vertical field of view, $\phi_h$: horizontal span angle. The cylinder is discretized based on desired pixel density to generate the final panoramic image.}
    \label{fig:spherical_to_panoramic}
\end{figure}

%% file: sections/methodology_07_panoramic.tex
\subsection{Panoramic Image Generation}
\label{subsec:panoramic_generation}

A valuable byproduct of our system is the generation of panoramic images. \begin{revisedtextblock}To generate these panoramic representations, we utilize all aligned event spherical frames, which are accumulated in a dedicated buffer throughout the system's operation.\end{revisedtextblock} The conversion process involves projecting the events from the unit sphere onto a surrounding cylinder, which is then unwrapped to form a 2D panoramic representation, as illustrated in \prettyref{fig:spherical_to_panoramic}. The projection process is defined by two key parameters: $\phi_v$, the vertical field of view, which determines the height of the cylindrical projection surface, and $\phi_h$, the horizontal span angle, which defines the width of the panoramic image. \prettyref{fig:panoramic_head} shows an example of the resulting panoramic image generated by our method.

The conversion from spherical coordinates to panoramic image coordinates involves projecting each spherical point \begin{revisedtextblock}$\mathbf{p}_i^s = (x, y, z)^T$\end{revisedtextblock} onto the cylinder surface using the equation:
\begin{revisedtextblock}
\begin{equation}
\mathbf{p}_i^c = (\frac{x}{\sqrt{x^2 + y^2}}, \frac{y}{\sqrt{x^2 + y^2}}, \frac{z}{\sqrt{x^2 + y^2}})^T
\end{equation}
\end{revisedtextblock}
This projection normalizes the x and y coordinates to the unit circle while preserving the relative height of z. The cylinder surface is then discretized based on the desired pixel density of the panoramic image, with the azimuthal angle of the projected point determining its horizontal position and its height on the cylinder determining its vertical position in the image.

A key advantage of our approach is that the entire mapping process operates in continuous spherical space, maintaining complete independence from any discretization requirements. Unlike traditional methods that require predefined resolutions, our method allows for the generation of panoramic images at arbitrary resolutions as needed, with the panorama generation process entirely decoupled from the core tracking and mapping pipeline. This design choice offers several benefits:
\begin{itemize}
    \item The spherical mapping and pose estimation operate in continuous space, ensuring their accuracy is independent of any discretization choices.
    \item Multiple panoramic images can be generated at different resolutions without modifying the underlying map data.
    \item The panorama generation serves as an independent post-processing step, preserving the real-time performance of the core system.
\end{itemize}

%% file: sections/experiments.tex
\section{Experiments}

In this section, we present a comprehensive evaluation of our proposed event-based rotational motion estimation method. We begin by detailing our experimental setup, including evaluation metrics, datasets, and hardware configuration (\prettyref{subsec:experimental_setup}). We then evaluate our method on both synthetic and real-world data: first on the ECRot dataset (\prettyref{subsec:ecrot_experiments}), followed by extended simulations using ESIM (\prettyref{subsec:extended_esim}) to assess performance under various challenging conditions, and then on real-world sequences (\prettyref{subsec:real_experiments}). \begin{revisedtextblockCR}We provide runtime analysis (\prettyref{subsec:runtime_analysis}), evaluate robustness to unmodeled translation (\prettyref{subsec:ecd_robustness}), and conduct ablation studies (\prettyref{subsec:ablation_studies}).\end{revisedtextblockCR}

\subsection{Experimental Setup}
\label{subsec:experimental_setup}

\subsubsection{Evaluation Metrics}
\label{subsubsec:evaluation_metrics}

To rigorously assess the performance of our proposed method, we employ two primary metrics for evaluating rotational motion estimation accuracy: Absolute Pose Error (APE) \cite{sturm2012benchmark} and Relative Pose Error (RPE) \cite{sturm2012benchmark}. These metrics provide a comprehensive evaluation of both global consistency and local accuracy of the estimated camera rotations.

\begin{revisedtextblock}
\paragraph{Absolute Pose Error (APE)} The APE measures global pose consistency. For timestamp $t_k$, the absolute rotation error is:\end{revisedtextblock}
\begin{equation}
\boldsymbol{\epsilon}_k = \log(\mathbf{R}'_k{}^\top \mathbf{R}_k)^\vee
\label{eq:ape}
\end{equation}
\begin{revisedtextblock}
where $\mathbf{R}_k$ and $\mathbf{R}'_k$ are estimated and ground truth rotations, and $\|\boldsymbol{\epsilon}_k\|_2$ represents the angular error.
\end{revisedtextblock}

To obtain a single representative value, we calculate the mean of these angular errors over all timestamps:
\begin{equation}
\overline{ape} = \frac{1}{N}\sum\nolimits_{k=1}^N \|\boldsymbol{\epsilon}_k\|_2
\label{eq:ape_mean}
\end{equation}
where $N$ is the total number of timestamps in the evaluated trajectory. This mean value $\overline{ape}$ provides an average measure of the angular error across the entire trajectory.

\paragraph{Relative Pose Error (RPE)} The RPE measures the local accuracy of the pose estimates by comparing the relative rotation between two timestamps in the estimated trajectory to the corresponding relative rotation in the ground truth trajectory. For a pair of timestamps $t_k$ and $t_{k+\Delta}$, the relative rotation error $\boldsymbol{\delta}_k$ is calculated as:
\begin{equation}
\boldsymbol{\delta}_k = \log\left(\left(\mathbf{R}'_k{}^\top \mathbf{R}'_{k+\Delta}\right)^{-1} \left(\mathbf{R}_k{}^\top \mathbf{R}_{k+\Delta}\right)\right)^\vee
\label{eq:rpe}
\end{equation}
where $\{\mathbf{R}_k, \mathbf{R}_{k+\Delta}\}$ and $\{\mathbf{R}'_k, \mathbf{R}'_{k+\Delta}\}$ are pairs of estimated and ground truth poses, respectively. Similar to the APE, the magnitude of $\boldsymbol{\delta}_k$, denoted as $\|\boldsymbol{\delta}_k\|_2$, represents the angular difference between the estimated and ground truth relative rotations.

The mean of these relative rotation errors $\overline{rpe}$ is then computed as:
\begin{equation}
\overline{rpe} = \frac{1}{M}\sum\nolimits_{k=1}^M \|\boldsymbol{\delta}_k\|_2
\label{eq:rpe_mean}
\end{equation}
where $M$ is the total number of relative pose pairs evaluated along the entire trajectory. To ensure fair comparison across methods with different estimation frequencies, we compute the RPE at fixed intervals corresponding to \SI{10}{\degree} of rotation in the ground truth trajectory. An $\overline{rpe}$ value of \SI{0.5}{\degree} indicates that estimated relative rotations over \SI{10}{\degree} intervals differ from ground truth by \SI{0.5}{\degree} on average, providing consistent local accuracy measurement across methods.

For the evaluation of APE and RPE across all experiments, we utilize the EVO toolkit \cite{grupp2017evo}, a software package designed for the evaluation of odometry and SLAM algorithms, ensuring consistent and reproducible assessments. To facilitate intuitive understanding of the error magnitudes, we convert all radian measurements to degrees in our reported results.

\subsubsection{Datasets}
\label{subsubsec:datasets}

To comprehensively evaluate our method's accuracy and robustness in rotation estimation, we utilize \begin{revisedtextblockCR}four\end{revisedtextblockCR} types of data sources: the ECRot synthetic dataset \cite{cmax-slam}, extended simulations generated using the ESIM simulator \cite{rebecq2018esim}, \begin{revisedtextblockCR}the Event Camera Dataset (ECD) \cite{mueggler2017event},\end{revisedtextblockCR} and our newly collected real-world sequences with LiDAR-based ground truth. \begin{revisedtextblock}Further details on their motion patterns, including event rates and average angular velocities ($\omega_{avg}$), are provided in \prettyref{tab:runtime_comparison} and \prettyref{tab:runtime_comparison_real}.\end{revisedtextblock}

\paragraph{ECRot Synthetic Dataset} We utilize all six synthetic sequences from the ECRot dataset: \texttt{bay}, \texttt{bicycle}, \texttt{city}, \texttt{street}, \texttt{town} and \texttt{playroom}. The first five sequences feature a resolution of $240\times180$ with a duration of \SI{5}{s}, while the \texttt{playroom} sequence has a resolution of $128\times128$ and spans \SI{2.5}{s}. Each sequence provides events and ground truth poses for quantitative evaluation.

\paragraph{Extended ESIM Simulations} Given the relatively short duration and moderate motion patterns in the ECRot dataset, we generate additional sequences using ESIM to evaluate our method under more challenging conditions, including various sequence durations (\SIrange{10}{80}{s}) and different angular velocities (\SIrange{49.83}{393.21}{\degree/\second}).

\paragraph{Real-World Sequences} We collected 16 real-world sequences using a synchronized setup consisting of an iniVation DVXplorer event camera ($640\times480$ resolution) and a Livox Avia LiDAR. The two sensors are hardware-synchronized, and their extrinsic parameters are calibrated using the edge-based calibration method proposed in \cite{xing2023target}. \begin{revisedtextblock}Among these sequences, the \texttt{limx-rotation} and \texttt{limx-sway} sequences were specifically collected using our LimX TRON 1 wheeled bipedal robot equipped with an iniVation DVXplorer event camera, as shown in \prettyref{fig:limx_tron1_setup}. Ground truth camera trajectories are derived by combining high-precision LiDAR poses from Point-LIO \cite{he2023point} with the calibrated LiDAR-event camera extrinsic parameters, providing accurate benchmarks for evaluating our rotation estimation results.\end{revisedtextblock}

\begin{revisedtextblockCR}
\paragraph{Event Camera Dataset (ECD)} To evaluate robustness to unmodeled translational motion, we use the ECD \cite{mueggler2017event}, which features unconstrained handheld 6-DOF motion with translation typically within \SI{\sim0.3}{\m}. Following CMax-SLAM's protocol \cite{cmax-slam}, we process four sequences (\texttt{boxes}, \texttt{dynamic}, \texttt{poster}, \texttt{shapes}), using the first 30 seconds of each.
\end{revisedtextblockCR}

\subsubsection{Hardware Configuration}
\label{subsubsec:hardware_config}

To ensure consistency and fair comparison across all experiments, we conducted our evaluations on a single laptop computer. The hardware configuration consists of an Intel Core i9-14900HX CPU, 32GB of RAM, and an NVIDIA RTX 4060 GPU.

\subsection{Experiments on ECRot Dataset}
\label{subsec:ecrot_experiments}

\begin{revisedtextblock}
\input{figs_and_tabs/fig_ecrot_synthetic_panoramic}
\input{figs_and_tabs/tab_ecrot_synthetic}
\input{figs_and_tabs/fig_roll_pitch_yaw_town}
\end{revisedtextblock}

We conducted experiments using all six synthetic sequences from the ECRot dataset \cite{cmax-slam}, comparing our method with several state-of-the-art approaches. \prettyref{tab:ecrot_synthetic_results} presents the quantitative results of our method compared to these approaches including SMT \cite{smt}, RTPT \cite{rtpt}, CM-GAE \cite{cm-gae}, and CMax-SLAM \cite{cmax-slam}.

Among all methods tested, none of them except CMax-SLAM and our EROAM demonstrated consistent performance. SMT, despite utilizing ground truth initialization to construct its initial map, fails to maintain stable pose estimation and quickly deviates from the initialized state, resulting in complete failure on all sequences. Moreover, SMT exhibits the highest computational overhead in our experiments. All other evaluated methods, including RTPT, CM-GAE, CMax-SLAM, and our EROAM, operate without requiring any prior pose information.

RTPT demonstrates better performance than SMT, maintaining tracking during initial camera rotations. However, it ultimately fails to process complete sequences despite multiple parameter tuning attempts, confirming the limitations reported in \cite{cmax-slam}. This failure occurs when the camera's field of view (FOV) approaches the map boundaries - RTPT's projection mechanism cannot effectively update the map in these situations, and attempts to increase the map size result in insufficient gradient information due to sparse projections. 

CM-GAE showed inconsistent performance - we report its best results after multiple parameter tuning attempts, with successful processing of only the bicycle, town, and playroom sequences, though its performance remains unstable. These three methods demonstrate limited general applicability. 

In contrast, both CMax-SLAM and our method successfully processed all six sequences. CMax-SLAM achieved this using their recommended sequence-specific parameters, while our method produced superior results using consistent parameters across all sequences, as evidenced by the lower $\overline{ape}$ and $\overline{rpe}$ values. We provide more detailed comparisons between CMax-SLAM and our method in \prettyref{subsec:extended_esim} and \prettyref{subsec:runtime_analysis}, where we evaluate performance under more challenging conditions and analyze computational efficiency.

To visualize the detailed performance differences, we present the rotation estimation results on the town sequence in \prettyref{fig:rotation_comparison}. The figure shows both complete trajectories and zoomed-in views of roll, pitch, and yaw angles. SMT and RTPT quickly deviate from the ground truth and fail. CM-GAE maintains tracking but exhibits noticeable drift across all angles. In contrast, our method closely tracks the ground truth trajectory across all three rotation angles, with CMax-SLAM showing similar but slightly less accurate performance, particularly visible in the detailed views.

For qualitative evaluation, we focus on panoramic mapping results from CM-GAE, CMax-SLAM, and our method, as shown in \prettyref{fig:ecrot_synthetic_panoramic_comparison}. While CMax-SLAM's original implementation generates panoramas at a fixed resolution of $1024\times512$ (shown in \prettyref{fig:panoramic_town_cmaxslam_origin}, with empty regions trimmed), our mapping framework operates in continuous spherical space rather than discrete panoramic space (detailed in \prettyref{subsec:panoramic_generation}). This continuous representation enables panorama generation at arbitrary resolutions - for this comparison, we chose a resolution of $7617\times2000$ to demonstrate the capability of our method.

To ensure fair visual comparison, we implement a unified panorama generation framework using each method's estimated trajectories. \prettyref{fig:panoramic_town_cmgae}, \prettyref{fig:panoramic_town_cmaxslam}, and \prettyref{fig:panoramic_town_eroam} show panoramas generated using identical parameters: an event window size of \SI{0.2}{ms} and consistent colorization scheme where the 90th percentile of event counts maps to intensity 255. All three methods' panoramas are regenerated using their respective estimated trajectories through this unified pipeline.

As evident from the zoomed-in regions in \prettyref{fig:ecrot_synthetic_panoramic_comparison}, our method produces panoramic images with significantly sharper edges and clearer structural details compared to other approaches. The improved quality is particularly noticeable in architectural features and text regions, demonstrating the benefits of our accurate rotation estimation.

\input{figs_and_tabs/fig_extended_esim_ape_rpe}
\begin{revisedtextblock}
\input{figs_and_tabs/tab_extended_esim_ape_rpe_omega}
\input{figs_and_tabs/tab_extended_esim_ape_rpe_duration}
\input{figs_and_tabs/fig_roll_pitch_yaw_lt_80}
\input{figs_and_tabs/fig_extended_esim_dm4}
\end{revisedtextblock}

\subsection{Extended Simulation Analysis}
\label{subsec:extended_esim}

While both CMax-SLAM and EROAM successfully process all ECRot sequences, these sequences are short (5s) and exhibit moderate motion patterns, as shown in \prettyref{fig:rotation_comparison}. For reference, the ECRot \texttt{bicycle} sequence has an average angular velocity of \SI{84.208}{\degree/\second} and acceleration of \SI{101.73}{\degree/\second\squared}. To evaluate the generality and robustness of our method, we conducted extended simulation experiments using ESIM \cite{rebecq2018esim} with the \texttt{bicycle} scene from ECRot dataset under more challenging conditions.

As shown in \prettyref{fig:angular_motion}, we designed two sets of sequences to test different aspects: (1) sequences with increasing angular velocities and accelerations (\texttt{DM-1}--\texttt{DM-8}), and (2) sequences of increasing duration (\texttt{LD-10}--\texttt{LD-80}). For the dynamic motion sequences, we progressively increased the magnitude and frequency of rotational motion while maintaining the 5s duration, with angular velocity ranging from \SI{49.83}{\degree/\second} to \SI{393.21}{\degree/\second} and angular acceleration from \SI{70.41}{\degree/\second\squared} to \SI{1121.32}{\degree/\second\squared}. For the duration test sequences, we used a moderate dynamic profile (average angular velocity: \SI{107.24}{\degree/\second}, average angular acceleration: \SI{200.93}{\degree/\second\squared}) and varied the sequence length from 10s to 80s.

\prettyref{tab:omega_comparison} and \prettyref{fig:omega_comparison} present the performance comparison under different angular velocities. While both methods perform well under moderate motion (\texttt{DM-1} and \texttt{DM-2}, $\overline{ape} \leq \SI{1.032}{\degree}$), CMax-SLAM's performance degrades significantly when angular velocity increases (\texttt{DM-4} onwards), with $\overline{ape}$ exceeding \SI{100}{\degree} in high-velocity scenarios. This degradation stems from a fundamental limitation of the contrast maximization framework: it assumes constant angular velocity within each time window used for estimation. Under high-dynamic motions with large angular accelerations, this assumption becomes invalid, leading to significant estimation errors. Although CMax-SLAM employs backend optimization, it cannot recover from these fundamental front-end estimation errors. This is clearly demonstrated in \prettyref{fig:dm4_panoramic_comparison}, where CMax-SLAM's panoramic reconstruction shows severe ghosting artifacts and double edges due to incorrect motion estimation under high angular velocities.

The comparison across different sequence durations, shown in \prettyref{tab:duration_comparison} and \prettyref{fig:duration_comparison}, reveals similar patterns. CMax-SLAM exhibits increasing drift as sequence length grows, with $\overline{ape}$ rising from \SI{2.312}{\degree} in \texttt{LD-10} to \SI{11.677}{\degree} in \texttt{LD-80}. This drift accumulation is clearly visualized in \prettyref{fig:ld_80_rpy_comparison_with_zoom}, where CMax-SLAM's estimated trajectory shows significant deviation from ground truth, particularly after \SI{40}{s}. This further demonstrates how initial estimation errors in the contrast maximization framework propagate over time, despite backend optimization attempts. In contrast, our method maintains consistent accuracy across all sequences, as our ES-ICP approach directly optimizes geometric constraints on the unit sphere without relying on assumptions about constant angular velocity.

These results demonstrate that our method not only achieves higher accuracy but also maintains consistent performance under challenging conditions where CMax-SLAM struggles, suggesting superior robustness and general applicability.

\begin{revisedtextblock}
\input{figs_and_tabs/tab_university_data_ape_rpe}
\input{figs_and_tabs/fig_roll_pitch_yaw_window_building}
\input{figs_and_tabs/fig_eroam_campus_pano_comparison}
\begin{revisedtextblockCR}
\input{cr_figs_and_tabs/fig_eroam_campus_pano_more_visualization}
\end{revisedtextblockCR}
\end{revisedtextblock}

\subsection{Experiments on Real-World Data}
\label{subsec:real_experiments}

To validate our method's performance in real-world scenarios, we collected and evaluated EROAM on our EROAM-campus dataset. The dataset contains 16 sequences captured using a custom-built synchronized sensor suite, consisting of an iniVation DVXplorer event camera ($640\times480$ resolution) and a Livox Avia LiDAR mounted on a tripod. Both sensors were hardware-synchronized, and their extrinsic parameters were calibrated using the edge-based calibration method proposed in \cite{xing2023target, yuan2021pixel}. We manually performed arbitrary 3-DoF rotational motions while recording data, and obtained ground truth camera trajectories by combining high-precision LiDAR poses from Point-LIO \cite{he2023point} with the calibrated LiDAR-event camera extrinsic parameters.

The quantitative comparison results across all sixteen sequences are presented in \prettyref{tab:real_world_evaluation}. SMT fails to process any of the sequences, despite multiple parameter tuning attempts. RTPT shows very limited success, only managing to process the \texttt{embankment} sequence with an $\overline{ape}$ of \SI{0.654}{\degree}, but failing on all other sequences. CM-GAE and CMax-SLAM demonstrate significant limitations, with CM-GAE exhibiting large estimation errors ($\overline{ape} > \SI{20}{\degree}$) on 6 out of 16 sequences and CMax-SLAM failing on 5 sequences with $\overline{ape} > \SI{20}{\degree}$. Even on the sequences where these methods do not completely fail, they still exhibit significantly higher errors compared to our method, with $\overline{ape}$ up to \SI{19.126}{\degree}.

In contrast, EROAM successfully processes all sequences while maintaining superior accuracy. Our method achieves the lowest error rates across all sequences, with $\overline{ape}$ below \SI{1.0}{\degree} in the vast majority of sequences. Even in challenging large-angle motion scenarios, such as the \texttt{limx-rotation} sequence where the LimX TRON1 robot performs over 3 full rotations, EROAM still demonstrates excellent performance with $\overline{ape}=\SI{1.720}{\degree}$ and $\overline{rpe}=\SI{0.939}{\degree}$. 

Our method demonstrates particular robustness in challenging real-world scenarios. In sequences containing dynamic objects such as pedestrians, motorcycles, and vehicles (\texttt{carpark}, \texttt{crossroad}, \texttt{vehicle}), EROAM maintains excellent performance thanks to our Regional Density Management (RDM) approach, which prevents the continuous accumulation of events from moving objects into the spherical map. Additionally, EROAM shows consistent accuracy on data collected from our actual robotic platform, with the \texttt{limx-rotation} and \texttt{limx-sway} sequences captured using the LimX TRON 1 wheeled bipedal robot demonstrating reliable rotation estimation even under the vibrations and motion patterns typical of real robotic operation.

The superior accuracy of our method is particularly evident in the \texttt{window-building} sequence, as shown in \prettyref{fig:window_building_rotation}. While both CM-GAE and CMax-SLAM exhibit significant drift and inconsistent estimation across all three rotation angles, EROAM closely tracks the ground truth trajectory throughout the sequence. The detailed views between \SI{15}{s} and \SI{20}{s} clearly demonstrate our method's ability to maintain accurate rotation estimation even during complex motion patterns.

This accuracy in rotation estimation directly translates to high-quality panoramic reconstruction results. As shown in \prettyref{fig:panoramic_comparison_real}, EROAM produces significantly clearer panoramas compared to CMax-SLAM. The comparison between the \texttt{front-gate} sequence results demonstrates our method's superior performance in maintaining structural clarity and avoiding ghosting artifacts. This advantage is further emphasized in the \texttt{embankment-360} sequence, where EROAM successfully reconstructs a complete 360° panorama with consistent quality.

Additional reconstruction results \prettyref{fig:eroam_additional_results}) demonstrate consistent performance across diverse environments.

\begin{revisedtextblock}
\input{figs_and_tabs/tab_ecrot_runtime}
\input{figs_and_tabs/tab_eroam_campus_runtime}
\end{revisedtextblock}

\subsection{Runtime Analysis}
\label{subsec:runtime_analysis}

\prettyref{tab:runtime_comparison} and \prettyref{tab:runtime_comparison_real} present the runtime comparison of RTPT, CM-GAE, CMax-SLAM, and our EROAM method on both synthetic ECRot sequences and real-world EROAM-campus sequences. Note that SMT is excluded from the comparison as it fails to process any sequences successfully and exhibits prohibitively high computational costs even during failed attempts. All experiments were conducted on a laptop with an Intel Core i9-14900HX CPU and 32GB RAM, with RTPT utilizing an NVIDIA RTX 4060 GPU while all other methods run on CPU only.

Our implementation uses a consistent processing frequency of $f = \SI{1000}{\Hz}$ across both datasets, demonstrating EROAM's capability to handle different resolution inputs while maintaining real-time performance. Processing times may be slightly shorter than sequence durations, which is expected when measuring real-time processing performance.

EROAM demonstrates superior computational efficiency across both datasets. For synthetic data, our method processes most sequences in approximately \SI{5}{s}. On real-world data, taking the \SI{31.413}{s} \texttt{embankment} sequence as an example, EROAM achieves remarkable speedups: 6.3$\times$ compared to CM-GAE, 5.3$\times$ compared to RTPT (despite its GPU acceleration), and 17.6$\times$ compared to CMax-SLAM. CMax-SLAM's performance degrades significantly with sequence duration due to its combination of contrast maximization and backend optimization, with processing time scaling poorly with event density. CM-GAE exhibits similar limitations. EROAM, in contrast, achieves real-time performance even for extended sequences - processing the \SI{47.169}{s} \texttt{lawn} sequence in \SI{47.158}{s}, while other methods require substantially longer: CM-GAE takes \SI{1318.581}{s} and CMax-SLAM needs \SI{2553.260}{s}.

This consistent performance stems from two key design choices. First, our spherical map representation significantly simplifies map maintenance. Unlike RTPT's complex map updates and feature management operations, EROAM employs efficient incremental updates to its spherical map using an ikd-Tree and manages point density with Regional Density Management (RDM), substantially reducing computational overhead.

Second, our ES-ICP algorithm fundamentally avoids the computational bottlenecks inherent in contrast maximization-based methods. While CM-GAE and CMax-SLAM require iterative contrast computation over event time windows and computationally intensive event warping operations after each velocity update, EROAM's ES-ICP solution:
\begin{itemize}
    \item Eliminates repeated event warping requirements
    \item Enables parallel processing of point-to-line distances and their Jacobians across CPU cores
    \item Maintains consistent computational load regardless of event rate or scene complexity
\end{itemize}

The experimental results clearly demonstrate EROAM's superior computational efficiency and scalability. By maintaining real-time performance across both synthetic and real-world scenarios while requiring only CPU resources, EROAM presents a practical solution for event-based rotational motion estimation in real-world applications.

%% file: figs_and_tabs/fig_ecrot_synthetic_panoramic.tex
\begin{figure*}[t]
    \centering
    {
    \setlength{\fboxrule}{0.1pt}
    
    \subfloat[Panorama generated using CM-GAE's estimated trajectory]{%
        \begin{tikzpicture}[
            zoomboxarray,
            spymargin=0.1em,
            caption margin=2ex,
            zoomboxes below,
            zoomboxarray columns=3,
            zoomboxarray rows=1,
            connect zoomboxes,
            zoombox paths/.append style={thick, red}]
            \node [image node] { {\color{gray}\fbox{\includegraphics[width=0.48\textwidth]{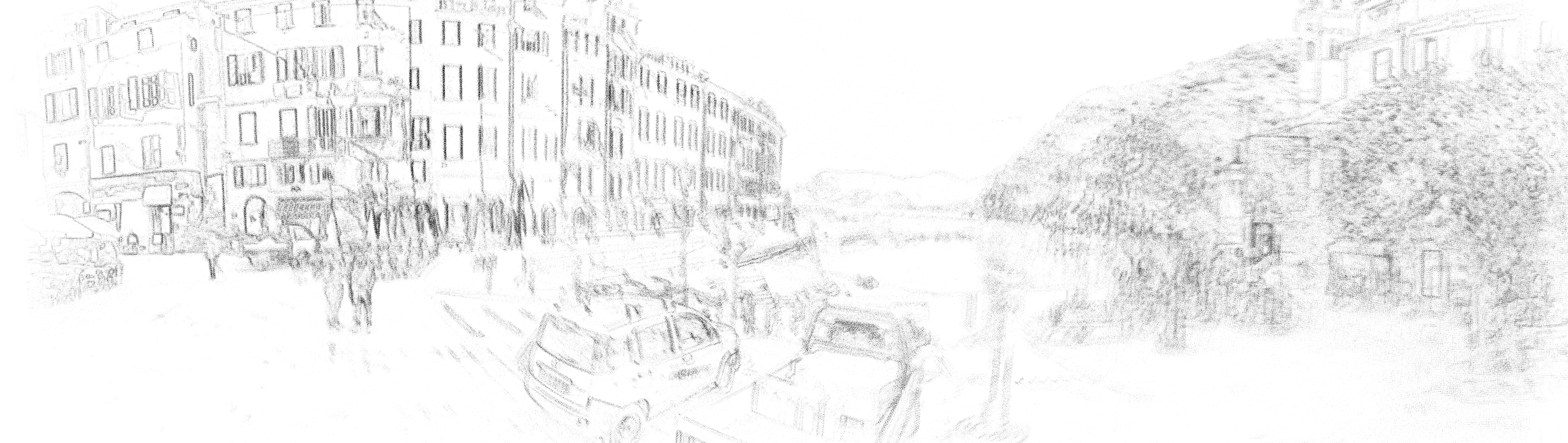}}} };
            \zoombox[magnification=8]{0.44,0.17}
            \zoombox[magnification=8]{0.49,0.69}
            \zoombox[magnification=8]{0.54,0.28}
        \end{tikzpicture}
        \label{fig:panoramic_town_cmgae}%
    }%
    \hfill
    \subfloat[Original CMax-SLAM panorama output (empty regions trimmed)]{%
        \begin{tikzpicture}[
            zoomboxarray,
            spymargin=0.1em,
            caption margin=2ex,
            zoomboxes below,
            zoomboxarray columns=3,
            zoomboxarray rows=1,
            connect zoomboxes,
            zoombox paths/.append style={thick, red}]
            \node [image node] { {\color{gray}\fbox{\includegraphics[width=0.48\textwidth,
            trim={53mm} {55.1mm} {52mm} {53mm},
            clip]{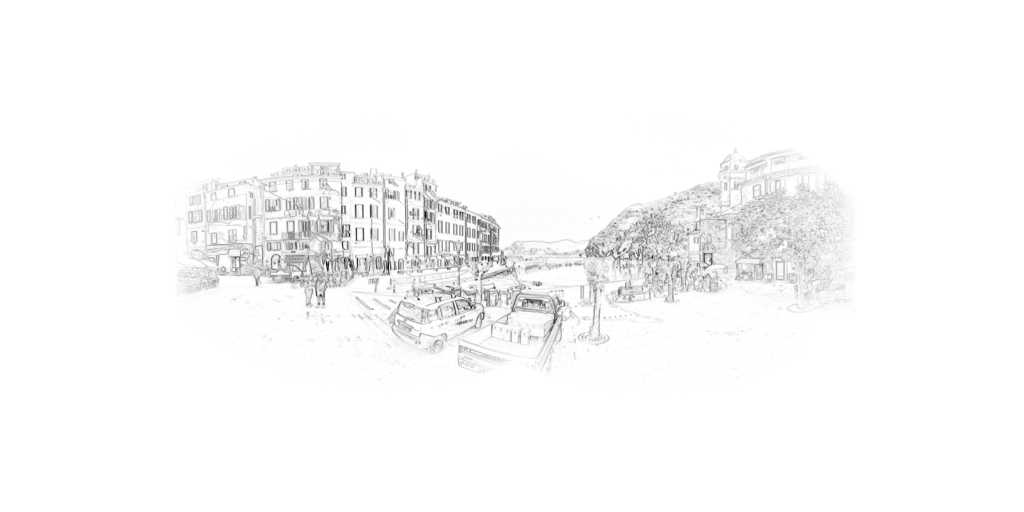}}} };
            \zoombox[magnification=8]{0.42,0.15}
            \zoombox[magnification=8]{0.47,0.62}
            \zoombox[magnification=8]{0.525,0.287}
        \end{tikzpicture}
        \label{fig:panoramic_town_cmaxslam_origin}%
    }%
    
    \vspace{4mm}
    
    \subfloat[Panorama generated using CMax-SLAM's estimated trajectory]{%
        \begin{tikzpicture}[
            zoomboxarray,
            spymargin=0.1em,
            caption margin=2ex,
            zoomboxes below,
            zoomboxarray columns=3,
            zoomboxarray rows=1,
            connect zoomboxes,
            zoombox paths/.append style={thick, red}]
            \node [image node] { {\color{gray}\fbox{\includegraphics[width=0.48\textwidth]{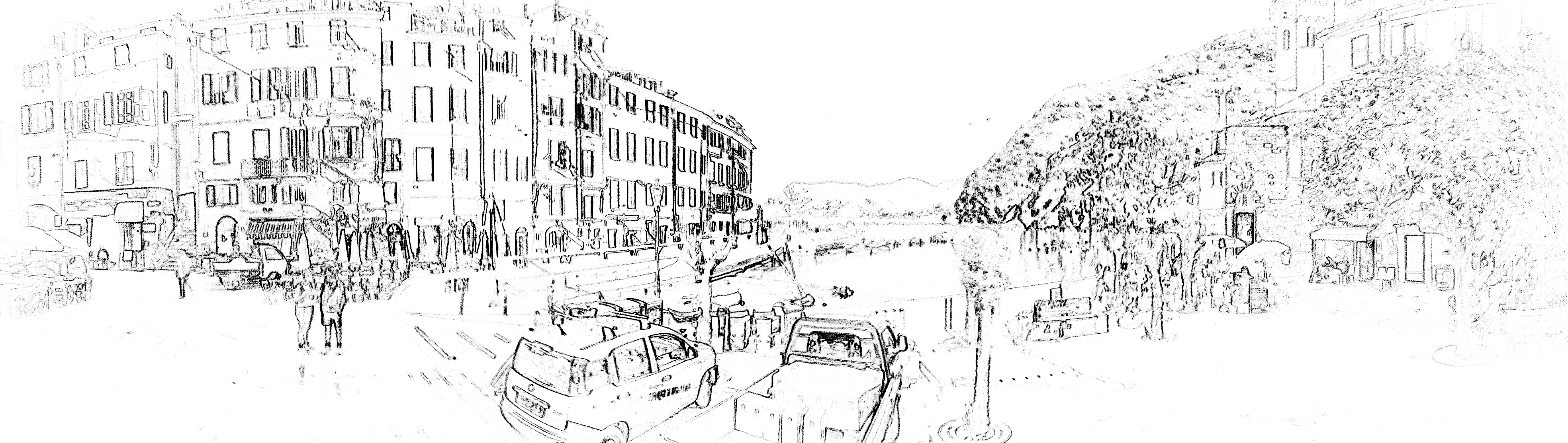}}} };
            \zoombox[magnification=8]{0.42,0.15}
            \zoombox[magnification=8]{0.47,0.65}
            \zoombox[magnification=8]{0.525,0.257}
        \end{tikzpicture}
        \label{fig:panoramic_town_cmaxslam}%
    }%
    \hfill
    \subfloat[Panorama generated using EROAM's (Ours) estimated trajectory]{%
        \begin{tikzpicture}[
            zoomboxarray,
            spymargin=0.1em,
            caption margin=2ex,
            zoomboxes below,
            zoomboxarray columns=3,
            zoomboxarray rows=1,
            connect zoomboxes,
            zoombox paths/.append style={thick, red}]
            \node [image node] { {\color{gray}\fbox{\includegraphics[width=0.48\textwidth]{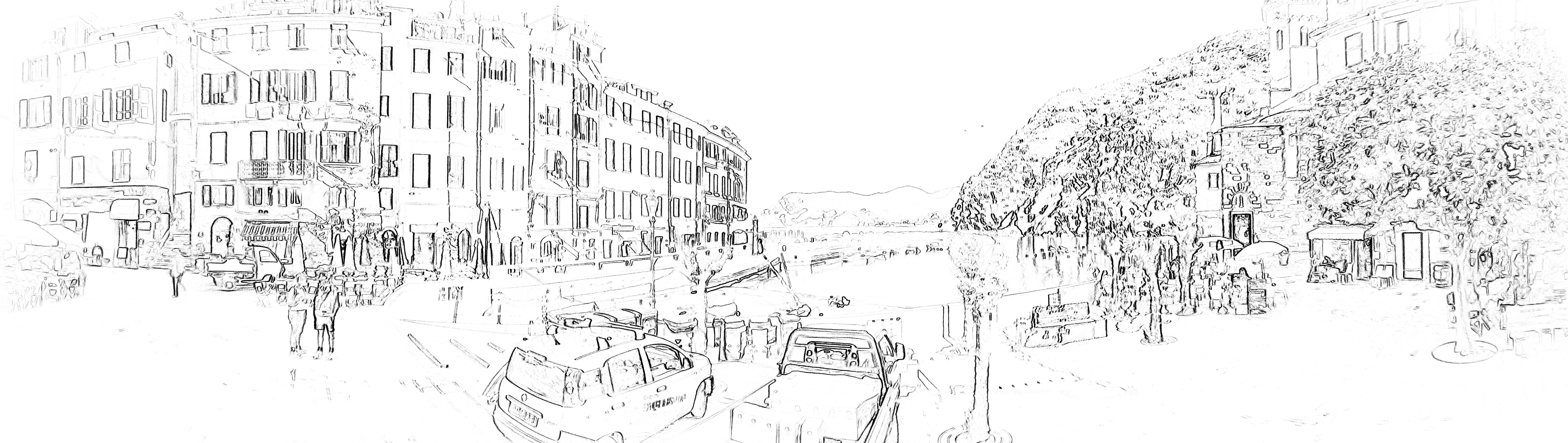}}} };
            \zoombox[magnification=8]{0.42,0.15}
            \zoombox[magnification=8]{0.47,0.63}
            \zoombox[magnification=8]{0.525,0.24}
        \end{tikzpicture}
        \label{fig:panoramic_town_eroam}%
    }%
    }
    \caption{Qualitative comparison of panoramic mapping results on the \texttt{town} sequence from ECRot dataset \cite{cmax-slam}. (a,c,d) show panoramas generated using our unified pipeline ($7617\times2000$ resolution) with different methods' estimated trajectories, where identical event window size (\SI{0.2}{ms}) and colorization scheme are applied. (b) shows the original panorama output from CMax-SLAM's implementation for reference. The zoomed-in regions highlight that our method produces sharper edges and clearer structural details compared to other approaches.}
    \label{fig:ecrot_synthetic_panoramic_comparison}
\end{figure*}

%% file: figs_and_tabs/tab_ecrot_synthetic.tex
\begin{table*}[ht]
\centering
\caption{Comparison of rotational motion estimation accuracy on ECRot\cite{cmax-slam} synthetic datasets.}
\label{tab:ecrot_synthetic_results}
\resizebox{0.8\textwidth}{!}{%
\begin{tabular}{l*{12}{c}}
\toprule
Sequence & \multicolumn{2}{c}{bay\cite{cmax-slam}} & \multicolumn{2}{c}{bicycle\cite{cmax-slam}} & \multicolumn{2}{c}{city\cite{cmax-slam}} & \multicolumn{2}{c}{street\cite{cmax-slam}} & \multicolumn{2}{c}{town\cite{cmax-slam}} & \multicolumn{2}{c}{playroom\cite{cmax-slam}} \\
\cmidrule(lr){2-3} \cmidrule(lr){4-5} \cmidrule(lr){6-7} \cmidrule(lr){8-9} \cmidrule(lr){10-11} \cmidrule(lr){12-13}
& $\overline{ape}$ (\si{\degree}) & $\overline{rpe}$ (\si{\degree}) & $\overline{ape}$ (\si{\degree}) & $\overline{rpe}$ (\si{\degree}) & $\overline{ape}$ (\si{\degree}) & $\overline{rpe}$ (\si{\degree}) & $\overline{ape}$ (\si{\degree}) & $\overline{rpe}$ (\si{\degree}) & $\overline{ape}$ (\si{\degree}) & $\overline{rpe}$ (\si{\degree}) & $\overline{ape}$ (\si{\degree}) & $\overline{rpe}$ (\si{\degree}) \\
\midrule
SMT \cite{smt} & - & - & - & - & - & - & - & - & - & - & - & - \\
RTPT \cite{rtpt} & - & - & - & - & - & - & - & - & - & - & - & - \\
CM-GAE \cite{cm-gae} & - & - & 0.856 & 0.244 & - & - & - & - & 2.600 & 0.603 & 3.684 & 0.823 \\
CMax-SLAM \cite{cmax-slam} & 0.802 & 0.141 & 0.727 & 0.098 & 1.113 & 0.294 & 0.517 & 0.181 & 1.238 & 0.130 & 0.727 & \textbf{0.095} \\
EROAM (\textbf{ours}) & \textbf{0.239} & \textbf{0.062} & \textbf{0.159} & \textbf{0.076} & \textbf{0.111} & \textbf{0.083} & \textbf{0.166} & \textbf{0.093} & \textbf{0.342} & \textbf{0.117} & \textbf{0.311} & 0.150 \\
\bottomrule
\end{tabular}%
}
\begin{tablenotes}
\item ``-" indicates a failure case where after multiple attempts, the method still either produces $\overline{ape}$ larger than 20\si{\degree}, crashes during execution, or outputs undefined quaternions.
\end{tablenotes}
\end{table*}

%% file: figs_and_tabs/fig_roll_pitch_yaw_town.tex
\begin{figure*}[t]
    \centering
    \includegraphics[width=\textwidth]{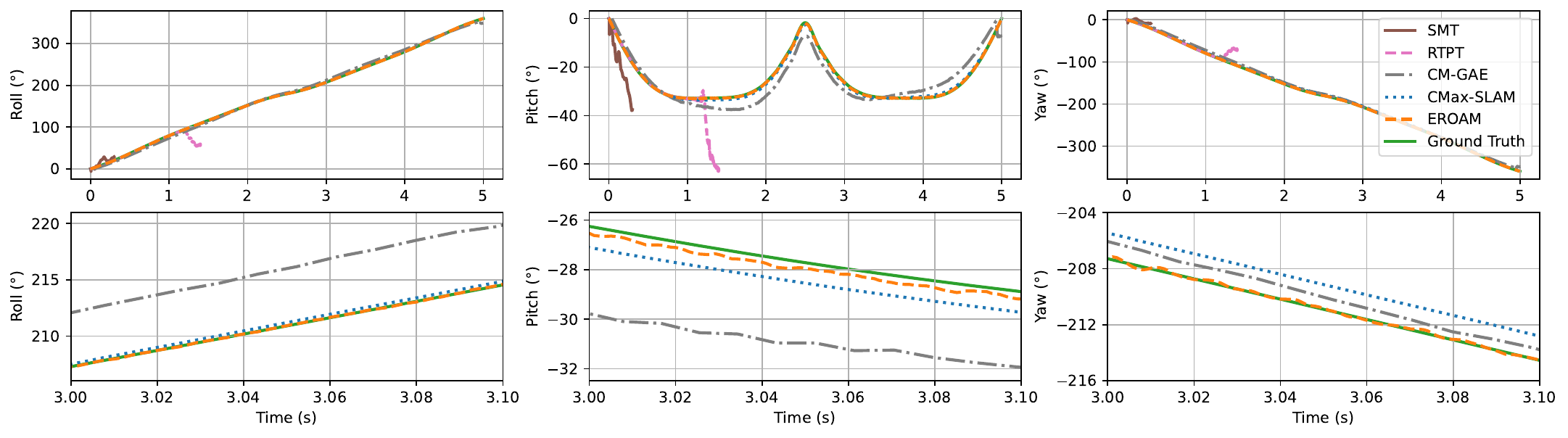}
    \caption{Rotation estimation results on the \texttt{town} sequence from ECRot dataset. The top row shows the full trajectories of roll, pitch, and yaw angles, while the bottom row shows the zoomed-in views from \SI{3.00}{s} to \SI{3.10}{s}. Our method (EROAM) achieves comparable accuracy with the ground truth, demonstrating robust rotation estimation performance. The trajectories of SMT and RTPT are truncated after significant deviation from ground truth for better visualization clarity.}
    \label{fig:rotation_comparison}
\end{figure*}

%% file: figs_and_tabs/fig_extended_esim_ape_rpe.tex
\begin{figure*}[!t]
    \centering
    \subfloat[Angular motion characteristics of \texttt{DM-1} to \texttt{DM-8} sequences]{%
        \includegraphics[width=0.32\textwidth]{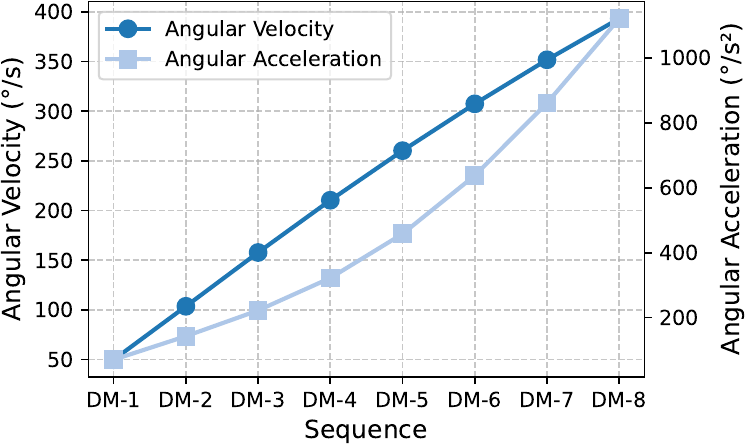}
        \label{fig:angular_motion}%
    }%
    \hfill
    \subfloat[Performance comparison on \texttt{DM-1} to \texttt{DM-8} sequences]{%
        \includegraphics[width=0.3225\textwidth]{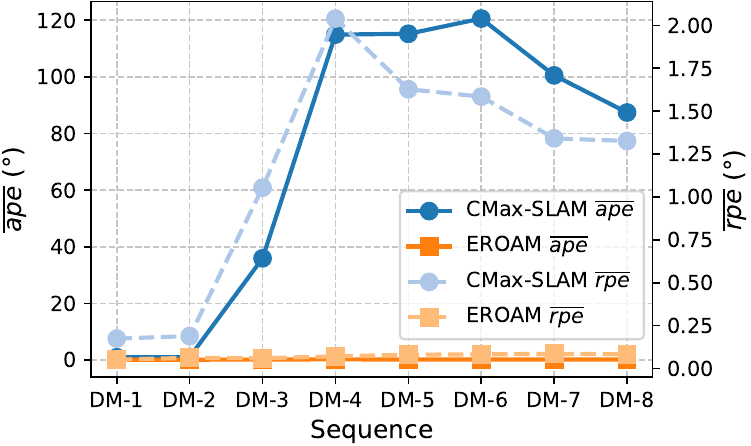}
        \label{fig:omega_comparison}%
    }%
    \hfill
    \subfloat[Performance comparison on \texttt{LD-10} to \texttt{LD-80} sequences]{%
        \includegraphics[width=0.32\textwidth]{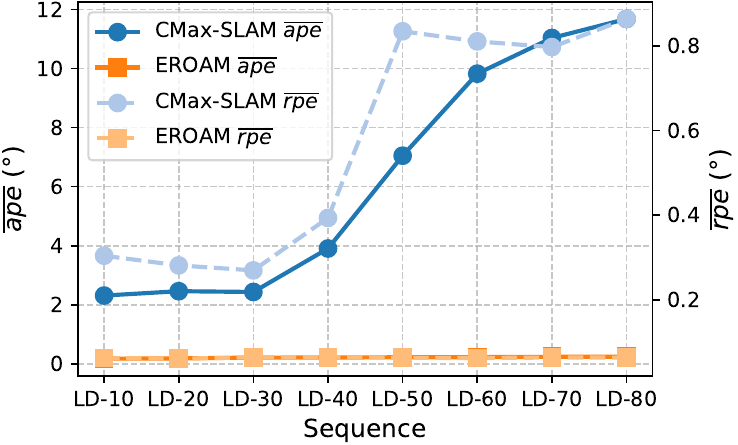}
        \label{fig:duration_comparison}%
    }%
    \caption{Quantitative evaluation results. (a) Shows the angular velocity (\SI{49.83}{\degree/\second} to \SI{393.21}{\degree/\second}) and acceleration (\SI{70.41}{\degree/\second\squared} to \SI{1121.32}{\degree/\second\squared}) characteristics of DM-1 to DM-8 sequences (each \SI{5}{s} long). (b) Compares the $\overline{ape}$ and $\overline{rpe}$ metrics across DM-1 to DM-8 sequences with increasing angular velocities. (c) Compares the $\overline{ape}$ and $\overline{rpe}$ metrics for LD-10 to LD-80 sequences of increasing durations (\SI{10}{s} to \SI{80}{s}).}
    \label{fig:quantitative_evaluation}
\end{figure*}

%% file: figs_and_tabs/tab_extended_esim_ape_rpe_omega.tex
\begin{table*}[!t]
\centering
\caption{Quantitative comparison of rotational state estimation under increasing angular velocities.}
\label{tab:omega_comparison}
\resizebox{\textwidth}{!}{%
\begin{tabular}{l*{16}{c}}
\toprule
Sequence & \multicolumn{2}{c}{DM-1} & \multicolumn{2}{c}{DM-2} & \multicolumn{2}{c}{DM-3} & \multicolumn{2}{c}{DM-4} & \multicolumn{2}{c}{DM-5} & \multicolumn{2}{c}{DM-6} & \multicolumn{2}{c}{DM-7} & \multicolumn{2}{c}{DM-8} \\
\cmidrule(lr){2-3} \cmidrule(lr){4-5} \cmidrule(lr){6-7} \cmidrule(lr){8-9} \cmidrule(lr){10-11} \cmidrule(lr){12-13} \cmidrule(lr){14-15} \cmidrule(lr){16-17}
& $\overline{ape}$ (\si{\degree}) & $\overline{rpe}$ (\si{\degree}) & $\overline{ape}$ (\si{\degree}) & $\overline{rpe}$ (\si{\degree}) & $\overline{ape}$ (\si{\degree}) & $\overline{rpe}$ (\si{\degree}) & $\overline{ape}$ (\si{\degree}) & $\overline{rpe}$ (\si{\degree}) & $\overline{ape}$ (\si{\degree}) & $\overline{rpe}$ (\si{\degree}) & $\overline{ape}$ (\si{\degree}) & $\overline{rpe}$ (\si{\degree}) & $\overline{ape}$ (\si{\degree}) & $\overline{rpe}$ (\si{\degree}) & $\overline{ape}$ (\si{\degree}) & $\overline{rpe}$ (\si{\degree}) \\
\midrule
CMax-SLAM\cite{cmax-slam} & 0.947 & 0.174 & 1.032 & 0.189 & 35.927 & 1.054 & 114.962 & 2.040 & 115.238 & 1.627 & 120.625 & 1.585 & 100.548 & 1.341 & 87.440 & 1.326 \\
EROAM (\textbf{ours}) & \textbf{0.139} & \textbf{0.155} & \textbf{0.192} & \textbf{0.093} & \textbf{0.333} & \textbf{0.078} & \textbf{0.237} & \textbf{0.119} & \textbf{0.307} & \textbf{0.091} & \textbf{0.731} & \textbf{0.111} & \textbf{0.300} & \textbf{0.116} & \textbf{0.176} & \textbf{0.105} \\
\bottomrule
\end{tabular}%
}
\end{table*}

%% file: figs_and_tabs/tab_extended_esim_ape_rpe_duration.tex
\begin{table*}[!t]
\centering
\caption{Assessment of long-term estimation stability across sequences of increasing duration.}\label{tab:duration_comparison}
\resizebox{\textwidth}{!}{%
\begin{tabular}{l*{16}{c}}
\toprule
Sequence & \multicolumn{2}{c}{LD-10} & \multicolumn{2}{c}{LD-20} & \multicolumn{2}{c}{LD-30} & \multicolumn{2}{c}{LD-40} & \multicolumn{2}{c}{LD-50} & \multicolumn{2}{c}{LD-60} & \multicolumn{2}{c}{LD-70} & \multicolumn{2}{c}{LD-80} \\
\cmidrule(lr){2-3} \cmidrule(lr){4-5} \cmidrule(lr){6-7} \cmidrule(lr){8-9} \cmidrule(lr){10-11} \cmidrule(lr){12-13} \cmidrule(lr){14-15} \cmidrule(lr){16-17}
& $\overline{ape}$ (\si{\degree}) & $\overline{rpe}$ (\si{\degree}) & $\overline{ape}$ (\si{\degree}) & $\overline{rpe}$ (\si{\degree}) & $\overline{ape}$ (\si{\degree}) & $\overline{rpe}$ (\si{\degree}) & $\overline{ape}$ (\si{\degree}) & $\overline{rpe}$ (\si{\degree}) & $\overline{ape}$ (\si{\degree}) & $\overline{rpe}$ (\si{\degree}) & $\overline{ape}$ (\si{\degree}) & $\overline{rpe}$ (\si{\degree}) & $\overline{ape}$ (\si{\degree}) & $\overline{rpe}$ (\si{\degree}) & $\overline{ape}$ (\si{\degree}) & $\overline{rpe}$ (\si{\degree}) \\
\midrule
CMax-SLAM\cite{cmax-slam} & 2.312 & 0.304 & 2.460 & 0.281 & 2.434 & 0.269 & 3.905 & 0.393 & 7.047 & 0.834 & 9.824 & 0.810 & 11.033 & 0.797 & 11.677 & 0.863 \\
EROAM (\textbf{ours}) & \textbf{0.131} & \textbf{0.083} & \textbf{0.128} & \textbf{0.078} & \textbf{0.133} & \textbf{0.083} & \textbf{0.115} & \textbf{0.129} & \textbf{0.121} & \textbf{0.106} & \textbf{0.120} & \textbf{0.102} & \textbf{0.126} & \textbf{0.082} & \textbf{0.119} & \textbf{0.095} \\
\bottomrule
\end{tabular}%
}
\end{table*}

%% file: figs_and_tabs/fig_roll_pitch_yaw_lt_80.tex
\begin{figure*}[t]
    \centering
    \includegraphics[width=\textwidth]{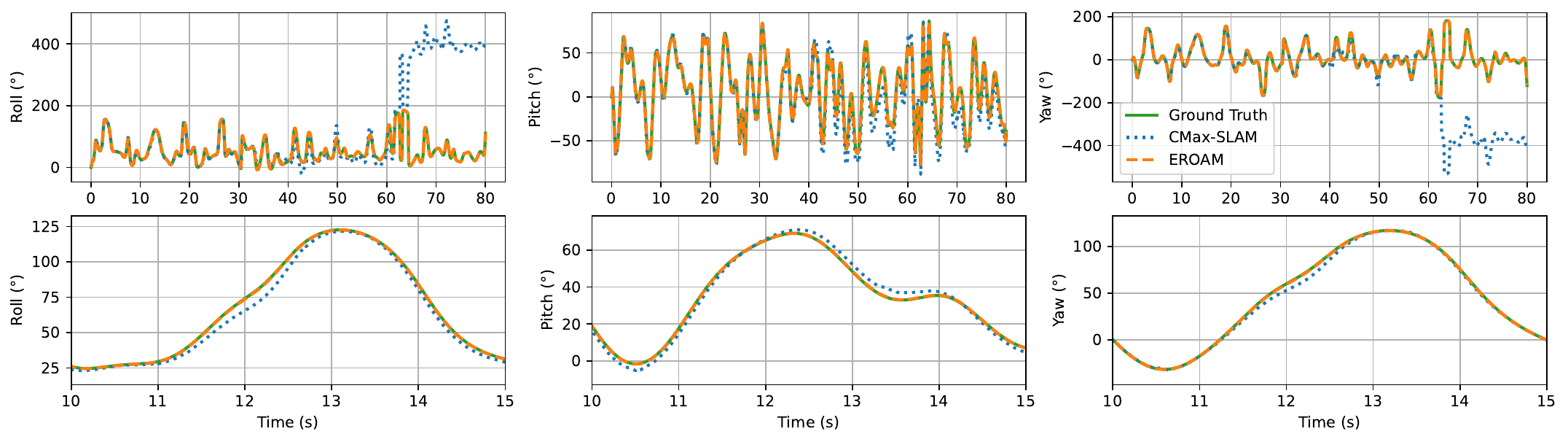}
    \caption{Comparison of rotational state estimation on the \texttt{LD-80} sequence. The top row shows the complete \SI{80}{s} trajectories for roll, pitch, and yaw angles, where CMax-SLAM exhibits significant consistency issues and deviates from ground truth starting from \SI{40}{s}. The bottom row presents detailed views between \SI{10}{s} and \SI{15}{s}, revealing that CMax-SLAM's estimation already shows noticeable errors even in this early stage. In contrast, EROAM maintains consistent accuracy throughout the entire sequence, closely aligning with ground truth in both global and local perspectives.}
    \label{fig:ld_80_rpy_comparison_with_zoom}
\end{figure*}

%% file: figs_and_tabs/fig_extended_esim_dm4.tex
\begin{figure*}[!t]
    \centering
    {
    \setlength{\fboxrule}{0.1pt}
    
    \subfloat[\begin{revisedtextblockCR}Panoramic image from CMax-SLAM\end{revisedtextblockCR}]{%
        \begin{tikzpicture}[
            zoomboxarray,
            spymargin=0.1em,
            caption margin=2ex,
            zoomboxes below,
            zoomboxarray columns=3,
            zoomboxarray rows=1,
            connect zoomboxes,
            zoombox paths/.append style={thick, red}]
            \node [image node] { {\color{gray}\fbox{\includegraphics[width=0.48\textwidth,
            trim={0mm} {6mm} {0mm} {0mm},
            clip]{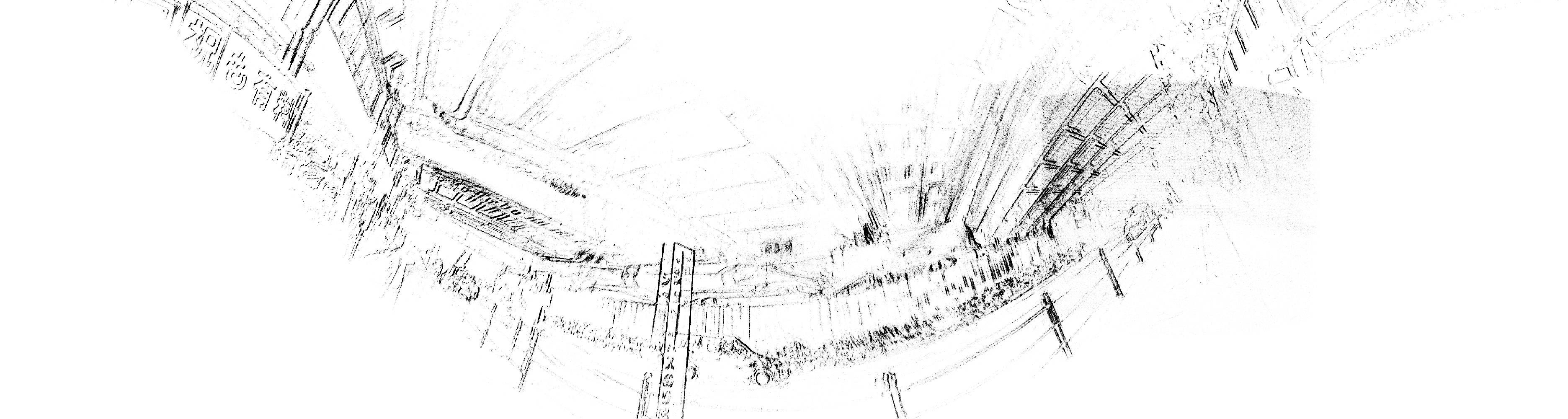}}} };
            \zoombox[magnification=4]{0.23,0.83}
            \zoombox[magnification=4]{0.43,0.32}
            \zoombox[magnification=4]{0.67,0.60}
        \end{tikzpicture}
        \label{fig:panoramic_dm4_cmaxslam}%
    }%
    \hfill
    \subfloat[Panoramic image from EROAM (Ours)]{%
        \begin{tikzpicture}[
            zoomboxarray,
            spymargin=0.1em,
            caption margin=2ex,
            zoomboxes below,
            zoomboxarray columns=3,
            zoomboxarray rows=1,
            connect zoomboxes,
            zoombox paths/.append style={thick, red}]
            \node [image node] { {\color{gray}\fbox{\includegraphics[width=0.48\textwidth]{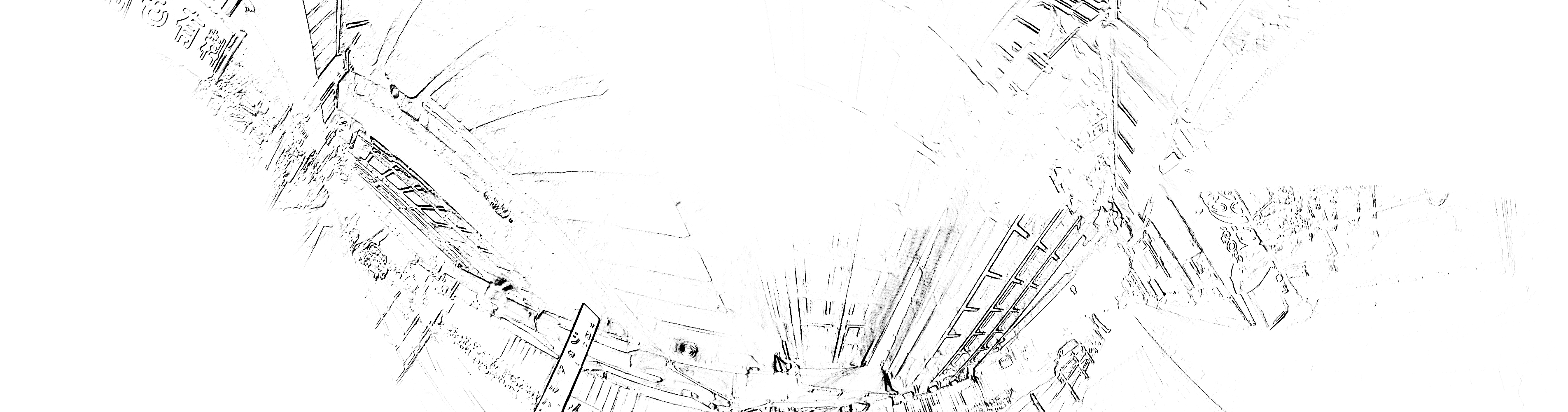}}} };
            \zoombox[magnification=4]{0.22,0.80}
            \zoombox[magnification=4]{0.36,0.26}
            \zoombox[magnification=4]{0.65,0.45}
        \end{tikzpicture}
        \label{fig:panoramic_dm4_eroam}%
    }%
    }
    \caption{Qualitative comparison of panoramic mapping results on the \texttt{DM-4} sequence with high angular velocity. CMax-SLAM's result exhibits significant ghosting artifacts and double edges in structures, indicating severe rotation estimation errors under high-dynamic motion. These artifacts are particularly visible in the zoomed-in regions, where single structures appear multiple times due to inconsistent motion estimates. In contrast, EROAM produces a panorama with sharp details and precisely aligned structures, demonstrating accurate rotation estimation throughout the sequence.}
    \label{fig:dm4_panoramic_comparison}
\end{figure*}

%% file: figs_and_tabs/tab_university_data_ape_rpe.tex
\begin{table*}[t]
\centering
\caption{Comparison of rotational motion estimation accuracy on EROAM-campus datasets.}
\label{tab:real_world_evaluation}
\resizebox{\textwidth}{!}{%

\begin{tabular}{l*{16}{c}}
\toprule
Sequence & \multicolumn{2}{c}{canteen-entrance} & \multicolumn{2}{c}{distant-building} & \multicolumn{2}{c}{embankment-360} & \multicolumn{2}{c}{embankment} & \multicolumn{2}{c}{lawn-360} & \multicolumn{2}{c}{lawn} & \multicolumn{2}{c}{lecture-hall} & \multicolumn{2}{c}{front-gate} \\
\cmidrule(lr){2-3} \cmidrule(lr){4-5} \cmidrule(lr){6-7} \cmidrule(lr){8-9} \cmidrule(lr){10-11} \cmidrule(lr){12-13} \cmidrule(lr){14-15} \cmidrule(lr){16-17}
& $\overline{ape}$ (\si{\degree}) & $\overline{rpe}$ (\si{\degree}) & $\overline{ape}$ (\si{\degree}) & $\overline{rpe}$ (\si{\degree}) & $\overline{ape}$ (\si{\degree}) & $\overline{rpe}$ (\si{\degree}) & $\overline{ape}$ (\si{\degree}) & $\overline{rpe}$ (\si{\degree}) & $\overline{ape}$ (\si{\degree}) & $\overline{rpe}$ (\si{\degree}) & $\overline{ape}$ (\si{\degree}) & $\overline{rpe}$ (\si{\degree}) & $\overline{ape}$ (\si{\degree}) & $\overline{rpe}$ (\si{\degree}) & $\overline{ape}$ (\si{\degree}) & $\overline{rpe}$ (\si{\degree}) \\
\midrule
SMT \cite{smt} & - & - & - & - & - & - & - & - & - & - & - & - & - & - & - & - \\
RTPT \cite{rtpt} & - & - & - & - & - & - & 0.654 & 0.403 & - & - & - & - & - & - & - & - \\
CM-GAE \cite{cm-gae} & 0.688 & 0.371 & 5.915 & 0.899 & 4.727 & 0.729 & 3.228 & 0.424 & 7.776 & 0.670 & - & - & 3.103 & 0.407 & - & - \\
CMax-SLAM \cite{cmax-slam} & 0.662 & 0.557 & 5.958 & 1.312 & - & - & 4.658 & 2.298 & 8.733 & 0.691 & - & - & 0.518 & 0.457 & 13.549 & 1.227 \\
EROAM (\textbf{ours}) & \textbf{0.489} & \textbf{0.295} & \textbf{0.239} & \textbf{0.247} & \textbf{1.450} & \textbf{0.231} & \textbf{0.499} & \textbf{0.233} & \textbf{1.248} & \textbf{0.356} & \textbf{0.490} & \textbf{0.208} & \textbf{0.241} & \textbf{0.186} & \textbf{0.475} & \textbf{0.244} \\
\midrule
\multicolumn{17}{c}{} \\
\midrule
Sequence & \multicolumn{2}{c}{rooftop} & \multicolumn{2}{c}{window-building} & \multicolumn{2}{c}{carpark} & \multicolumn{2}{c}{condo} & \multicolumn{2}{c}{crossroad} & \multicolumn{2}{c}{vehicle} & \multicolumn{2}{c}{limx-rotation} & \multicolumn{2}{c}{limx-sway} \\
\cmidrule(lr){2-3} \cmidrule(lr){4-5} \cmidrule(lr){6-7} \cmidrule(lr){8-9} \cmidrule(lr){10-11} \cmidrule(lr){12-13} \cmidrule(lr){14-15} \cmidrule(lr){16-17}
& $\overline{ape}$ (\si{\degree}) & $\overline{rpe}$ (\si{\degree}) & $\overline{ape}$ (\si{\degree}) & $\overline{rpe}$ (\si{\degree}) & $\overline{ape}$ (\si{\degree}) & $\overline{rpe}$ (\si{\degree}) & $\overline{ape}$ (\si{\degree}) & $\overline{rpe}$ (\si{\degree}) & $\overline{ape}$ (\si{\degree}) & $\overline{rpe}$ (\si{\degree}) & $\overline{ape}$ (\si{\degree}) & $\overline{rpe}$ (\si{\degree}) & $\overline{ape}$ (\si{\degree}) & $\overline{rpe}$ (\si{\degree}) & $\overline{ape}$ (\si{\degree}) & $\overline{rpe}$ (\si{\degree}) \\
\midrule
SMT \cite{smt} & - & - & - & - & - & - & - & - & - & - & - & - & - & - & - & - \\
RTPT \cite{rtpt} & - & - & - & - & - & - & - & - & - & - & - & - & - & - & - & - \\
CM-GAE \cite{cm-gae} & 14.063 & 0.695 & 10.954 & 1.227 & 16.567 & 1.076 & 19.126 & 2.430 & - & - & - & - & - & - & - & - \\
CMax-SLAM \cite{cmax-slam} & 0.841 & 0.406 & 8.201 & 4.698 & - & - & - & - & - & - & 6.119 & 0.891 & 14.473 & 1.412 & 1.141 & 0.549 \\
EROAM (\textbf{ours}) & \textbf{0.438} & \textbf{0.208} & \textbf{0.256} & \textbf{0.167} & \textbf{0.612} & \textbf{0.222} & \textbf{0.543} & \textbf{0.280} & \textbf{0.539} & \textbf{0.204} & \textbf{0.666} & \textbf{0.291} & \textbf{1.720} & \textbf{0.939} & \textbf{0.684} & \textbf{0.426} \\
\bottomrule
\end{tabular}

}

\begin{tablenotes}
\item ``-" indicates a failure case where after multiple attempts, the method still either produces $\overline{ape}$ larger than 20\si{\degree}, crashes during execution, or outputs undefined quaternions.
\end{tablenotes}
\end{table*}

%% file: figs_and_tabs/fig_roll_pitch_yaw_window_building.tex
\begin{figure*}[!t]
    \centering
    \includegraphics[width=\textwidth]{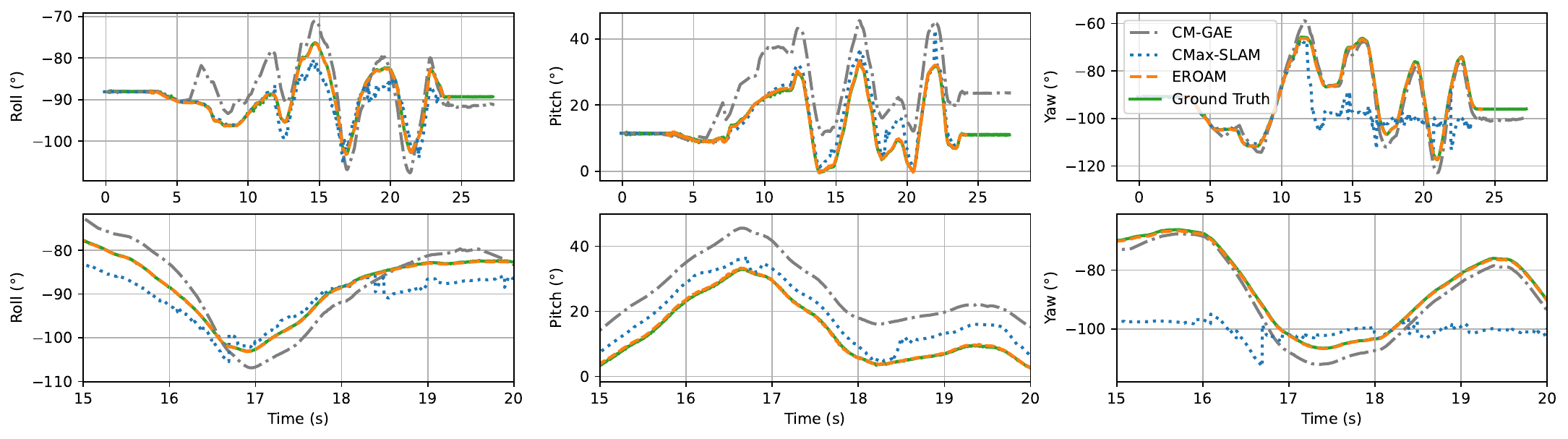}
    \caption{Comparison of rotational state estimation on the \texttt{window-building} sequence. The top row shows the complete trajectories for roll, pitch, and yaw angles, while the bottom row presents detailed views between \SI{15}{s} and \SI{20}{s}. Our method (EROAM) maintains accurate tracking throughout the sequence, closely following the ground truth trajectory, while CM-GAE and CMax-SLAM show noticeable deviations, particularly evident in the zoomed-in views.}
    \label{fig:window_building_rotation}
\end{figure*}

%% file: figs_and_tabs/fig_eroam_campus_pano_comparison.tex
\begin{figure*}[!t]
    \centering
    {
    \setlength{\fboxrule}{0.1pt}
    
    \subfloat[\begin{revisedtextblockCR}CMax-SLAM result on \texttt{front-gate} sequence\end{revisedtextblockCR}]{%
        \begin{tikzpicture}[
            zoomboxarray,
            spymargin=0.1em,
            caption margin=2ex,
            zoomboxes below,
            zoomboxarray columns=3,
            zoomboxarray rows=1,
            connect zoomboxes,
            zoombox paths/.append style={thick, red}]
            \node [image node] { {\color{gray}\fbox{\includegraphics[width=0.47\textwidth,
            trim={0mm} {22mm} {0mm} {56mm},
            clip]{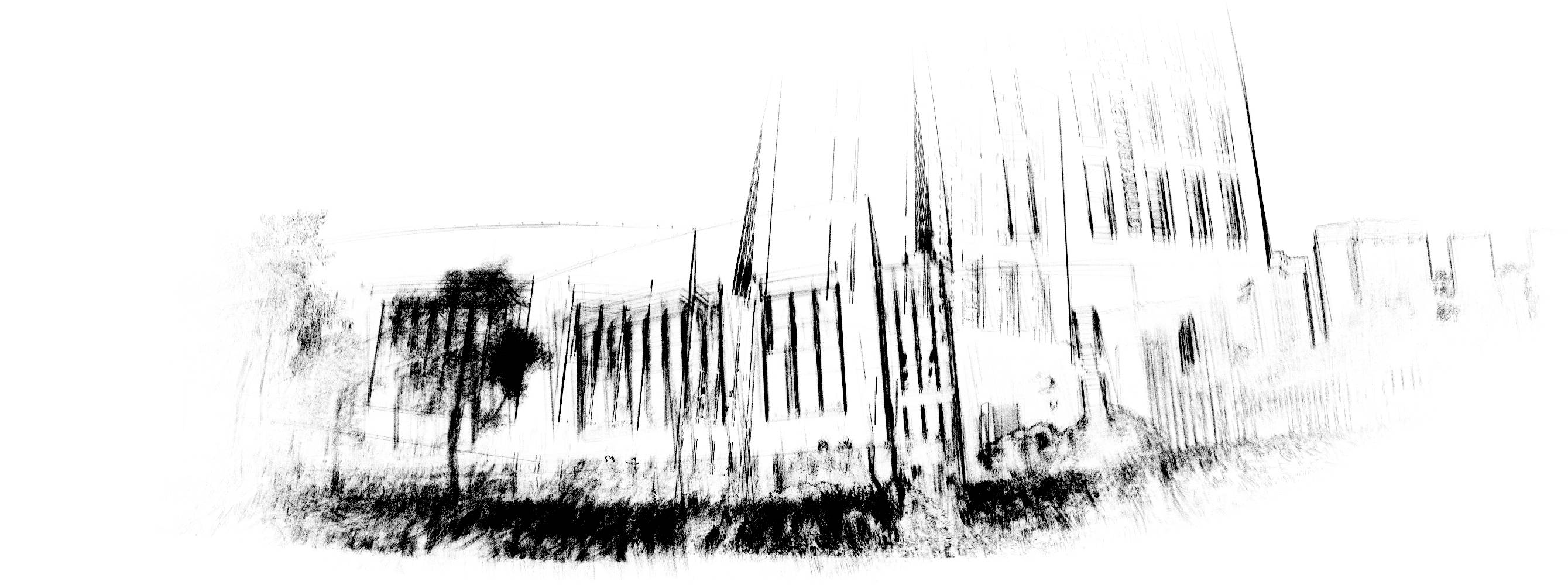}}} };
            \zoombox[magnification=4]{0.30,0.43}
            \zoombox[magnification=4]{0.40,0.48}
            \zoombox[magnification=4]{0.71,0.65}
        \end{tikzpicture}
        \label{fig:panoramic_front_gate_cmaxslam}%
    }%
    \hfill
    \subfloat[EROAM result on \texttt{front-gate} sequence]{%
        \begin{tikzpicture}[
            zoomboxarray,
            spymargin=0.1em,
            caption margin=2ex,
            zoomboxes below,
            zoomboxarray columns=3,
            zoomboxarray rows=1,
            connect zoomboxes,
            zoombox paths/.append style={thick, red}]
            \node [image node] { {\color{gray}\fbox{\includegraphics[width=0.47\textwidth,
            trim={0mm} {10mm} {0mm} {58mm},
            clip]{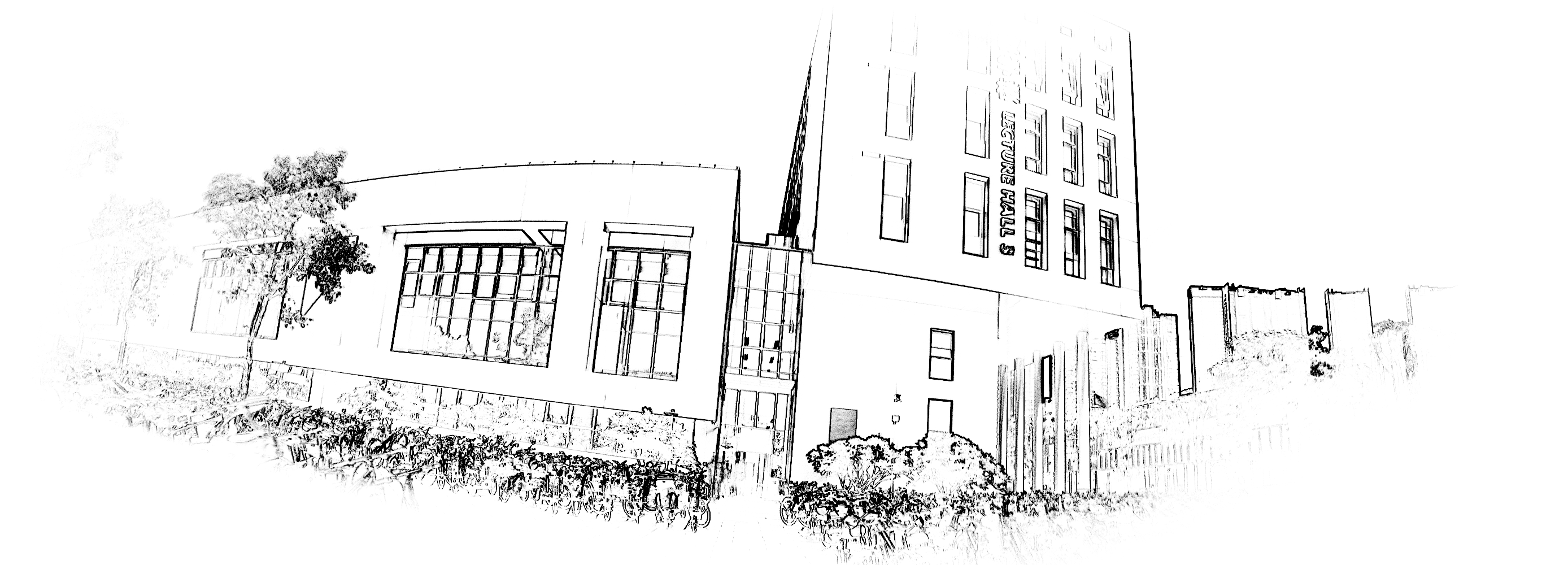}
            }} };
            \zoombox[magnification=4]{0.19,0.65}
            \zoombox[magnification=4]{0.36,0.62}
            \zoombox[magnification=4]{0.62,0.60}
        \end{tikzpicture}
        \label{fig:panoramic_front_gate_eroam}%
    }%
    
    
    \subfloat[\begin{revisedtextblockCR}CMax-SLAM result on \texttt{embankment-360} sequence\end{revisedtextblockCR}]{%
        {\color{gray}\fbox{\includegraphics[width=0.98\textwidth,
        trim={0mm} {90mm} {0mm} {80mm},
        clip]{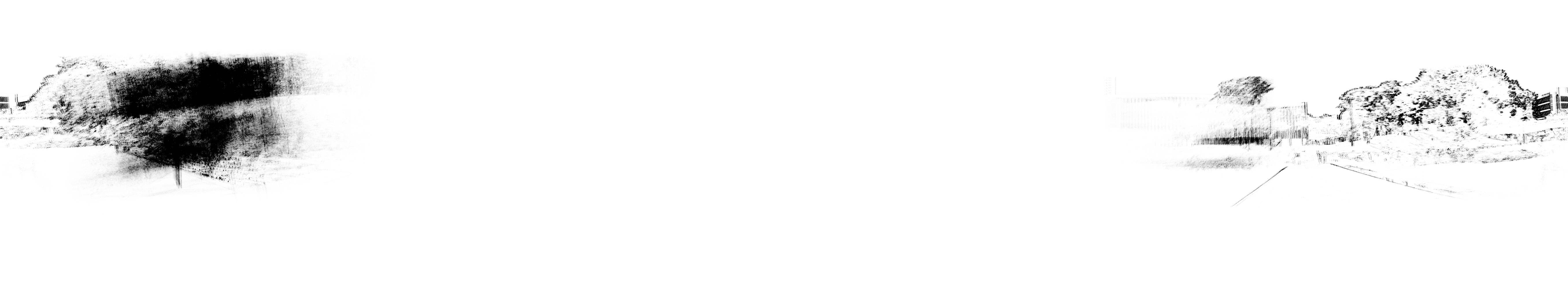}}}
        \label{fig:panoramic_embankment_cmaxslam}%
    }%
    
    \subfloat[EROAM result on \texttt{embankment-360} sequence]{%
        {\color{gray}\fbox{\includegraphics[width=0.98\textwidth,
        trim={0mm} {90mm} {0mm} {80mm},
        clip]{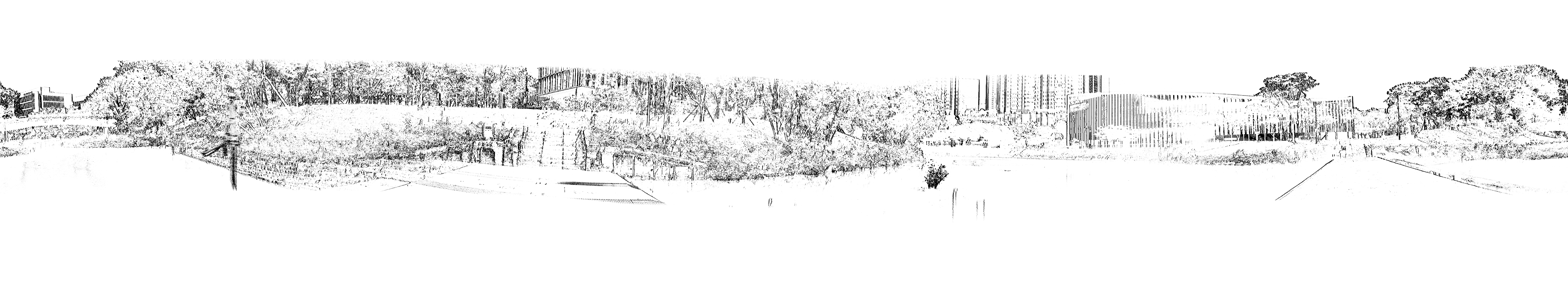}}}
        \label{fig:panoramic_embankment_eroam}%
    }%
    }
    \caption{Qualitative comparison of panoramic mapping results on real-world sequences. The zoomed-in regions in (a,b) highlight the superior structural clarity achieved by our method compared to CMax-SLAM's results with ghosting artifacts. The full \SI{360}{\degree} panoramas in (c,d) demonstrate EROAM's ability to maintain consistent reconstruction quality across extended views.}
    \label{fig:panoramic_comparison_real}
\end{figure*}

%% file: cr_figs_and_tabs/fig_eroam_campus_pano_more_visualization.tex
\begin{figure*}[!t]
    \centering
    \setlength{\fboxrule}{0.1pt}
    
    \subfloat[EROAM result on \texttt{distant-building} sequence]{%
        {\color{gray}\fbox{\includegraphics[width=0.48\textwidth,
        trim={79mm} {100mm} {79mm} {0mm},
        clip]{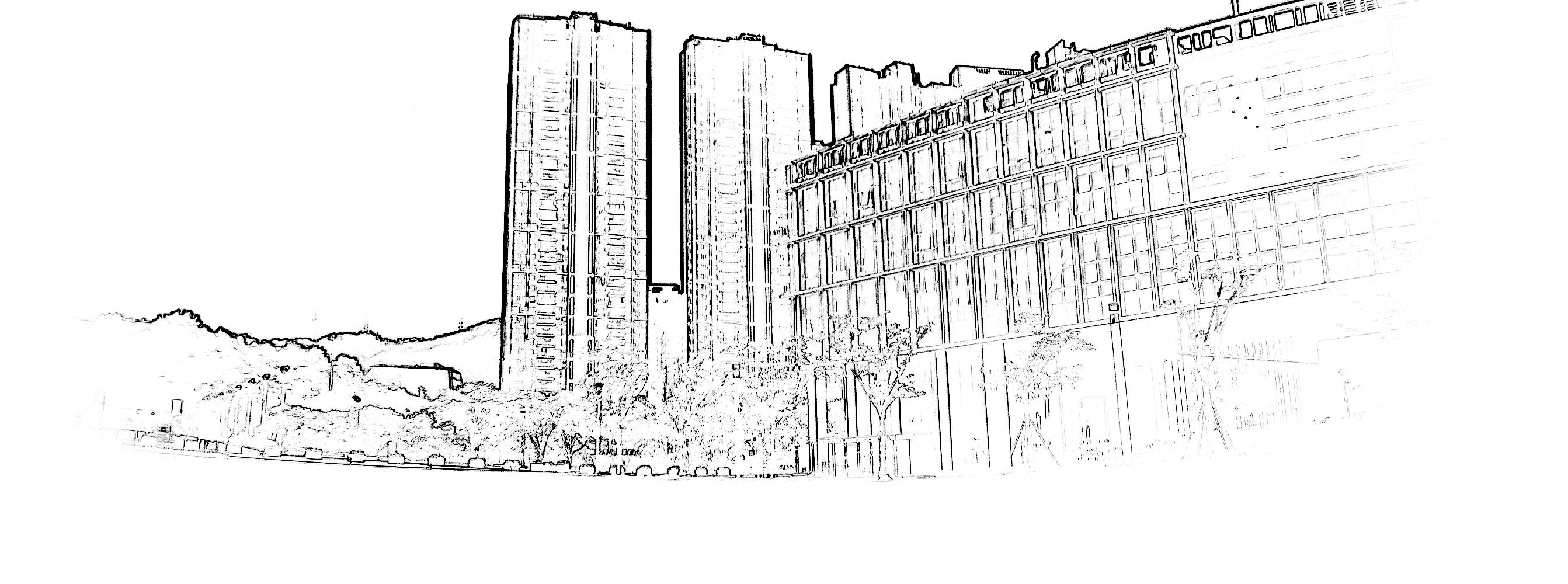}}}
        \label{fig:panoramic_distant_building}%
    }%
    \hfill
    \subfloat[EROAM result on \texttt{crossroad} sequence]{%
        {\color{gray}\fbox{\includegraphics[width=0.48\textwidth,
        trim={130mm} {47mm} {130mm} {40mm},
        clip]{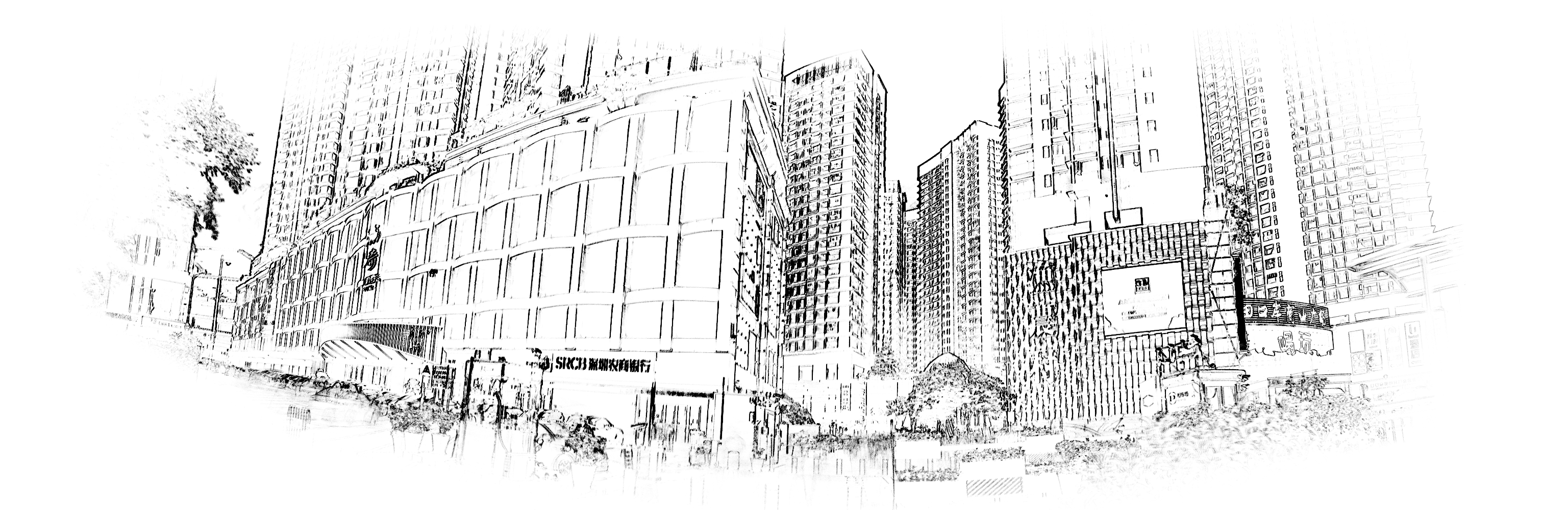}}}
        \label{fig:panoramic_crossroad}%
    }%
    
    \caption{Additional visualization results from EROAM on different sequences.}
    \label{fig:eroam_additional_results}
\end{figure*}

%% file: figs_and_tabs/tab_ecrot_runtime.tex
\begin{table*}[!t]
    \centering
    \caption{Runtime comparison on ECRot sequences.}
    \label{tab:runtime_comparison}
    \scalebox{0.80}{%
    \begin{tabular}{lcccrrrr}
        \toprule
        \multirow{2}{*}{Sequence} & \multirow{2}{*}{Duration (\si{\second})} & \multirow{2}{*}{Event Rate ($\times 10^6$\si{ev/s})} & \multirow{2}{*}{$\omega_{avg}$(\si{\degree/\second})} & \multicolumn{4}{c}{Runtime (\si{\second})} \\
        \cmidrule{5-8}
        & & & & RTPT \cite{rtpt} & CM-GAE \cite{cm-gae} & CMax-SLAM \cite{cmax-slam} & EROAM (\textbf{ours}) \\
        \midrule
        bay & 5.000 & 12.357 & 84.208 & 67.245 & 114.744 & 262.272 & \textbf{4.996} \\
        bicycle & 5.000 & 3.716 & 84.208 & 35.620 & 9.283 & 63.632 & \textbf{4.996} \\
        city & 5.000 & 15.026 & 84.208 & 79.172 & 151.906 & 336.170 & \textbf{4.995} \\
        street & 5.000 & 10.385 & 84.208 & 56.532 & 117.441 & 197.090 & \textbf{4.997} \\
        town & 5.000 & 11.074 & 84.208 & 62.541 & 68.969 & 194.032 & \textbf{4.996} \\
        playroom & 2.500 & 1.451 & 154.746 & 3.785 & \textbf{2.423} & 18.987 & 2.538 \\
        \bottomrule
    \end{tabular}%
    }
\end{table*}

%% file: figs_and_tabs/tab_eroam_campus_runtime.tex
\begin{table*}[!t]
    \centering
    \caption{Runtime comparison on EROAM-campus sequences.}
    \label{tab:runtime_comparison_real}
    \scalebox{0.80}{%
    \begin{tabular}{lcccrrrrr}
        \toprule
        \multirow{2}{*}{Sequence} & \multirow{2}{*}{Duration (\si{\second})} & \multirow{2}{*}{Event Rate ($\times 10^6$\si{ev/s})} & \multirow{2}{*}{$\omega_{avg}$(\si{\degree/\second})} & \multicolumn{4}{c}{Runtime (\si{\second})} \\
        \cmidrule{5-8}
        & & & & RTPT\textsuperscript{*} \cite{rtpt} & CM-GAE \cite{cm-gae} & CMax-SLAM \cite{cmax-slam} & EROAM (\textbf{ours}) \\
        \midrule
        canteen-entrance & 19.590 & 6.865 & 62.115 & 200.557 & 441.424 & 520.178 & \textbf{19.578} \\
        distant-building & 29.196 & 2.650 & 63.060 & 97.019 & 191.737 & 386.502 & \textbf{29.208} \\
        embankment & 31.413 & 3.473 & 64.870 & 166.443 & 196.604 & 552.528 & \textbf{31.449} \\
        embankment-360 & 18.101 & 6.233 & 97.792 & 174.606 & 181.257 & 782.102 & \textbf{18.125} \\
        lawn & 47.169 & 13.802 & 77.099 & - & 1318.581 & 2553.260 & \textbf{47.158} \\
        lawn-360 & 16.090 & 8.888 & 83.732 & 226.574 & 347.039 & 558.310 & \textbf{16.079} \\
        lecture-hall & 18.368 & 4.064 & 50.790 & 99.170 & 915.236 & 256.658 & \textbf{18.366} \\
        front-gate & 26.153 & 7.300 & 80.189 & 294.268 & 270.939 & 794.393 & \textbf{26.148} \\
        rooftop & 50.267 & 4.349 & 64.182 & 321.165 & 410.770 & 845.354 & \textbf{50.338} \\
        window-building & 27.514 & 9.304 & 54.567 & 335.886 & 877.523 & 1417.175 & \textbf{27.508} \\
        carpark & 38.878 & 9.751 & 77.642 & 486.052 & 780.133 & 1108.190 & \textbf{38.877} \\
        condo & 49.389 & 12.077 & 76.176 & - & 988.569 & 1663.687 & \textbf{49.370} \\
        crossroad & 44.512 & 11.171 & 74.227 & 664.389 & 820.304 & 1342.736 & \textbf{44.502} \\
        vehicle & 36.043 & 14.673 & 85.943 & 647.659 & 883.591 & 1203.566 & \textbf{36.032} \\
        limx-rotation & 22.090 & 16.188 & 106.683 & 479.243 & 606.214 & 695.631 & \textbf{21.988} \\
        limx-sway & 19.990 & 17.438 & 113.613 & 407.918 & 754.939 & 715.156 & \textbf{19.978} \\
        \bottomrule
    \end{tabular}%
    }

\begin{tablenotes}
\item 
\footnotesize{
\hspace{0.8cm}\textsuperscript{*}RTPT fails to process the \texttt{lawn} and \texttt{condo} sequences due to their large size.
}
\end{tablenotes}

\end{table*}

%% file: sections/experiments_ecd.tex
\subsection{Robustness to Unmodeled Translational Motion}
\label{subsec:ecd_robustness}
\input{cr_figs_and_tabs/fig_reviewer2_ecd_comparison_qualitative}
\input{cr_figs_and_tabs/tab_reviewer2_ecd_comparison_quantitative}

While EROAM is designed for pure rotational motion, some degree of translational motion is inevitable in practice. To evaluate robustness when this assumption is violated, we assess EROAM on the Event Camera Dataset (ECD)~\cite{mueggler2017event}, which features unconstrained handheld 6-DOF motion with translation typically within \SI{\sim0.3}{\m} and associated parallax effects, representing a more challenging scenario than our previous pure rotation evaluations. Following CMax-SLAM's evaluation protocol~\cite{cmax-slam}, we process all four sequences (\texttt{boxes}, \texttt{dynamic}, \texttt{poster}, and \texttt{shapes}), using the first 30 seconds of each.

\prettyref{fig:ecd_robustness_comparison} presents panoramic reconstructions from three sources. Critically, the rightmost column shows panoramas generated using only the rotational component of the 6-DOF ground truth. These are severely blurred and fail to register scene edges. This visualization confirms that when parallax from translation is present, even the ground truth rotation cannot properly align events, rendering it an invalid reference for performance evaluation.

In contrast, both EROAM (left column) and CMax-SLAM (middle column) produce coherent panoramic reconstructions, demonstrating that both methods can operate despite the presence of unmodeled translation. However, the reconstructed edges are noticeably less sharp compared to pure rotation scenarios (\prettyref{fig:ecrot_synthetic_panoramic_comparison}), which is expected since no single $SO(3)$ trajectory can perfectly align all events when parallax exists.

For transparency and completeness, we report the quantitative $\overline{ape}$ and $\overline{rpe}$ metrics in \prettyref{tab:ecd_evaluation}. However, it is critical to emphasize that $\overline{ape}$ and $\overline{rpe}$ are not effective metrics for performance evaluation on these sequences. The ground truth motion is 6-DOF, yet this evaluation is constrained to only 3-DOF rotation. As \prettyref{fig:ecd_robustness_comparison} visually confirms with the severely blurred GT (Rotation Only) panorama, the 3-DOF rotational component of the ground truth is itself an invalid reference that fails to align the events due to significant translational parallax. Therefore, the quantitative results in this context cannot measure accuracy, their purpose is to demonstrate that EROAM and CMax-SLAM exhibit comparable behavior when processing this unmodeled translational motion, which aligns with our qualitative findings.

This experiment demonstrates EROAM's robustness to moderate translational motion. However, when translation becomes substantially larger, the parallax effects overwhelm the pure rotation model, as discussed in \prettyref{sec:limitations}.

%% file: cr_figs_and_tabs/fig_reviewer2_ecd_comparison_qualitative.tex
\begin{figure}[!t]
    \centering
    \small
    \newcommand{\clipleft}{0mm}
    \newcommand{\clipbottom}{60mm}
    \newcommand{\clipright}{0mm}
    \newcommand{\cliptop}{10mm}
    
    \setlength{\fboxrule}{0.1pt}
    \renewcommand{\arraystretch}{0.5} 
    \setlength{\tabcolsep}{1pt}     
    
    \begin{tabular}{c c c c}
         & EROAM (Ours) & CMax-SLAM \cite{cmax-slam} & GT (Rotation Only) \\
         
        \rotatebox{90}{\hspace{1em}\texttt{boxes}} &
        \color{gray}\fbox{\includegraphics[width=0.29\columnwidth,
        trim={\clipleft} {\clipbottom} {\clipright} {\cliptop},
        clip]{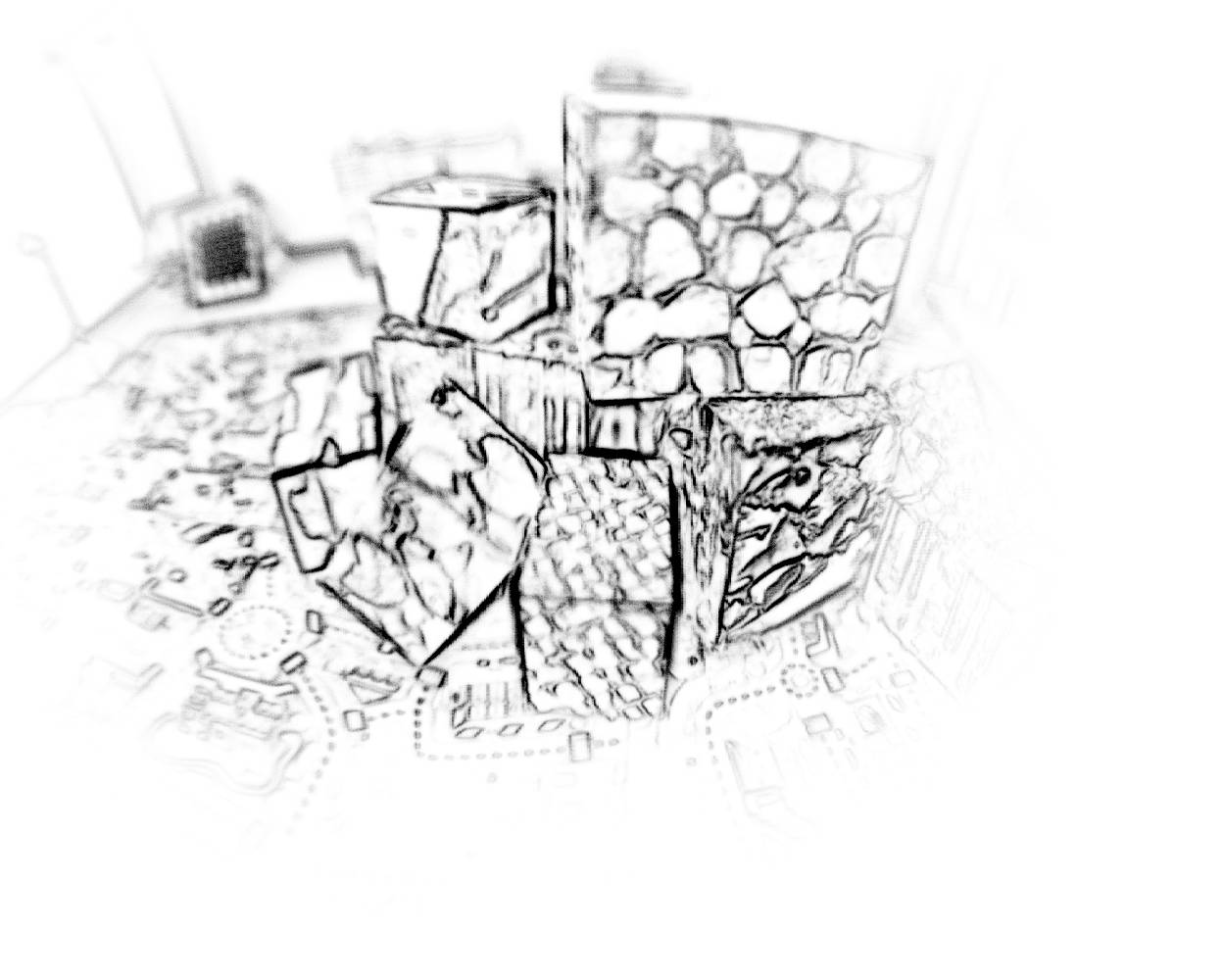}} &
        \color{gray}\fbox{\includegraphics[width=0.29\columnwidth,
        trim={\clipleft} {\clipbottom} {\clipright} {\cliptop},
        clip]{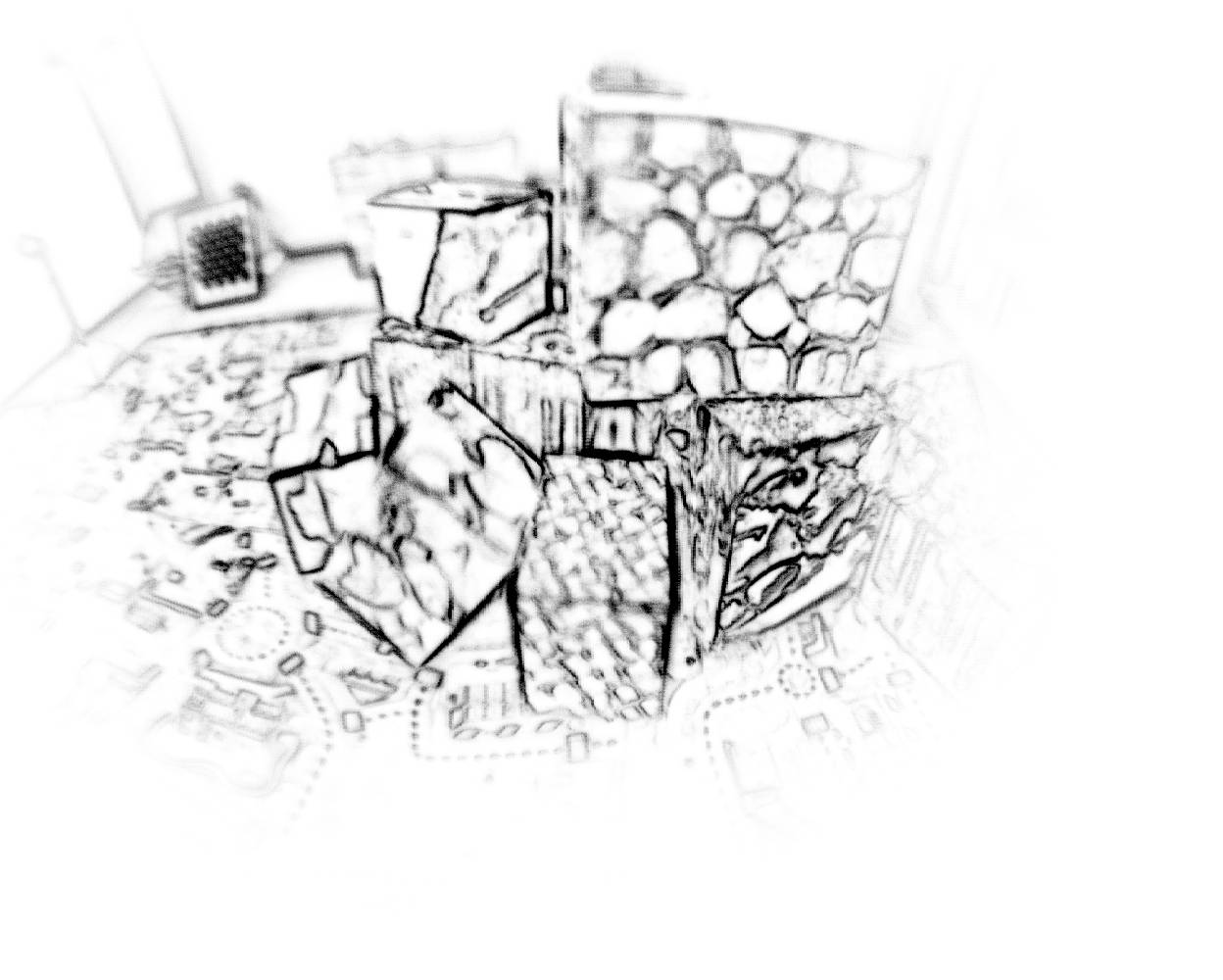}} &
        \color{gray}\fbox{\includegraphics[width=0.29\columnwidth,
        trim={\clipleft} {\clipbottom} {\clipright} {\cliptop},
        clip]{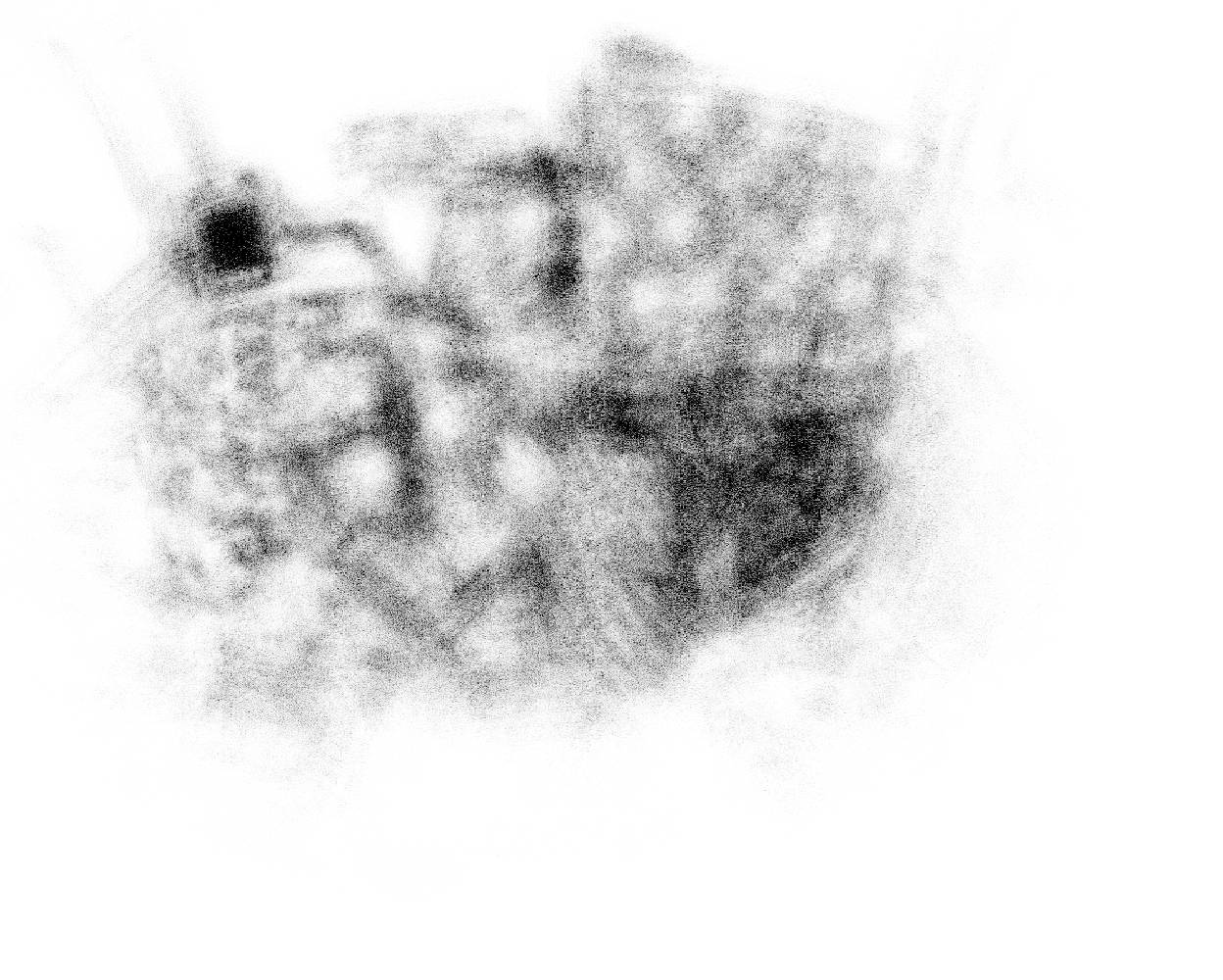}} \\
        
        \rotatebox{90}{\hspace{1em}\texttt{dynamic}} &
        \color{gray}\fbox{\includegraphics[width=0.29\columnwidth,
        trim={\clipleft} {\clipbottom} {\clipright} {\cliptop},
        clip]{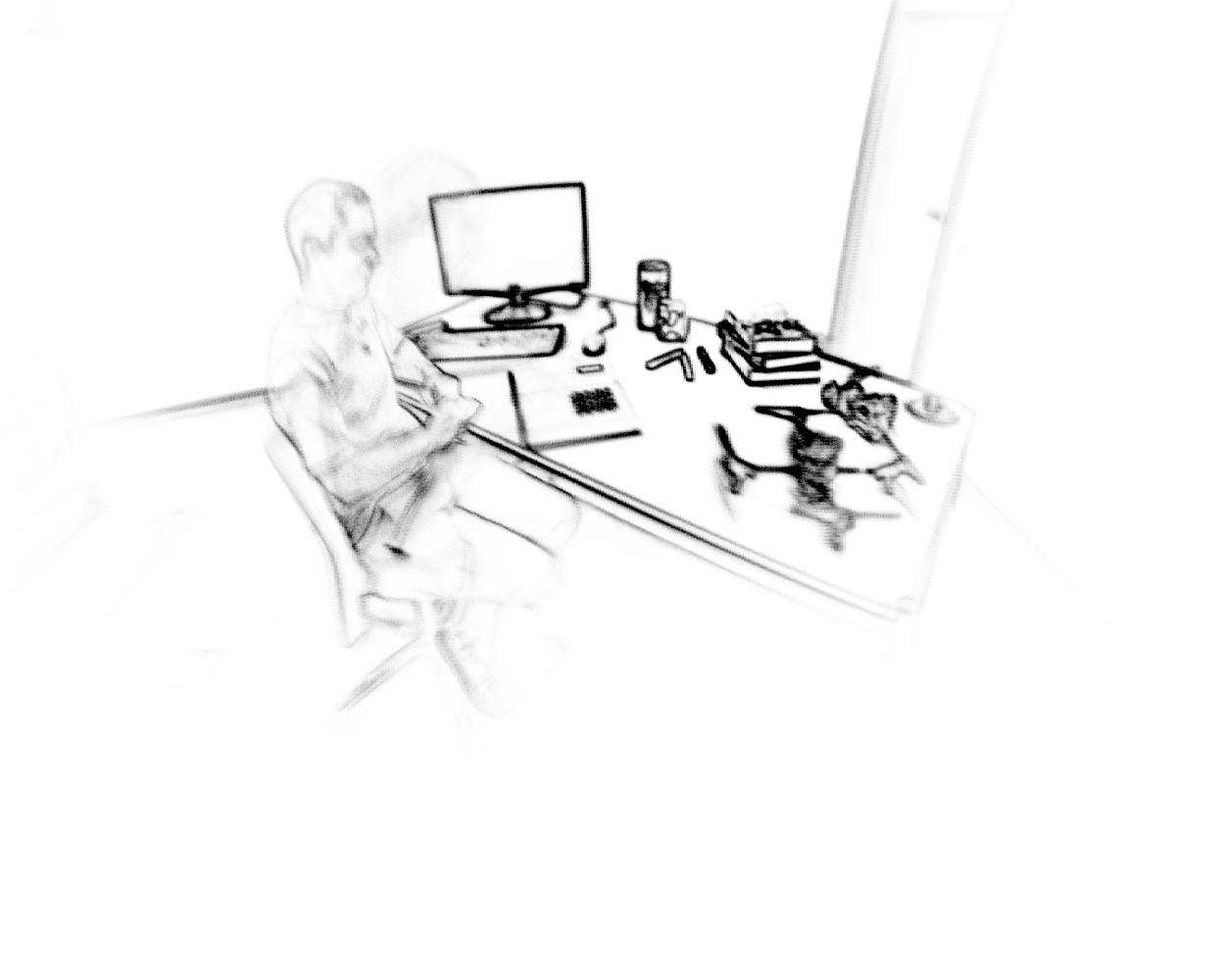}} &
        \color{gray}\fbox{\includegraphics[width=0.29\columnwidth,
        trim={\clipleft} {\clipbottom} {\clipright} {\cliptop},
        clip]{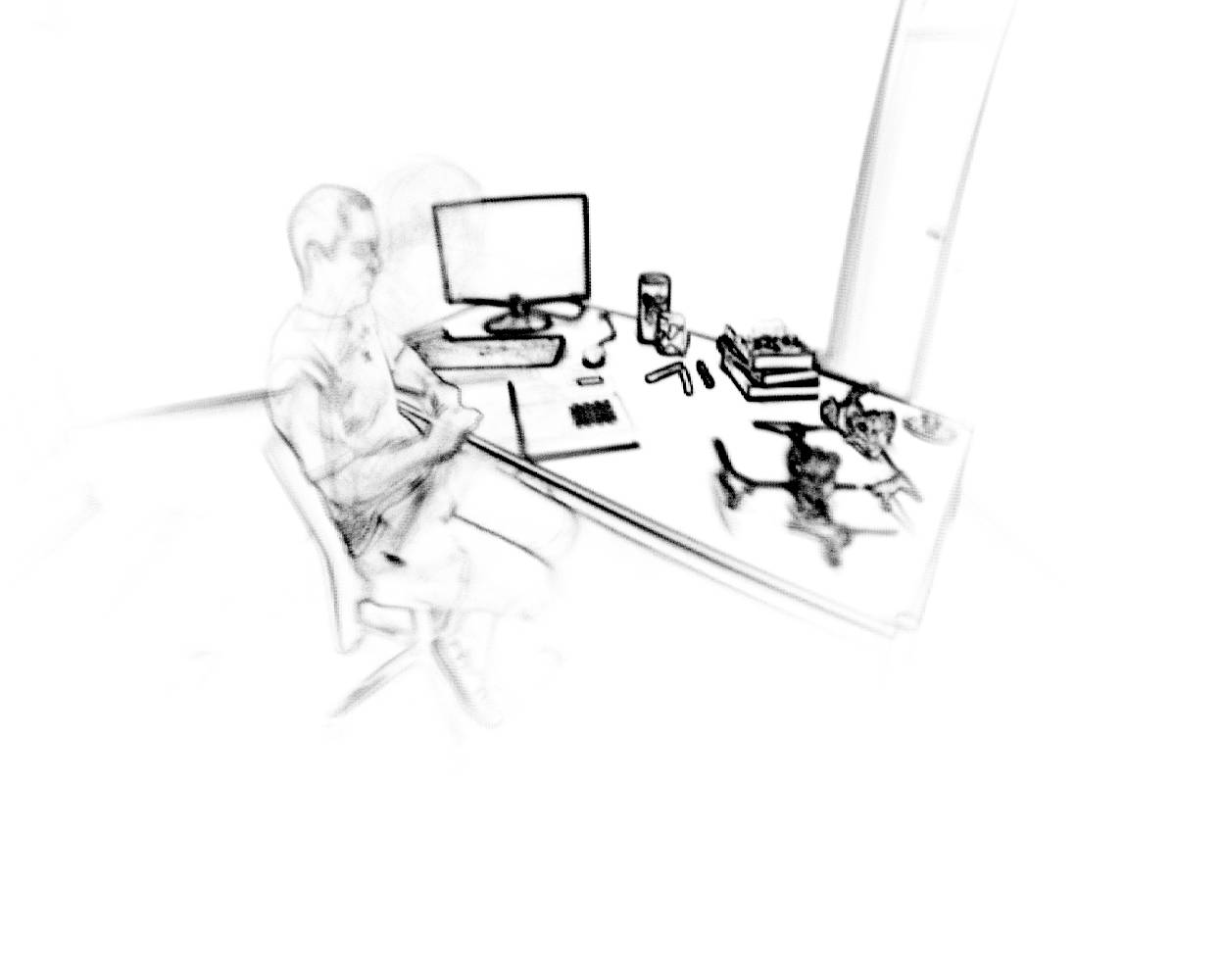}} &
        \color{gray}\fbox{\includegraphics[width=0.29\columnwidth,
        trim={\clipleft} {\clipbottom} {\clipright} {\cliptop},
        clip]{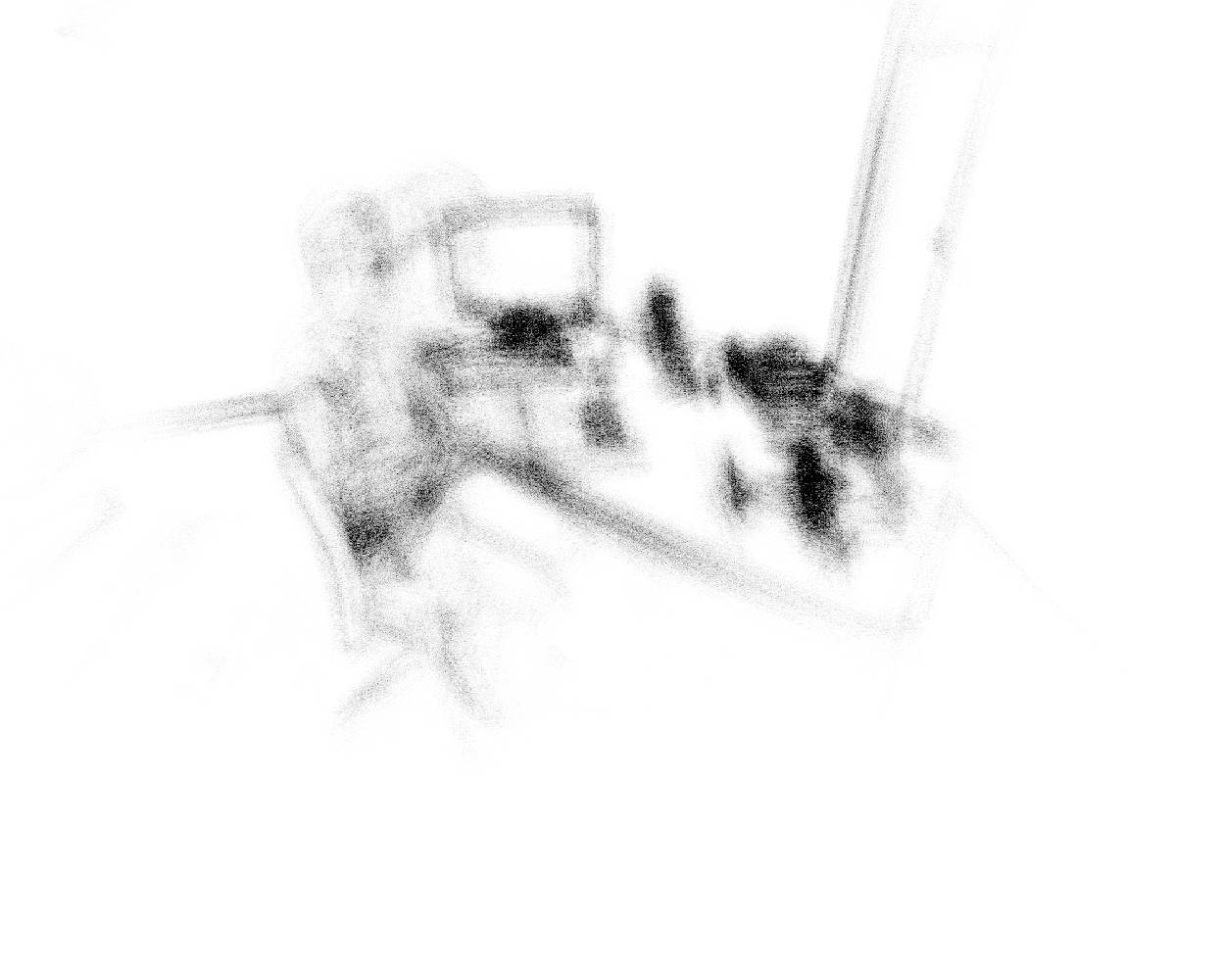}} \\
        
        \rotatebox{90}{\hspace{1em}\texttt{poster}} &
        \color{gray}\fbox{\includegraphics[width=0.29\columnwidth,
        trim={\clipleft} {\clipbottom} {\clipright} {\cliptop},
        clip]{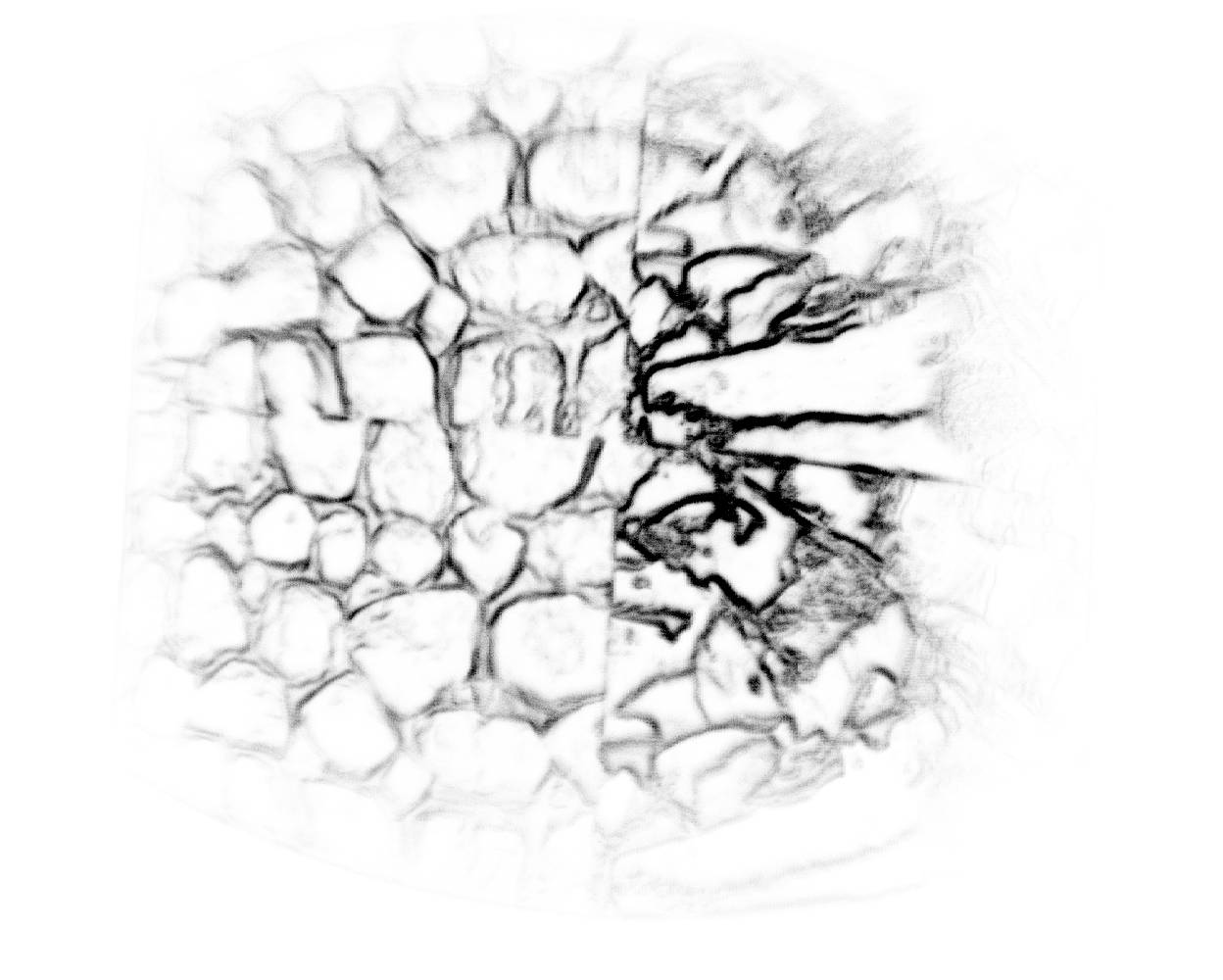}} &
        \color{gray}\fbox{\includegraphics[width=0.29\columnwidth,
        trim={\clipleft} {\clipbottom} {\clipright} {\cliptop},
        clip]{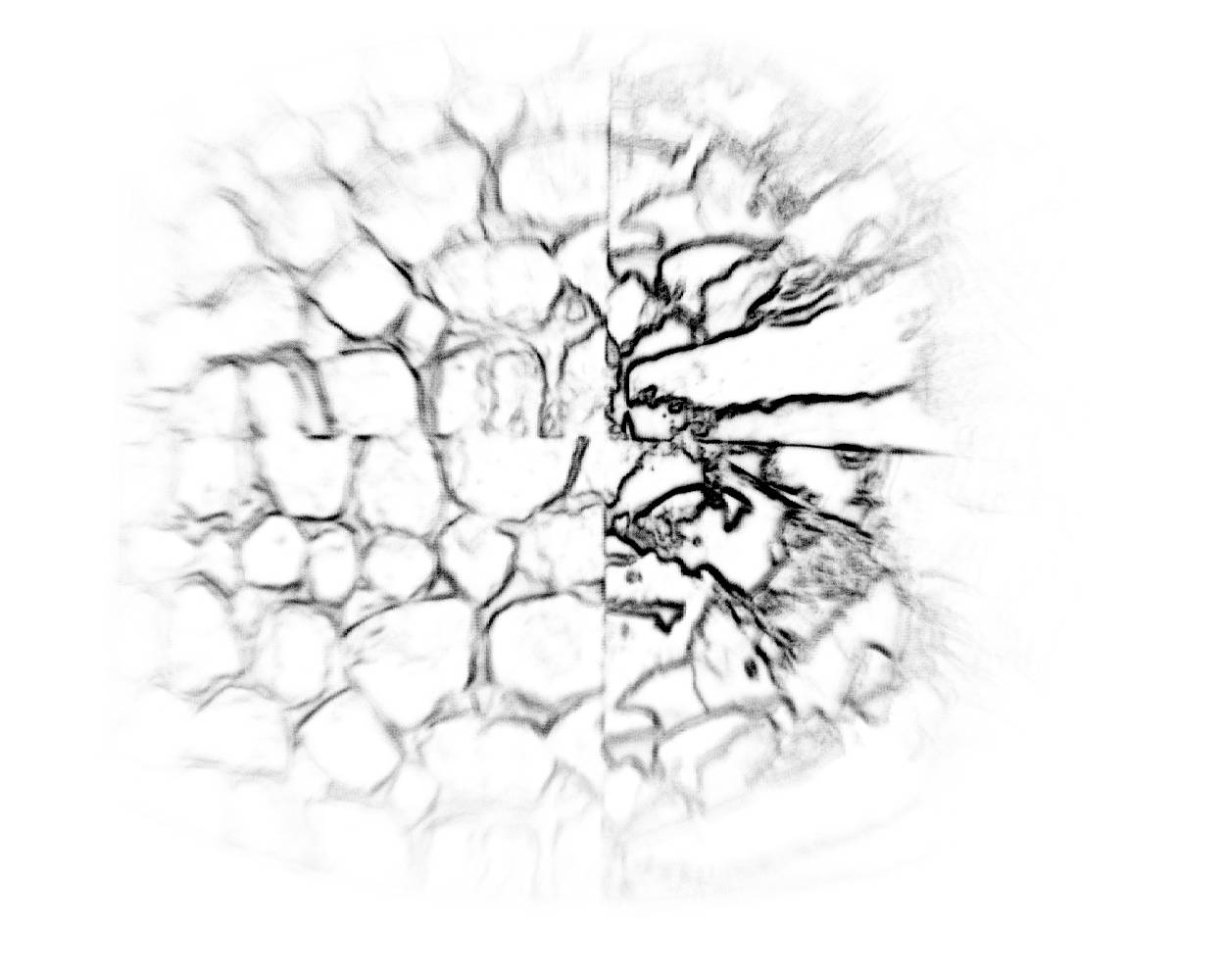}} &
        \color{gray}\fbox{\includegraphics[width=0.29\columnwidth,
        trim={\clipleft} {\clipbottom} {\clipright} {\cliptop},
        clip]{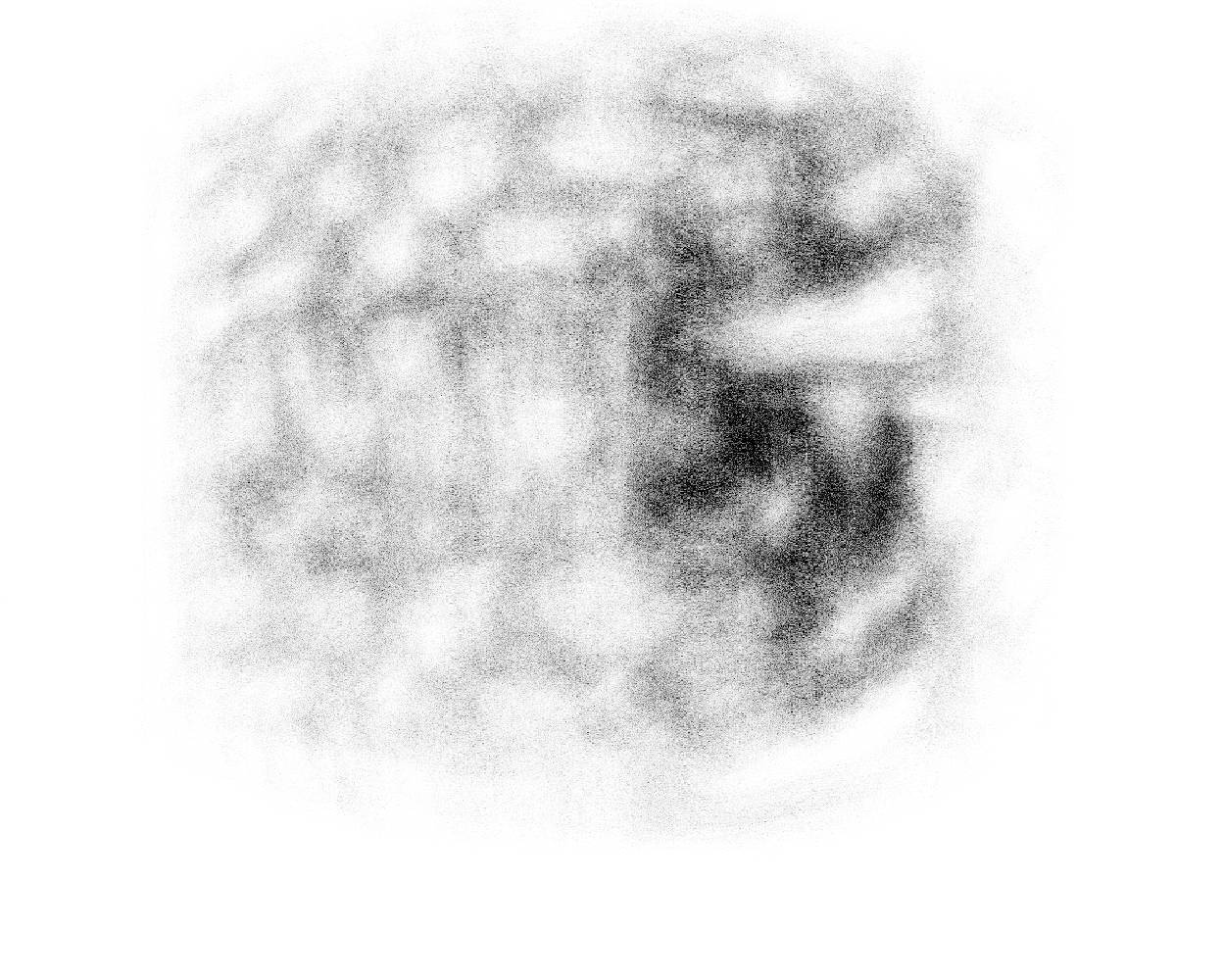}} \\
        
        \rotatebox{90}{\hspace{1em}\texttt{shapes}} &
        \color{gray}\fbox{\includegraphics[width=0.29\columnwidth,
        trim={\clipleft} {\clipbottom} {\clipright} {\cliptop},
        clip]{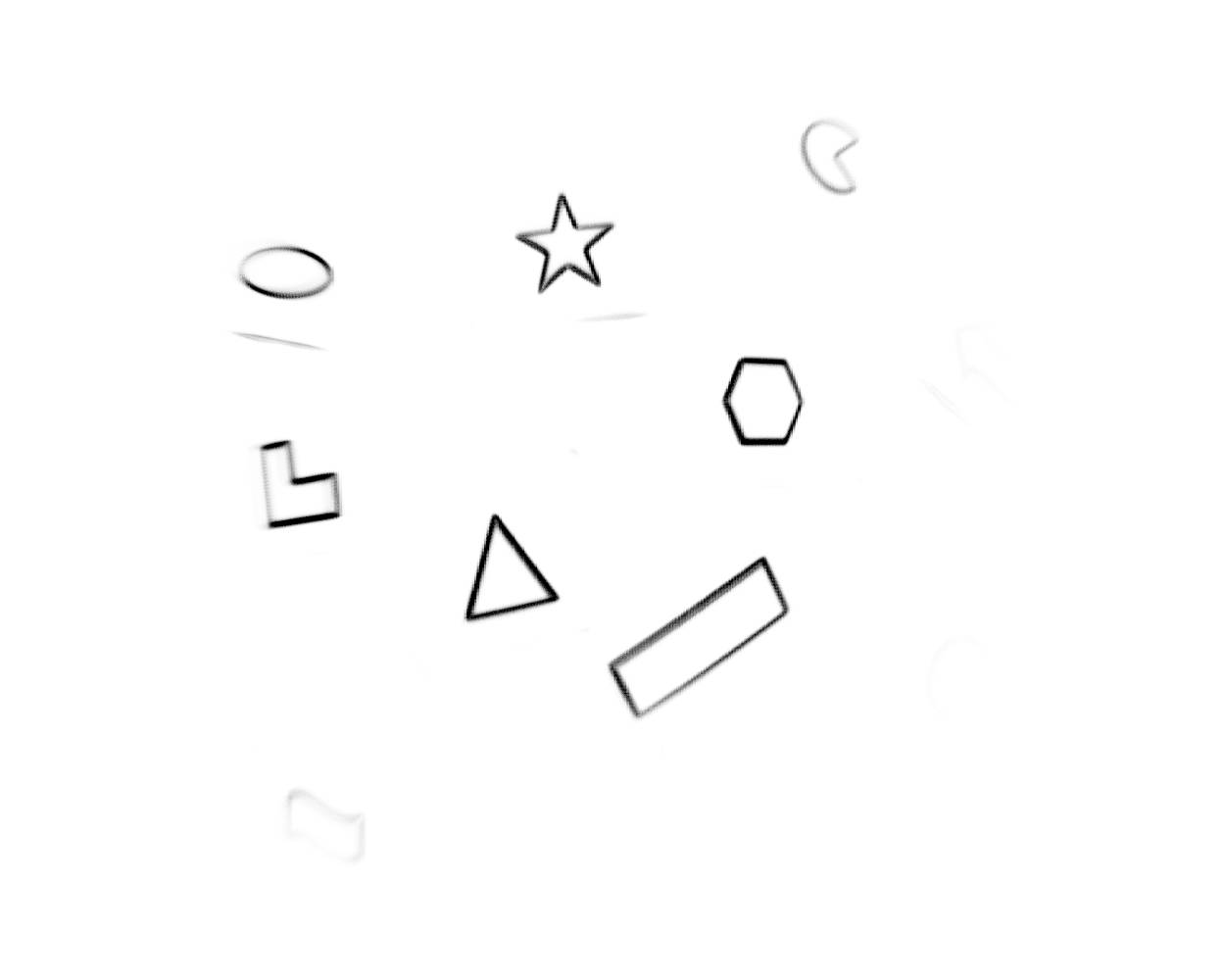}} &
        \color{gray}\fbox{\includegraphics[width=0.29\columnwidth,
        trim={\clipleft} {\clipbottom} {\clipright} {\cliptop},
        clip]{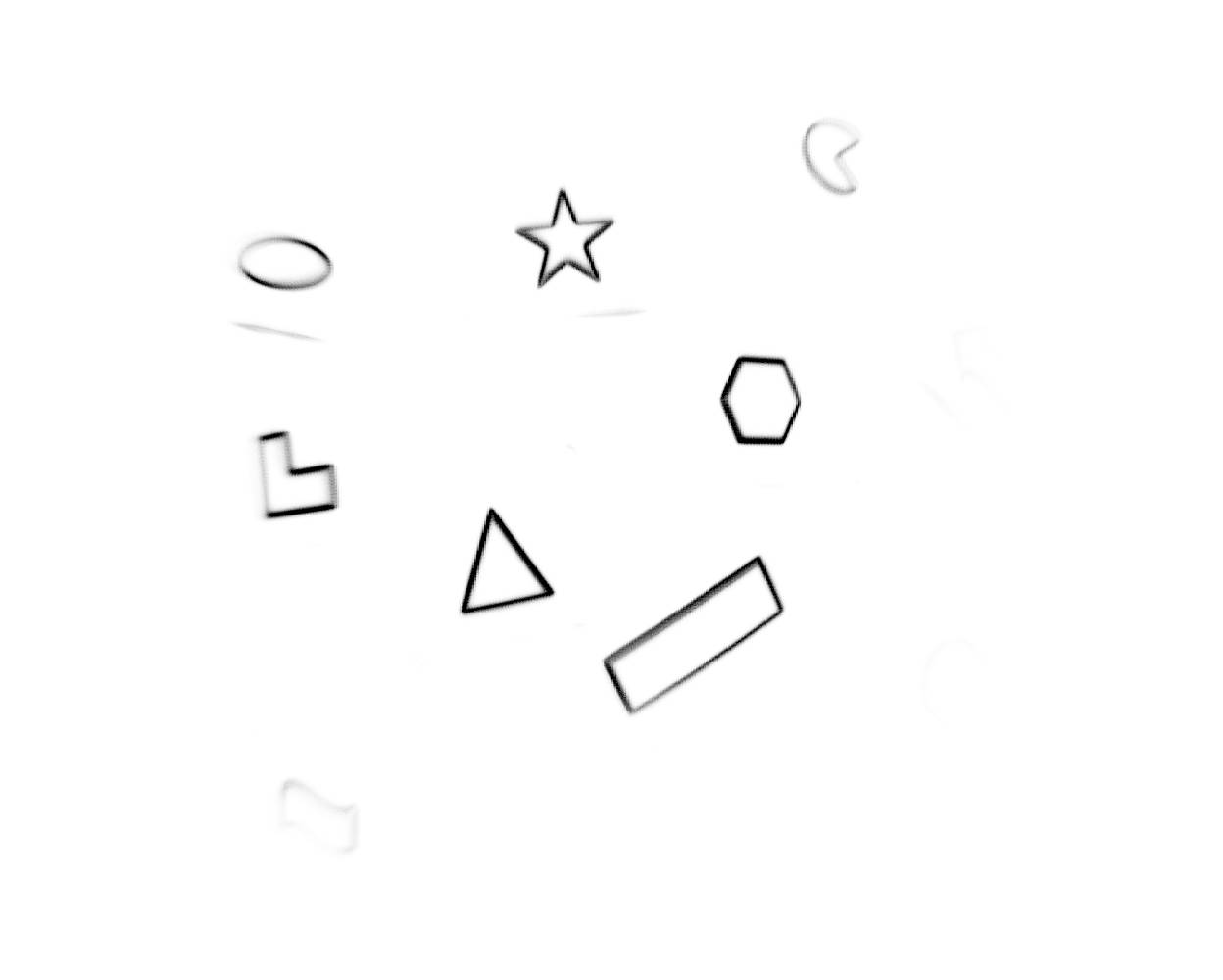}} &
        \color{gray}\fbox{\includegraphics[width=0.29\columnwidth,
        trim={\clipleft} {\clipbottom} {\clipright} {\cliptop},
        clip]{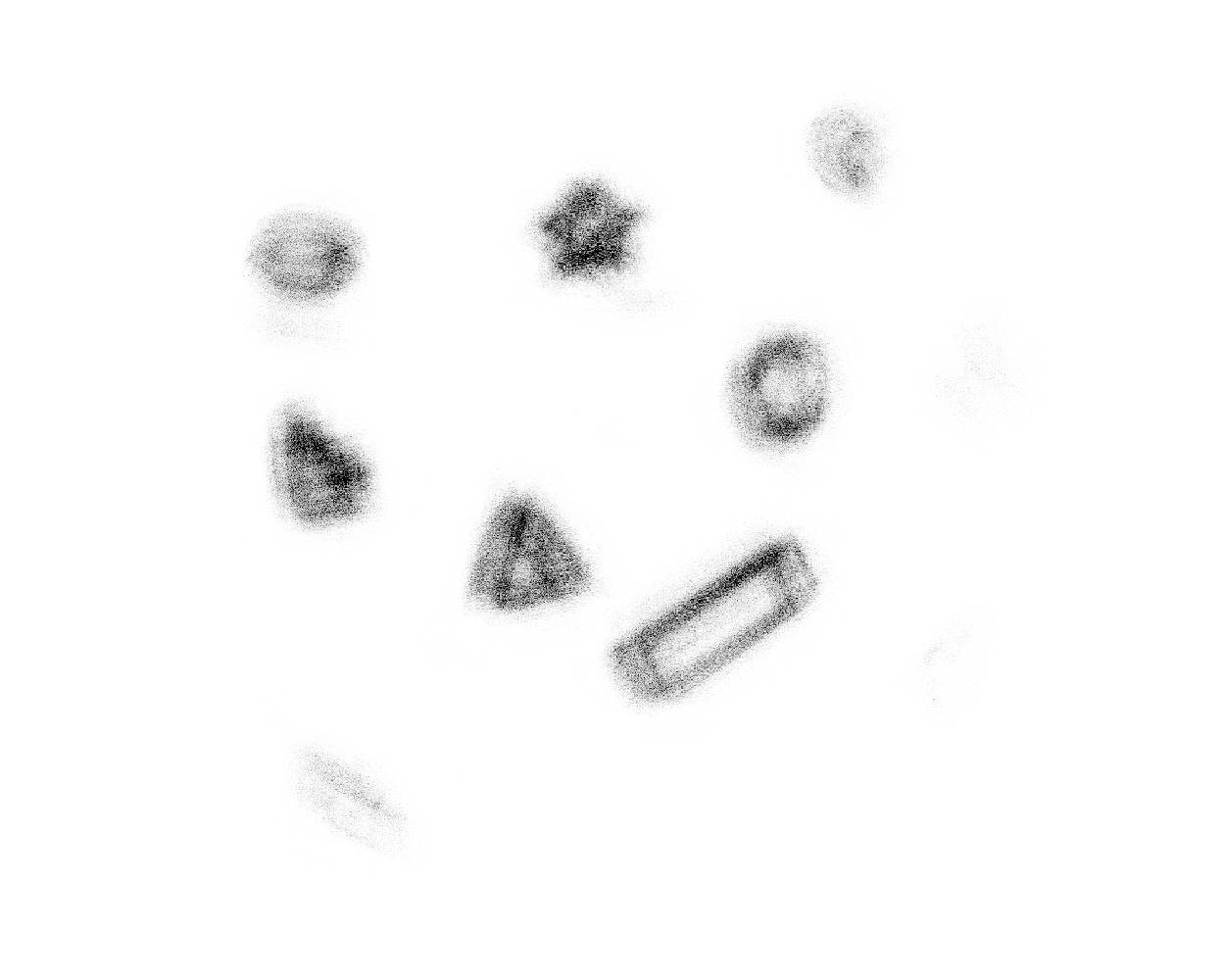}} \\
    \end{tabular}
    
    \caption{Robustness evaluation on ECD dataset \cite{mueggler2017event} with unmodeled translation. Left: EROAM. Middle:  CMax-SLAM \cite{cmax-slam}. Right: GT rotation-only panorama. The severe blurring in GT panoramas confirms that parallax from translation makes quantitative rotation metrics misleading. Both methods produce coherent results despite minor translation, though edges are less sharp than in pure rotation scenarios.}
    \label{fig:ecd_robustness_comparison}
\end{figure}

%% file: cr_figs_and_tabs/tab_reviewer2_ecd_comparison_quantitative.tex
\begin{table}[!t]
\centering
\caption{Quantitative evaluation on the ECD dataset \cite{mueggler2017event}.}
\label{tab:ecd_evaluation}
\resizebox{\columnwidth}{!}{%
\begin{tabular}{l*{8}{c}}
\toprule
Sequence & \multicolumn{2}{c}{boxes} & \multicolumn{2}{c}{dynamic} & \multicolumn{2}{c}{poster} & \multicolumn{2}{c}{shapes} \\
\cmidrule(lr){2-3} \cmidrule(lr){4-5} \cmidrule(lr){6-7} \cmidrule(lr){8-9}
  & $\overline{ape}$ (\si{\degree}) & $\overline{rpe}$ (\si{\degree}) & $\overline{ape}$ (\si{\degree}) & $\overline{rpe}$ (\si{\degree}) & $\overline{ape}$ (\si{\degree}) & $\overline{rpe}$ (\si{\degree}) & $\overline{ape}$ (\si{\degree}) & $\overline{rpe}$ (\si{\degree}) \\
\midrule
CMax-SLAM \cite{cmax-slam} & 2.752 & 1.251 & 1.758 & 0.890 & 4.720 & 1.384 & 2.964 & 1.471 \\
EROAM (\textbf{ours}) & 2.680 & 1.138 & 2.491 & 0.918 & 4.711 & 1.310 & 3.135 & 1.404 \\
\bottomrule
\end{tabular}%
}

\end{table}

%% file: sections/experiments_ablation.tex
\subsection{Ablation Studies}
\label{subsec:ablation_studies}
To analyze how different design choices and parameters affect EROAM's performance, we present comprehensive ablation studies. These studies specifically investigate the impact of four key elements: the number of events per frame ($n$), the processing frequency ($f$), the ES-ICP distance metric (point-to-line vs. point-to-point), and the Regional Density Management (RDM) mechanism.

\subsubsection{Impact of Number of Events per Frame ($n$)}
\input{figs_and_tabs/tab_ablation_n}
The number of events $n$ used per processing frame directly affects both the estimation accuracy and computational efficiency of our system. \prettyref{tab:ablation_n_parameter_extended_ms} presents results across various settings of $n$, evaluated on both synthetic ECRot and real-world EROAM-campus sequences.

The results indicate that a sufficient number of events is crucial for the ES-ICP algorithm to reliably establish geometric constraints and converge to an accurate solution. With very small event slices ($n \leq 400$), ES-ICP struggles, leading to high error rates or complete tracking failures. For instance, with $n=200$, all tested sequences show $\overline{ape}$ exceeding \SI{110}{\degree}, indicating complete tracking loss. Performance stabilizes significantly when $n \geq 800$, with $\overline{ape}$ dropping to \SI{0.140}{\degree} for the \texttt{bicycle} sequence. For our main experiments, we selected $n=1500$. This choice (approximated by $n=1600$ in the ablation table) effectively balances accuracy and computational load. At $n=1600$, the ES-ICP computation time ($t_{es}$) consistently remains between \SI{0.62}{\ms} and \SI{0.72}{\ms}, well within the \SI{1}{\ms} budget required for \SI{1000}{\Hz} operation. While further increasing $n$ (to 3200, 6400, or 12800) might offer marginal accuracy improvements in some sequences, it significantly increases computation time. For instance, $t_{es}$ exceeds \SI{1}{\ms} when $n=3200$, reaching as high as \SI{3.288}{\ms} for the \texttt{town} sequence when $n=12800$. Thus, $n=1500$ provides a practical trade-off, ensuring robust performance without compromising real-time capabilities.

\subsubsection{Effect of Processing Frequency ($f$)}
\input{figs_and_tabs/tab_ablation_f}
Processing frequency $f$ determines how often rotation estimates are computed and directly impacts the system's ability to track high-speed motions. \prettyref{tab:ablation_freq_f_parameter} shows performance across different frequency settings, with a fixed $n=1500$.

High-frequency estimation is paramount for EROAM's success. The Iterative Closest Point (ICP) algorithm, which ES-ICP is based on, is susceptible to converging to local minima if the initial guess for the transformation is not sufficiently close to the true solution. By operating at a high frequency (e.g., \SI{1000}{\Hz}), the actual rotation between consecutive frames becomes very small. When we use the solution from the previous ES-ICP step as the initial guess for the current frame, this initial guess is inherently very close to the true current pose. This dramatically improves the likelihood of ES-ICP converging to the correct global minimum, which is a core reason for EROAM's high accuracy.

The experimental results support this understanding. At very low frequencies ($f = \SI{50}{\Hz}$), large inter-frame rotations lead to significant tracking challenges, resulting in high errors or failures (e.g., $\overline{ape} = \SI{94.866}{\degree}$ for \texttt{bicycle} at $f = \SI{50}{\Hz}$). Performance improves substantially as frequency increases, with $f \geq \SI{400}{\Hz}$ providing consistently good results. At $f = \SI{800}{\Hz}$, all sequences demonstrate excellent accuracy, with ES-ICP computation times ($t_{es}$) under \SI{0.9}{\ms}.

Conversely, extremely high frequencies ($f \geq \SI{1600}{\Hz}$), while maintaining good accuracy on some sequences, can lead to diminishing returns or even slight degradation. More importantly, they substantially increase total processing time due to the overhead of handling many more frames (e.g., processing time for \texttt{town} increases from \SI{4.979}{\s} at $f = \SI{800}{\Hz}$ to \SI{11.281}{\s} at $f = \SI{3200}{\Hz}$).

Based on these findings, we selected $f = \SI{1000}{\Hz}$ as our standard operational frequency. This allows EROAM to process a relatively small number of events per frame but at an extremely high rate, offering an optimal balance between accuracy, robustness to dynamic motion, and computational efficiency.

\subsubsection{Point-to-Line vs. Point-to-Point ES-ICP}
\input{figs_and_tabs/tab_ablation_esicp_and_rdm}

\input{figs_and_tabs/fig_reviewer9_p2p_vs_p2l}
To assess the impact of the ES-ICP distance metric, we compared our point-to-line (P2L) approach with a traditional point-to-point (P2P) implementation. The results are presented in \prettyref{tab:p2l_vs_p2p}, with a qualitative comparison shown in \prettyref{fig:p2l_vs_p2p_panorama_comparison}.

The P2L approach consistently outperforms P2P in terms of accuracy across all datasets. This advantage is crucial for robust tracking, as an erroneous estimate from P2P could negatively affect subsequent frames by providing a poor initial guess. The accuracy gap becomes particularly significant in challenging sequences. For example, on the \texttt{distant-building} sequence, P2L achieves an $\overline{ape}$ of \SI{0.274}{\degree} compared to P2P's \SI{1.938}{\degree}. The difference is even more dramatic for the dynamic \texttt{DL-8} sequence, where P2L maintains accurate tracking ($\overline{ape} = \SI{0.460}{\degree}$) while P2P fails completely ($\overline{ape} = \SI{17.159}{\degree}$). As demonstrated by comparing our P2L result in \prettyref{fig:panoramic_distant_building} with the P2P result in \prettyref{fig:p2l_vs_p2p_panorama_comparison}, P2L's superior accuracy translates to clearer panoramic reconstructions with well-aligned edges, whereas P2P results in noticeable ghosting and misalignments. This performance advantage stems from P2L's ability to leverage local structural information by fitting lines to neighboring points rather than relying solely on individual point correspondences. By considering the underlying edge structure represented by multiple map points, P2L establishes more robust geometric constraints for alignment, especially in regions with sparse or uneven event distributions.
While P2P offers computational advantages, with ES-ICP computation times approximately 20-30\% faster than P2L due to simpler nearest neighbor operations and the elimination of line fitting, the substantial accuracy and robustness improvements provided by P2L, particularly in challenging scenarios, justify its selection as our standard approach. Furthermore, P2L itself remains computationally efficient. For applications where raw speed is paramount over accuracy in simpler scenes, our open-source implementation also includes the P2P option.

\subsubsection{Effectiveness of Regional Density Management}
\input{figs_and_tabs/fig_reviewer9_rdm_time_pie}
Our Regional Density Management (RDM) approach is crucial for maintaining both long-term performance and map quality. \prettyref{tab:rdm_comparison} demonstrates its effectiveness by comparing EROAM's performance with and without RDM, and \prettyref{fig:rdm_runtime_comparison} illustrates the impact on computational resource allocation.

The core challenge RDM addresses is map maintenance for real-time operation without sacrificing detail through spatial downsampling. As keyframes are continuously added to the map (stored in an ikd-Tree), two issues arise: the time to update the ikd-Tree increases, and the growing number of points in the tree slows down nearest neighbor searches crucial for ES-ICP. This is especially problematic in long sequences or scenarios where the camera repeatedly scans the same area, leading to redundant points that bloat the map, degrade performance, and potentially reduce ES-ICP accuracy.
RDM effectively mitigates these issues. While both approaches (with and without RDM) achieve comparable accuracy (similar $\overline{ape}$ and $\overline{rpe}$ values), RDM significantly improves computational efficiency. This is most evident in longer sequences like \texttt{LT-80}, where RDM reduces the k-d tree update time ($t_{kd}$) by 84\%, from \SI{8930.521}{\ms} to \SI{1410.348}{\ms}.
\prettyref{fig:rdm_runtime_comparison} further clarifies RDM's optimization of computational resources. Without RDM, ikd-Tree operations consume 11.0\% of the total processing time. With RDM enabled, this overhead drops dramatically to just 1.7\%. The ES-ICP time ($t_{es}$) also sees a slight decrease with RDM, as a more efficiently managed map with fewer redundant points accelerates the nearest neighbor searches within ES-ICP. Notably, the \texttt{Waiting Time} (system idle time) increases from 10.4\% to 23.3\% with RDM, indicating that the system processes frames faster and has more performance headroom.
These results validate our decision to incorporate RDM. It enables efficient map management by controlling point density without the accuracy loss associated with downsampling, ensuring detail preservation across all observed regions and robust real-time performance, especially during extended operation.

%% file: figs_and_tabs/tab_ablation_n.tex
\begin{table*}[!t]
\centering
\caption{Ablation study on the impact of $n$}
\label{tab:ablation_n_parameter_extended_ms}
\resizebox{1.0\textwidth}{!}{%
\begin{tabular}{c|cccc|cccc|cccc|cccc}
\toprule
\multirow{2}{*}{$n$} & \multicolumn{4}{c|}{bicycle} & \multicolumn{4}{c|}{town} & \multicolumn{4}{c|}{distant-building} & \multicolumn{4}{c}{embankment} \\
\cmidrule{2-17}
 & $\overline{ape}$ (\si{\degree}) & $\overline{rpe}$ (\si{\degree}) & $t_{es}$ (\si{ms}) & $t$ (\si{s}) & $\overline{ape}$ (\si{\degree}) & $\overline{rpe}$ (\si{\degree}) & $t_{es}$ (\si{ms}) & $t$ (\si{s}) & $\overline{ape}$ (\si{\degree}) & $\overline{rpe}$ (\si{\degree}) & $t_{es}$ (\si{ms}) & $t$ (\si{s}) & $\overline{ape}$ (\si{\degree}) & $\overline{rpe}$ (\si{\degree}) & $t_{es}$ (\si{ms}) & $t$ (\si{s}) \\
\midrule
200   & 116.988 & 10.271 & 0.409 & 4.996 & 123.442 & 10.226 & 0.396 & 4.997 & 135.360 & 7.353 & 0.414 & 29.218 & 110.486 & 6.978 & 0.428 & 31.459 \\
400   & 0.986 & 0.191 & 0.566 & 4.998 & 95.452 & 8.461 & 0.498 & 4.996 & 89.487 & 7.562 & 0.519 & 29.218 & 88.200 & 6.529 & 0.497 & 31.459 \\
800   & 0.140 & 0.074 & 0.621 & 4.997 & 0.477 & 0.088 & 0.609 & 4.996 & 0.593 & 0.472 & 0.617 & 29.218 & 0.755 & 0.303 & 0.610 & 31.459 \\
1600  & 0.121 & 0.038 & 0.693 & 5.003 & 0.234 & 0.053 & 0.718 & 4.997 & 0.231 & 0.234 & 0.625 & 29.218 & 0.558 & 0.236 & 0.692 & 31.459 \\
3200  & 0.133 & 0.044 & 1.114 & 7.141 & 0.205 & 0.037 & 1.229 & 7.741 & 0.233 & 0.206 & 0.936 & 29.218 & 0.507 & 0.258 & 1.009 & 31.973 \\
6400  & 0.124 & 0.039 & 1.339 & 8.442 & 0.162 & 0.041 & 2.183 & 13.647 & 0.246 & 0.224 & 1.405 & 34.069 & 0.528 & 0.280 & 1.603 & 45.870 \\
12800 & 0.123 & 0.039 & 1.329 & 8.339 & 0.159 & 0.048 & 3.288 & 20.384 & 0.236 & 0.215 & 1.763 & 39.575 & 0.529 & 0.247 & 1.815 & 50.818 \\
\bottomrule
\end{tabular}%
}
\end{table*}

%% file: figs_and_tabs/tab_ablation_f.tex
\begin{table*}[!t]
\centering
\caption{Ablation study on the impact of frequency $f$}
\label{tab:ablation_freq_f_parameter}
\resizebox{1.0\textwidth}{!}{%
\begin{tabular}{c|cccc|cccc|cccc|cccc}
\toprule
\multirow{2}{*}{$f$ (\si{Hz})} & \multicolumn{4}{c|}{bicycle} & \multicolumn{4}{c|}{town} & \multicolumn{4}{c|}{distant-building} & \multicolumn{4}{c}{embankment} \\
\cmidrule{2-17}
 & $\overline{ape}$ (\si{\degree}) & $\overline{rpe}$ (\si{\degree}) & $t_{es}$ (\si{ms}) & $t$ (\si{s}) & $\overline{ape}$ (\si{\degree}) & $\overline{rpe}$ (\si{\degree}) & $t_{es}$ (\si{ms}) & $t$ (\si{s}) & $\overline{ape}$ (\si{\degree}) & $\overline{rpe}$ (\si{\degree}) & $t_{es}$ (\si{ms}) & $t$ (\si{s}) & $\overline{ape}$ (\si{\degree}) & $\overline{rpe}$ (\si{\degree}) & $t_{es}$ (\si{ms}) & $t$ (\si{s}) \\
\midrule
50   & 94.866 & 52.260 & 1.630 & 4.966 & 100.675 & 34.735 & 1.551 & 4.954 & 138.721 & 11.231 & 1.603 & 29.189 & 11.963 & 4.581 & 1.405 & 31.449 \\
100  & 18.819 & 0.608 & 2.141 & 4.977 & 65.508 & 3.022 & 1.869 & 4.974 & 0.258 & 0.196 & 1.731 & 29.199 & 0.693 & 0.272 & 1.744 & 31.449 \\
200  & 1.001 & 0.076 & 1.922 & 4.983 & 19.086 & 0.580 & 1.683 & 4.978 & 0.248 & 0.251 & 1.398 & 29.204 & 0.490 & 0.292 & 1.617 & 31.449 \\
400  & 0.071 & 0.042 & 1.301 & 4.983 & 2.376 & 0.131 & 1.240 & 4.992 & 0.224 & 0.194 & 1.200 & 29.207 & 0.444 & 0.216 & 1.230 & 31.449 \\
800  & 0.091 & 0.045 & 0.831 & 4.986 & 0.193 & 0.063 & 0.823 & 4.979 & 0.255 & 0.276 & 0.810 & 29.208 & 0.439 & 0.295 & 0.800 & 31.450 \\
1600 & 0.110 & 0.044 & 0.568 & 5.988 & 0.169 & 0.057 & 0.564 & 5.946 & 4.621 & 0.321 & 0.483 & 29.208 & 0.486 & 0.212 & 0.495 & 31.450 \\
3200 & 0.206 & 0.050 & 0.419 & 8.920 & 0.205 & 0.066 & 0.539 & 11.281 & 7.973 & 0.325 & 0.422 & 32.730 & 0.481 & 0.266 & 0.469 & 43.780 \\
\bottomrule
\end{tabular}%
}
\end{table*}

%% file: figs_and_tabs/tab_ablation_esicp_and_rdm.tex
\begin{table*}[!t]
\centering
\begin{minipage}{0.44\textwidth}
  \centering
  \caption{Comparison between Point-to-Line (P2L) and Point-to-Point (P2P) methods}
  \label{tab:p2l_vs_p2p}
  \resizebox{\textwidth}{!}{%
    \begin{tabular}{c|cccc|cccc}
    \toprule
    \multirow{2}{*}{Dataset} & \multicolumn{4}{c|}{P2L} & \multicolumn{4}{c}{P2P} \\
    \cmidrule{2-9}
     & $\overline{ape}$ (\si{\degree}) & $\overline{rpe}$ (\si{\degree}) & $t_{es}$ (\si{ms}) & $t$ (\si{s}) & $\overline{ape}$ (\si{\degree}) & $\overline{rpe}$ (\si{\degree}) & $t_{es}$ (\si{ms}) & $t$ (\si{s}) \\
    \midrule
    bicycle & 0.092 & 0.045 & 0.664 & 5.009 & 0.104 & 0.045 & 0.533 & 4.996 \\
    town & 0.266 & 0.068 & 0.674 & 4.984 & 0.433 & 0.154 & 0.540 & 5.000 \\
    distant-building & 0.274 & 0.249 & 0.605 & 29.209 & 1.938 & 1.138 & 0.488 & 29.218 \\
    embankment & 0.439 & 0.252 & 0.689 & 31.449 & 0.947 & 0.270 & 0.509 & 31.459 \\
    DL-8 & 0.460 & 0.092 & 0.721 & 4.989 & 17.159 & 0.709 & 0.594 & 4.998 \\
    LT-80 & 0.140 & 0.063 & 0.683 & 80.122 & 0.175 & 0.082 & 0.476 & 80.001 \\
    \bottomrule
    \end{tabular}%
  }
\end{minipage}
\hfill
\begin{minipage}{0.545\textwidth}
  \centering
  \caption{Performance comparison with and without Regional Density Management (RDM)}
  \label{tab:rdm_comparison}
  \resizebox{\textwidth}{!}{%
    \begin{tabular}{c|ccccc|ccccc}
    \toprule
    \multirow{2}{*}{Dataset} & \multicolumn{5}{c|}{w/ RDM} & \multicolumn{5}{c}{w/o RDM} \\
    \cmidrule{2-11}
     & $\overline{ape}$ (\si{\degree}) & $\overline{rpe}$ (\si{\degree}) & $t_{es}$ (\si{ms}) & $t_{kd}$ (\si{ms}) & $t$ (\si{s}) & $\overline{ape}$ (\si{\degree}) & $\overline{rpe}$ (\si{\degree}) & $t_{es}$ (\si{ms}) & $t_{kd}$ (\si{ms}) & $t$ (\si{s}) \\
    \midrule
    bicycle & 0.119 & 0.041 & 0.635 & 234.223 & 4.996 & 0.137 & 0.038 & 0.691 & 409.045 & 5.014 \\
    town & 0.177 & 0.035 & 0.639 & 221.014 & 4.995 & 0.206 & 0.037 & 0.685 & 410.574 & 5.006 \\
    distant-building & 0.240 & 0.236 & 0.551 & 68.388 & 29.218 & 0.261 & 0.228 & 0.595 & 354.598 & 29.217 \\
    embankment & 0.537 & 0.244 & 0.623 & 89.756 & 31.459 & 0.500 & 0.235 & 0.632 & 406.953 & 31.458 \\
    DL-8 & 0.226 & 0.066 & 0.692 & 632.552 & 4.998 & 0.284 & 0.061 & 2.151 & 2235.392 & 13.308 \\
    LT-80 & 0.140 & 0.069 & 0.665 & 1410.348 & 80.122 & 0.165 & 0.064 & 0.726 & 8930.521 & 80.770 \\
    \bottomrule
    \end{tabular}%
  }
\end{minipage}
\end{table*}

%% file: figs_and_tabs/fig_reviewer9_p2p_vs_p2l.tex
\begin{figure}[!t]
    \centering
    {
    \setlength{\fboxrule}{0.1pt}
    {\color{gray}\fbox{\includegraphics[width=0.98\columnwidth,
    trim={0mm} {55mm} {0mm} {25mm}, 
    clip]{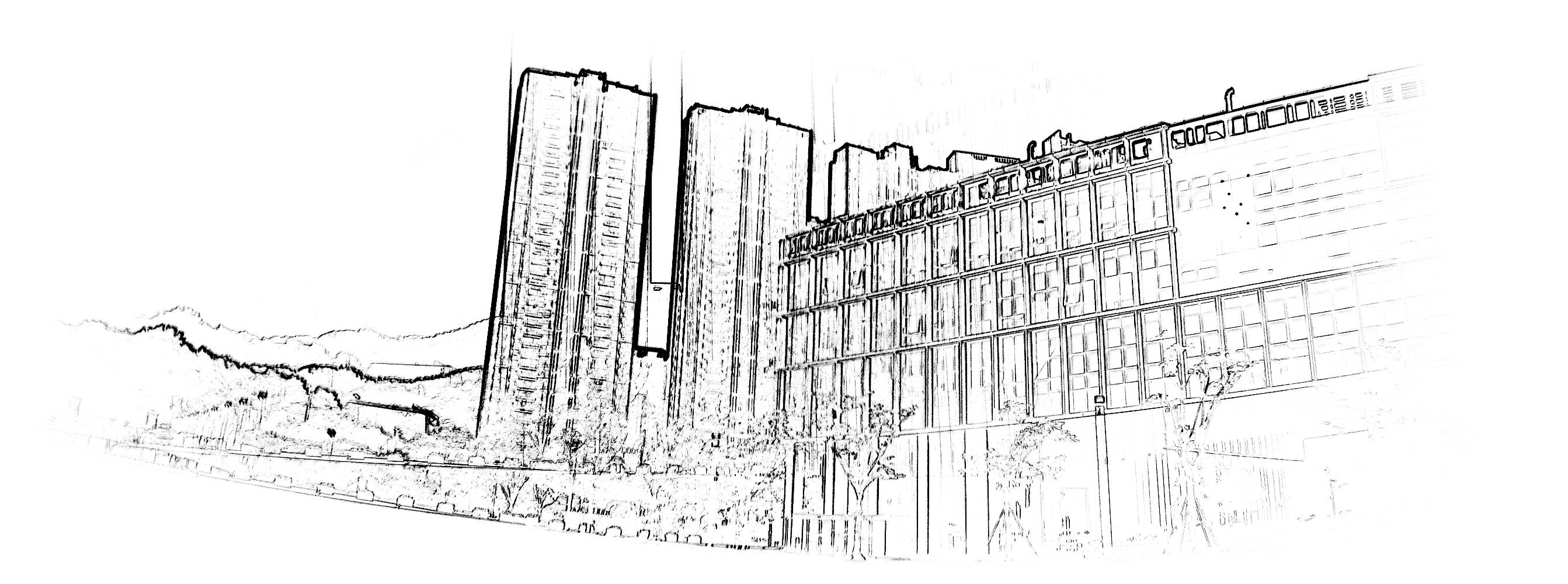}}}
    }
    \caption{Panoramic image generated using Point-to-Point (P2P) ES-ICP algorithm on the \texttt{distant-building} sequence. This result exhibits noticeable ghosting effects and misalignments compared to our proposed Point-to-Line (P2L) method shown in \prettyref{fig:panoramic_distant_building}. The P2P approach suffers from double edges and blurring, especially in building contours and architectural details.}
    \label{fig:p2l_vs_p2p_panorama_comparison}
\end{figure}

%% file: figs_and_tabs/fig_reviewer9_rdm_time_pie.tex
\begin{figure}[!t]
    \centering
    \begin{overpic}[width=0.98\columnwidth,
        trim={2mm} {0mm} {2mm} {0mm},
        clip]{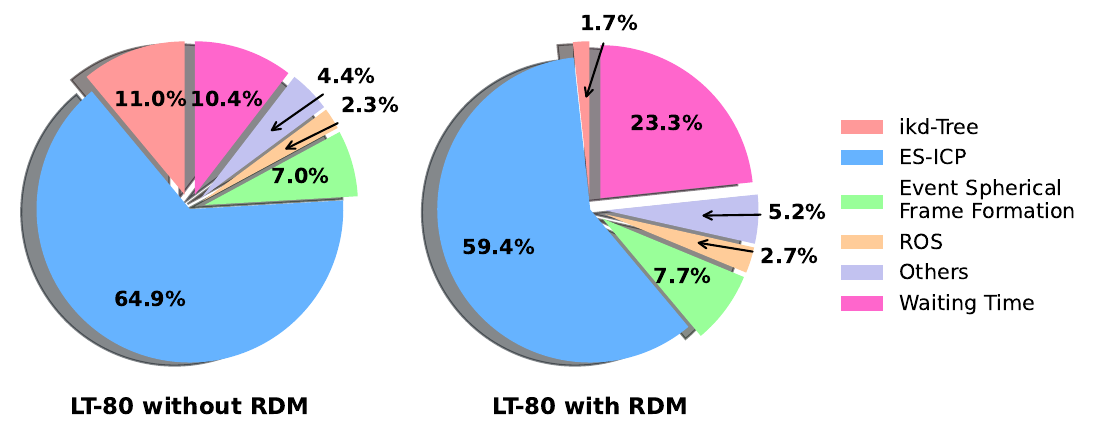}
    \end{overpic}
    \caption{Comparison of computational time distribution across EROAM components on the \texttt{LT-80} sequence with and without Regional Density Management (RDM). The pie charts demonstrate how RDM optimizes processing efficiency by significantly reducing the ikd-Tree operation overhead.}
    \label{fig:rdm_runtime_comparison}
\end{figure}

%% file: sections/limitation.tex
\section{Limitations}
\label{sec:limitations}
\input{figs_and_tabs/fig_robot_platforms}
\input{figs_and_tabs/fig_reviewer8_cear_pano}

\begin{revisedtextblockCR}
While EROAM demonstrates robustness to moderate translational motion as shown in \prettyref{subsec:ecd_robustness}, its performance degrades when translation becomes substantially larger. This limitation stems from the fundamental pure rotational motion assumption underlying our geometric formulation.
\end{revisedtextblockCR}

\begin{revisedtextblockCR}
To illustrate this boundary, we evaluated EROAM on the \texttt{env2\_backflip1} sequence from the CEAR dataset \cite{zhu2024cear}, which features a Mini-Cheetah quadruped robot \cite{katz2019mini} performing a backflip maneuver (\prettyref{fig:cear_quad_backflip}). This scenario combines three challenging factors: (1) large translation (\SI{0.7}{\m}), (2) simultaneous \SI{360}{\degree} rotation, and (3) extreme close-range scenes (\SIrange{0.3}{0.5}{\m}) when facing the floor. Under these conditions, parallax effects become severe and the motion cannot be approximated as pure rotation.
\end{revisedtextblockCR}

\prettyref{fig:panoramic_cear} shows the panoramic reconstruction. While EROAM attempts to align events within its $SO(3)$ framework, the estimated trajectory diverges from actual motion, incorrectly estimating completion of rotation before the backflip finishes. This causes misalignment manifesting as blurring and shadows in the panorama's central region.

\begin{revisedtextblockCR}
This failure case, combined with the successful ECD results (\prettyref{subsec:ecd_robustness}), empirically defines EROAM's operational spectrum: the method maintains functionality with moderate translation (up to \SI{\sim0.3}{\m} as in ECD sequences) but degrades when translation is substantially large and combined with close-range parallax. For applications requiring 6-DOF estimation in such scenarios, full SLAM systems would be more appropriate.
\end{revisedtextblockCR}

%% file: figs_and_tabs/fig_robot_platforms.tex
\begin{figure}[!t]
    \centering
    \subfloat[]{
        \includegraphics[height=3.2cm]{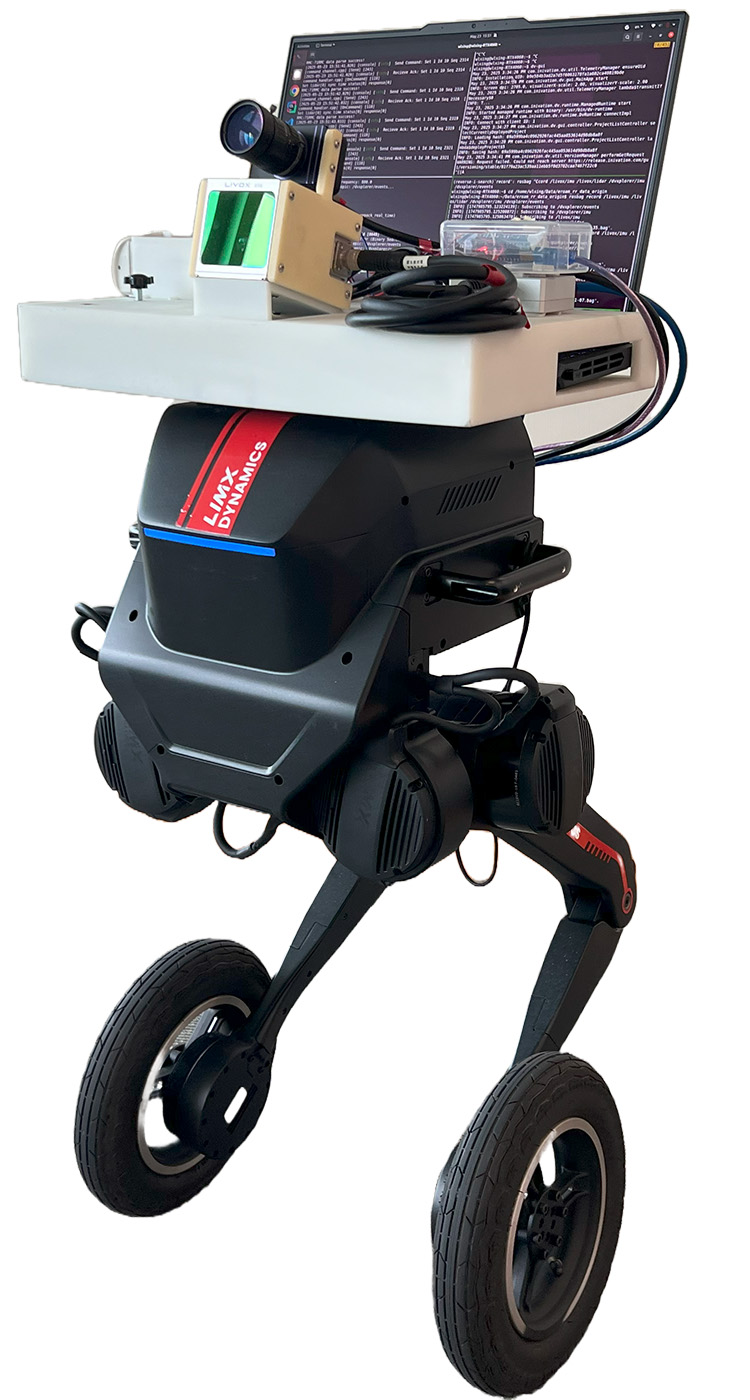}
        \label{fig:limx_tron1_setup}
    }
    \hfill
    \subfloat[]{
        \includegraphics[height=3.2cm]{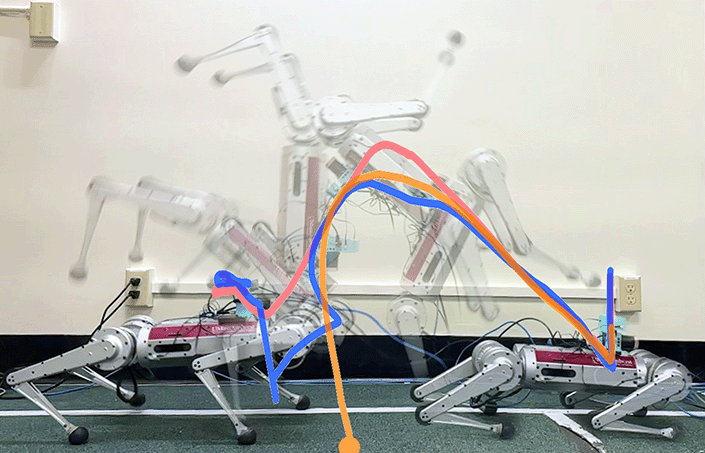}
        \label{fig:cear_quad_backflip} 
    }
    \caption{Event camera-equipped robotic platforms. 
             (a) The LimX TRON 1 wheeled bipedal robot, our experimental platform, equipped with an iniVation DVXplorer event camera. 
             (b) A Mini-Cheetah quadruped robot with an event camera performing a backflip, featured in the CEAR dataset~\cite{zhu2024cear}. Image (b) from~\cite{zhu2023event}.}
    \label{fig:event_camera_robots}
\end{figure}

%% file: figs_and_tabs/fig_reviewer8_cear_pano.tex
\begin{figure}[!t]
    \centering
    {
    \setlength{\fboxrule}{0.1pt}
    
    \color{gray}\fbox{\includegraphics[width=0.98\columnwidth]{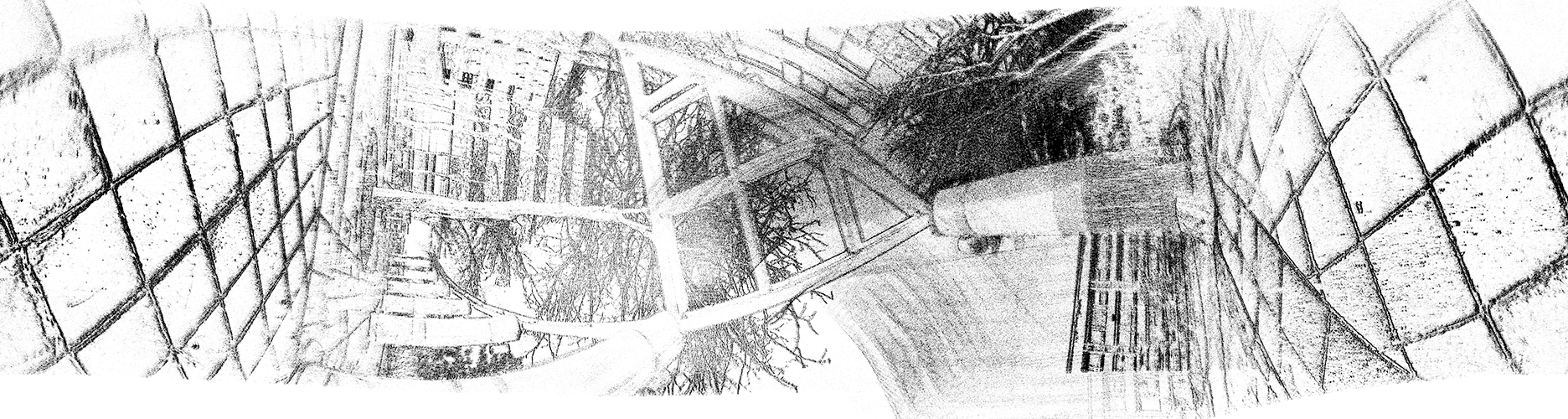}}}%

    \caption{Panoramic image reconstructed from the \texttt{env2\allowdisplaybreaks\_backflip1} sequence in~\cite{zhu2024cear}. The sequence was recorded with an event camera-equipped quadruped robot executing a backflip maneuver. The image is rotated \SI{90}{\degree}.}
    \label{fig:panoramic_cear}%
\end{figure}

%% file: sections/conclusion.tex
\section{Conclusion}

In this paper, we have introduced EROAM, a novel approach to event-based rotational motion estimation. Our method advances the state of the art through several key innovations. First, we propose a spherical representation that simplifies rotational motion formulation \begin{revisedtextblockCR}while operating in a continuous spherical domain\end{revisedtextblockCR}. Second, we develop an Event Spherical ICP algorithm that efficiently processes events in parallel while maintaining high accuracy. Third, we present a real-time implementation that achieves superior performance. Through extensive evaluation on both synthetic and real-world datasets, we have demonstrated EROAM's robust performance across various challenging conditions, including high angular velocities and extended sequences. The success of our ICP-based approach suggests promising potential for broader applications in event camera tasks, offering significant advantages in both computational efficiency and accuracy compared to existing contrast maximization frameworks.